\documentclass[11pt,letterpaper,onecolumn]{IEEEtran}

\usepackage{setspace}
\usepackage[pagebackref=true,breaklinks=true,letterpaper=true,colorlinks,bookmarks=false]{hyperref}
\usepackage{epsfig}
\usepackage{graphicx}
\usepackage{amsmath}
\usepackage{amssymb}

\usepackage{rotating}
\usepackage[belowskip=-6pt,aboveskip=0pt]{caption}
\usepackage[belowskip=0pt,aboveskip=0pt]{subcaption}
\usepackage{threeparttable}
\usepackage{breqn}
\usepackage{tikz}

\graphicspath{{figures/}}

\newcommand{\ud}{\,\mathrm{d}}

\newcommand{\R}{\mathbb{R}}

\newcommand{\ip}[3]{\left< {#1}, {#2} \right>_{#3}}

\newcommand{\comment}[1]{ }

\newcommand{\boundellipse}[3]% center, xdim, ydim
{(#1) ellipse (#2 and #3)
}

\begin{document}
\title{Tracking via Motion Estimation with Physically Motivated
  Inter-Region Constraints}
\author{Omar Arif, Ganesh Sundaramoorthi, Byung-Woo Hong, Anthony
  Yezzi
  \thanks{O.~Arif is with the Department of Electrical Engineering, King Abdullah University of Science and Technology (KAUST), Thuwal, Saudi Arabia}%
  \thanks{G.~Sundaramoorthi is with the Department of Electrical
    Engineering and Department of Applied Mathematics and
    Computational Science, King Abdullah University of Science and
    Technology (KAUST), Thuwal, Saudi Arabia}%
  \thanks{B.~W.~Hong is with the Department of Computer Science,
    Chung-Ang University, Seoul, Korea}
  \thanks{A.~Yezzi is with the School of Electrical \& Computer
    Engineering, Georgia Institute of Technology, Atlanta, USA}
}

\maketitle

\begin{abstract}
  In this paper, we propose a method for tracking structures (e.g.,
  ventricles and myocardium) in cardiac images (e.g., magnetic
  resonance) by propagating forward in time a previous estimate of the
  structures via a new deformation estimation scheme that is motivated
  by physical constraints of fluid motion.  The method employs within
  structure motion estimation (so that differing motions among
  different structures are not mixed) while simultaneously satisfying
  the physical constraint in fluid motion that at the interface
  between a fluid and a medium, the normal component of the fluid's
  motion must match the normal component of the motion of the
  medium. We show how to estimate the motion according to the previous
  considerations in a variational framework, and in particular, show
  that these conditions lead to PDEs with boundary conditions at the
  interface that resemble Robin boundary conditions and induce
  coupling between structures. We illustrate the use of this motion
  estimation scheme in propagating a segmentation across frames and
  show that it leads to more accurate segmentation than traditional
  motion estimation that does not make use of physical
  constraints. Further, the method is naturally suited to interactive
  segmentation methods, which are prominently used in practice in
  commercial applications for cardiac analysis, where typically a
  segmentation from the previous frame is used to predict a
  segmentation in the next frame. We show that our propagation scheme
  reduces the amount of user interaction by predicting more accurate
  segmentations than commonly used and recent interactive commercial
  techniques.
\end{abstract}

\section{Introduction}
Accurate boundary detection of deforming structures from time-varying
medical images (e.g., cardiac MRI) is an important step in many
clinical applications that study structure and function of organs
non-invasively.  While many methods have been proposed to determine
the boundary of a deforming structure by segmenting each frame based
on image intensity statistics and incorporating training data
(see Section~\ref{subsec:segmentation_related}), it is sometimes
easier to exploit the temporal coherence of the structure, and apply a
tracking framework. That is, one matches a current estimate of the
structure of interest at time $t$ to the image at time $t+1$ to detect
the organ at time $t+1$. This requires an accurate registration
between images.

A difficulty in registration stems from the aperture problem - many
different registrations are able to explain two images, and therefore,
regularization is needed to constrain the set of possible
solutions. Typically, global regularization is used. However, the
image consists of many structures, each moving with different
motions/deformations, and global regularization implies that smoothing
is performed across multiple structures\footnote{It may seem enough to
  register only the organ at time $t$ (not the entire image) to a
  subset of the image at time $t+1$, thus smoothing within the
  organ. However, the approach is problematic as the background
  registration helps limit the possible registrations of the organ of
  interest, aiding the registration of the organ. See
  Figures~\ref{fig:registrations_LV} and \ref{fig:registrations_RV}
  for an experiment.}. Therefore, motion/deformation information from
surrounding structures is used in the registration estimate within the
structure of interest; this leads to errors in the registration and in
the structure segmentation.

\comment{
While
regularization within organs is desired, that alone does
not satisfy physically motivated constraints from fluid
mechanics. Indeed, the motion of a fluid within a surrounding medium
is such that the normal component of the motion of the fluid at the
boundary matches the normal component of the motion of the surrounding
medium. This is derived from the observation that the fluid and the
medium do not separate. Further, the No-Slip condition of fluid
mechanics states that the fluid's motion relative to the boundary is
zero.
}

In this work, we derive a registration method that estimates motion
separately in each structure by performing only within structure
regularization (so that motions from heterogeneous structures are not
mixed) while satisfying physical motion constraints (from fluid
mechanics) across the boundary between two structures.  Specifically,
the physical constraint is that the motion of a fluid in a medium at
the interface is such that the normal components of the motions are
the same. We derive a motion estimation scheme that incorporates these
physical considerations.  We use our new motion estimation scheme in a
tracking algorithm to segment structures.  That is, given the initial
segmentation in the first frame, our algorithm automatically
propagates the segmentation to the next frame.  Although our
methodology is not restricted to a particular imaging modality or a
particular structure, we focus on a prominent application where the
physically motivated considerations are natural - segmenting the left
(LV), right ventricle (RV), and the surrounding heart muscle from
cardiac MRI. We demonstrate that our new motion estimation scheme
leads to more accurate segmentation of the LV and RV than using
traditional global regularization.

Further, the main motivation for our new motion estimation scheme is
segmentation propagation in interactive segmentation of image
sequences. Segmentation propagation is a basic step to predict the
segmentation in the next frame from the current frame in interactive
methods. Interactive approaches are still the norm in commercial
medical applications (particularly cardiac MRI) as fully automated
segmentation (see discussion in
Section~\ref{subsec:segmentation_related}) is still at the research
stage. An accurate segmentation propagation reduces the amount of
interaction for the user. We show that our propagation scheme, with
physically viable motion, leads to better segmentation propagation
than recent existing commercially available software for cardiac
segmentation, and would thus require less interaction.

\comment{
, and such In interactive segmentation of image sequences, it is common to
use algorithms to predict the segmentation in the next frame from
previous frame, and thus accurate propagation schemes are needed to
reduce the amount of work to the user.

We also demonstrate that
our physically motivated motion estimation scheme leads to better
segmentation propagations than 
}

\subsection{Related Work: Cardiac Segmentation}
\label{subsec:segmentation_related}

We now give a brief review of the cardiac image segmentation
literature, which places our work in appropriate context. For a more
thorough review of literature, we direct the reader to recent review
articles \cite{petitjean2011review,zhuangchallenges2013}.

Early methods for automatic segmentation of cardiac images involve the
use of image partitioning algorithms (e.g., active contours
\cite{kass1988snakes} implemented via level sets
\cite{osher1988fronts}, graph cuts \cite{boykov2001fast}, or convex
optimization methods \cite{bresson2007fast}), which optimize energies
that integrate basic image features such as edges \cite{pham2001fem},
intensity statistics \cite{cordero2011unsupervised}, motion cues
\cite{cremers2005motion}, and basic smoothness priors of the
partition. These methods are good at partitioning the image into
regions of homogeneous statistics, however, the regions of the
partition do not typically select regions that correspond to physical
objects/structures (e.g., the left/right ventricle, or the
myocardium). Therefore, there have been two approaches to augment
basic segmentation algorithms to determine object boundaries: {\bf
  training-based} approaches where manually segmented objects from
training images are used to construct a model which is then used in
automatic segmentation, and the second approach is {\bf interactive}
approaches where human interaction is used to correct errors from
basic segmentation approaches.

Since our contribution lies in reducing the amount of interaction
required in interactive approaches, we give only a brief review of
training based approaches next before moving to review interactive
approaches. In training-based approaches, training data is used to
construct a model of the heart. Early approaches constructed models of
the heart manually by using simple geometric approximations of the
heart made by observing training images, e.g., a generic model of the
heart constructed from truncated ellipses \cite{sermesant2006cardiac}
that are used to model the ventricles. More specific models for the
heart tailored to the training data use the training data of manually
segmented organs more directly. Some approaches (e.g.,
\cite{stegmann2005unsupervised,van2006spasm,koikkalainen2008methods,heimann2009statistical,zhuang2010registration,zhang20104,zhang2011deformable})
make use of active shape and appearance models
\cite{cootes1995active,cootes2001active} where manual landmarks around
the boundary of the object in training images model the shape, and
texture descriptors describing a neighborhood around the landmark are
used to model object appearance. Such landmarks and descriptors allow
for natural use of PCA to generate a statistical model.  More precise
models of shape are based on performing PCA of segmented objects using
mesh-based approaches \cite{ecabert2008automatic} or geometric level
set representations \cite{tsai2003shape,paragios2003level}. The
previous methods construct static models of the heart, however, the
heart is a dynamic object, and thus, dynamic models of shape
\cite{schaerer2010dynamic,zhu2010segmentation} are constructed by
considering the shape from multiple frames in the sequence as a
time-varying object. Once the heart model is constructed from training
data, to perform object segmentation of an image, the model must be
fitted to the image. This can be done in a number of ways, the most
common method for shape models
(e.g. \cite{chen2002using,cremers2003shape,tsai2003shape}) is to
restrict the optimization of the energies based on basic image
features (discussed in the previous paragraph) to the shapes
determined by the parametric shape model. Other approaches that have
both shape and appearance information deform the average
shape/appearance, i.e., the atlas, via registration to fit the target
image, thereby determining the object segmentation
\cite{lorenzo2004segmentation,zhuang2010registration,Isgum:TMI:09,kirisli2010fully}.

Although a fully automatic solution to segmentation of the heart (as
in the methods above) and its sub-structures is the ideal goal, these
methods are not accurate enough (especially when there is deviation
from the training set, e.g., in cases of disease) to be used in many
cardiac applications (e.g.,
\cite{zhuangchallenges2013,lalande1999automatic,paragios2002variational,tsai2003shape,kaus2004automated,senegas2004model,lotjonen2004statistical,lin2006automated,van2006spasm,lynch2006automatic}). Therefore,
in practical commercial applications, interactive approaches to
segmentation in which a user corrects the prediction of automatic
segmentation are still the norm. Various techniques in the computer
vision community (e.g.,
\cite{Xue2005,boykov2000interactive,cremers2007probabilistic}) have
been designed to incorporate interaction in the form of seed points
(belonging to the object and the background). These methods modify
energies based on simple image features (discussed in the first
paragraph) to incorporate constraints from the seed points entered by
the user. Other interactive methods allow for a manual or
semi-automated segmentation of the first frame in the cardiac sequence
and then attempt to propagate the segmentation to sub-sequent frames,
thereby predicting a segmentation that needs little interaction to
correct \cite{schnabel2001generic,noble2002myocardial}. Several
methods exist to propagate the initial segmentation (e.g., by
registration \cite{schnabel2001generic,noble2002myocardial}) and/or by
using the manual segmentation in the first frame as initialization to
an automated segmentation algorithm. Several commercial softwares for
interactive heart segmentation have been designed. For example, the
recent software Medviso \cite{heiberg2010design,sjogren2012semi}
allows the user to input an initial segmentation, which is then
propagated to subsequent frames in order to segment various structures
including the ventricles and myocardium. The software also allows for
various other manual interactions to correct any errors in the
propagation. The algorithm is a culmination of many techniques
including registration to propagate the segmentation as well as the
use of automated methods training data to encode shape priors.

\subsection{Related Work: Registration}

The goal of our method is to improve the propagation method in
interactive techniques so that less interaction is required by the
user. We do so by deriving a registration technique that better models
the underlying physics of the motion of ventricles and myocardium, in
particular constraints formed from interactions between adjacent
regions (e.g., ventricles and the surrounding muscle). Since our works
relates to registration, we give a brief review of recent related work in
registration.

The goal of registration is to find pixel-wise correspondence between
two images in a sequence. The difficulty arises from the aperture
ambiguity: there are infinitely many transformations that map one
image to the other, and thus regularization must be used to constrain
the solution. The pioneering work \cite{horn1981determining} from the
computer vision literature uses a uniform global smoothness penalty to
estimate the registration under small pixel displacements. Larger
deformations with global regularization that lead to diffeomorphic
registrations, a property of a proper registration in typical medical
images, has been considered by
\cite{beg2005computing,vercauteren2008symmetric}. For cardiac images
(as well as other medical images), there are multiple objects
(sub-structures) that each have different motion characteristics, thus
global regularization across adjacent objects (mixing heterogenous
motion characteristics) is not desired, and moreover, there are
physical constraints between adjacent regions that a registration
based on global regularization does not satisfy. In particular, in
cardiac applications (and others), the ventricles and the heart muscle
(myocardium), both mostly fluids, are such that the normal component
on the boundary of the motions (velocity) of both structures are
equal. The previous constraint is from the fact that the ventricle and
the surrounding muscle do not separate during motion.  Further, the
No-Slip condition \cite{munson1990fundamentals} from fluid mechanics
for viscous fluids states that the motion of the fluid relative to the
boundary (the tangent component) is zero. The scale at which this
happens is small compared with the resolution of the imaging device,
and thus we allow the tangent components to be arbitrary across the
boundary, although our technique can easily be modified if the No-Slip
condition is desired to be enforced.

There has been recent work in the medical imaging literature that has
considered other forms of regularization rather than global
regularization. Indeed, in the case of lung registration, the lung
slides along the rib-cage, and the motion of both these structures are
different and thus global regularization is not desired. In
\cite{pace2011deformable,schmidt2012fast} an anisotropic global
regularization is used to favor smoothing in the tangential direction
near organ boundaries. In \cite{risser2011diffeomorphic}, Log-Demons
\cite{vercauteren2008symmetric} is generalized so that smoothing is
performed on the tangent component of the registration within organs,
and the normal component is globally smoothed on the whole image. Our
approach differs from these works in that our technique is motivated
by the physical constraints of fluid motion present in the
heart. Further, in our method, regularization is only performed within
homogeneous structures (different than \cite{risser2011diffeomorphic}
which smoothes the normal component across structures, mixing
inhomeogeneous motions), and smoothing along the tangential direction
of the boundary in \cite{pace2011deformable,schmidt2012fast} does not
necessarily ensure that the normal motions equality on the
boundary. While \cite{risser2011diffeomorphic} achieves equal normal
motions, it does so by smoothing across tstructures.

\comment{
whereas our
approach is done in a more generic way, having smoothness in the
normal direction if the data dictates it (which is not typically the
case in heart data where there is in fact only continuity but not
smoothness of the normal component).
}

\subsection{Organization of Paper}
In Section~\ref{sec:motion_model}, we specify the motion and
registration model between frames in an image sequence, in particular,
the motion constraint between structures. In
Section~\ref{sec:infinitesimal_deformation}, we use the motion model
to setup a variational problem for estimating the motion given the two
images assuming infinitesimal motion, and then show how to estimate
the motion using two different methods. In
Section~\ref{sec:infinitesimal_deformation}, we show how to estimate
non-infinitesimal motion that are typical between frames and
simultaneously propagate the segmentation from the previous frame to
the next frame. Finally, we in Section~\ref{sec:experiments} show a
series of experiments to verify our method as well as compare it to an
existing recent commercial software package for interactive cardiac
segmentation.

\section{Motion and Registration Model}
\label{sec:motion_model}

In this section, we state the assumptions that we use to derive our
method for piecewise registration whose normal component is continuous
across organ boundaries. We assume the standard brightness constancy
plus noise model:
\begin{equation} \label{eq:brightness_constancy}
  J_1(x) = J_0(w^{-1}(x)) + \eta(x), \, x\in \Omega
\end{equation}
where $\Omega\subset \R^n$ ($n=2,3$) is the image domain, $J_0, J_1 :
\Omega \to \R$ are the images sampled from the time-varying imagery at
two consecutive times, $w : \Omega \to \Omega$ is the registration
between frames $J_0$ and $J_1$, and $\eta$ is a noise process. The
structure of interest in $J_0$ is denoted $R \subset \Omega$ and the
background is denoted $\Omega \backslash R$. Our model can be extended
to any number of organs, but we forgo the details for simplicity of
presentation. The structure in frame $J_1$ is then assumed to be
$w(R)$.

The registration is an invertible map, and thus we represent it as an
integration of a time varying infinitesimal velocity field (following
standard representation in the fluid mechanics literature):
\begin{equation} \label{eq:warp_vel}
  w(x) = \phi_T(x), \, \phi_s(x) = x+\int_0^s
  v_{\tau}(\phi_{\tau}(x)) \ud \tau, \, s\in [0,T]
\end{equation}
where $T>0$, $v_{\tau} : \Omega \to \R^n$ is a velocity field, and
$\phi_{\tau} : \Omega \to \Omega$ for every $\tau \in [0,T]$. The map
$\phi_{\tau}$ is such that $\phi_{\tau}(x)$ indicates the mapping of
$x$ after it flows along the velocity field for time $\tau$, which is
an artificial time parameter.

We assume that the motion/deformation of the structure of interest $R$
and the surrounding region have different characteristics and
therefore the registration $w$ consists of two components (one can
easily extend to more components, but we use only two for simplicity
of notation), $w^{in}$ and $w^{out}$ defined inside the organ of
interest $R$ and outside the organ $\Omega \backslash R$, resp.  This
can be achieved with a velocity field that has two components
$v^{in}_{\tau}, v^{out}_{\tau}$ (both smooth within their domains):
\begin{equation} \label{eq:warp_decomp}
  w(x) = 
  \begin{cases}
    w^{in}(x) & x \in R \\
    w^{out}(x) & x \in \Omega \backslash R
  \end{cases},\quad 
  v_{\tau}(x) = 
  \begin{cases}
    v^{in}_{\tau}(x) & x \in R_{\tau} \\
    v^{out}_{\tau}(x) & x \in \Omega\backslash R_{\tau}
  \end{cases},
\end{equation}
where $R_{\tau}=\phi_{\tau}(R)$. This implies that $w^{in}$ and
$w^{out}$ are smooth and invertible. When the structure $R$ contains a
fluid (as in the ventricles) and $\Omega\backslash R$ is the
surrounding medium (e.g., myocardium), the normal as in our case of
interest, there is a physical constraint from fluid mechanics imposed
on $v^{in}_{\tau}$ and $v^{out}_{\tau}$ at the boundary $\partial
R_{\tau}$. The constraint is that the normal component of
$v^{in}_{\tau}$ and $v^{out}_{\tau}$ are equal on $\partial R_{\tau}$:
\begin{equation} \label{eq:vel_cont}
  v^{in}_{\tau}(x) \cdot N(x) = v^{out}_{\tau}(x) \cdot N(x), \, x\in \partial R_{\tau},
\end{equation}
where $N$ indicates the surface normal of $\partial R_{\tau}$. The
condition implies that the two mediums $R$ and $\Omega\backslash R$ do
not separate when deformed by the infinitesimal motion.  Further, the
No-Slip Condition from fluid mechanics \cite{munson1990fundamentals}
implies that the motion of the fluid relative to the surrounding
medium is zero at the interface, thus, the tangent components of
$v^{in}_{\tau}$ is zero. However, the scale at which the tangential
component is approximately zero may be at a scale much smaller than
determined by the resolution of the imaging device. Therefore, we do
not enforce this constraint, although if desired, the constraint can
easily be incorporated in our framework in the next section.

Modeling the velocity with two separate components allows for
different deformation for the structure of interest and surrounding
medium, and the constraint \eqref{eq:vel_cont} couples the two
components of velocity, in a physically plausible manner. This implies
\emph{continuity} of the normal component of $v_{\tau}$ across
$\partial R_{\tau}$, but not necessarily differentiability, as
dictated by materials of different chemical composition.

\section{Energy-Based Formulation for Infinitesimal Deformations}
\label{sec:infinitesimal_deformation}

In this section, we consider the case when the registration can be
approximated as $w(x)=x+v(x)$ where $v$ is an infinitesimal
deformation, and we show how one computes $v$ such that
\eqref{eq:vel_cont} is satisfied while applying regularization only
within regions and not across regions.

We design an optimization problem so that the solution determines the
infinitesimal deformation $v$ of interest.  In a tracking framework,
$R$ is given (e.g., at the initial frame or the previous estimate from
a previous frame), and the goal is to determine $v$ and then $R+v=\{
x+v(x) \,:\, x\in \Omega \}$ is the object in the next frame (assuming
the deformation between frames is small).  The energy that we consider
is
\begin{align}
  \label{eq:energy_lin}
  E(v_i, v_o; R, I) &= 
  \frac 1 2 \int_R 
  \left( |J_1- I+\nabla I \cdot v_i|^2 +  \alpha_i |\nabla v_i|^2\right) \ud x + 
  \frac 1 2 \int_{\Omega\backslash R} 
  \left( |J_1- I+\nabla I \cdot v_o|^2 +  \alpha_o |\nabla v_o|^2\right) \ud x\\
  \label{eq:boundary_constraint}
  \mbox{subject to } & v_i\cdot N =  v_o \cdot N \mbox{ on } \partial R
\end{align}
The infinitesimal deformations are defined $v_i : R\to \R^n$ and $v_o
: \Omega\backslash R \to \R^n$ (as in \eqref{eq:warp_decomp}). The
operator $\nabla$ is the spatial gradient defined within $R$ or
$\Omega\backslash R$ (without crossing $\partial R$).  Note that the
first term in each of the above integrals arises from a linearization
of \eqref{eq:brightness_constancy} (when $I=J_0$), and due to the
well-known aperture problem, regularization (the second term in each
of the integrals) is required to invert \eqref{eq:warp_decomp}. It
should be emphasized that the regularization is done \emph{separately}
within each of the regions.  The weights $\alpha_i, \alpha_o>0$
indicate the amount of regularity desired within each region.

\begin{figure}
  \centering
  {\small 
    \begin{tabular}{c@{}c@{}c@{}c}
      \includegraphics[width=.2\linewidth,clip=true, trim=20 30 20
      30]{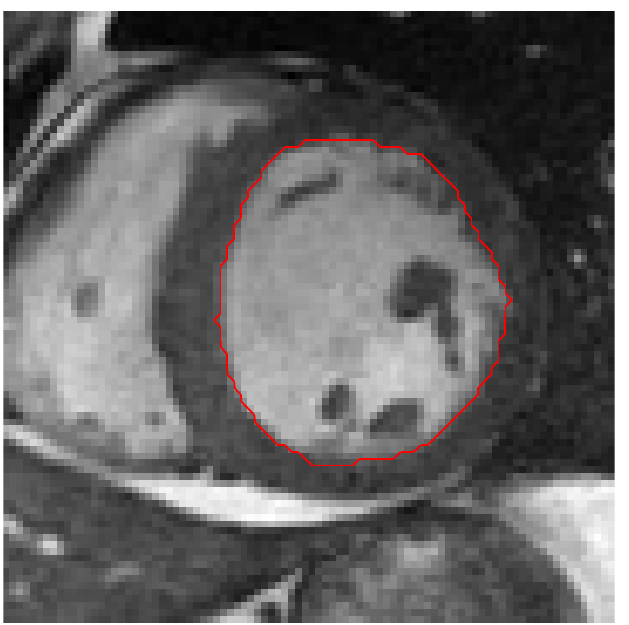} &
      \includegraphics[width=.2\linewidth,clip=true, trim=20 30 20
      30]{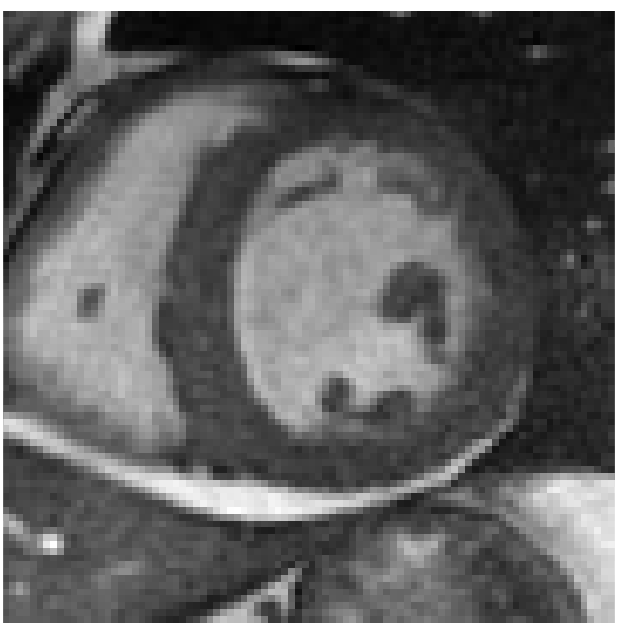} &
      \includegraphics[width=.27\linewidth]{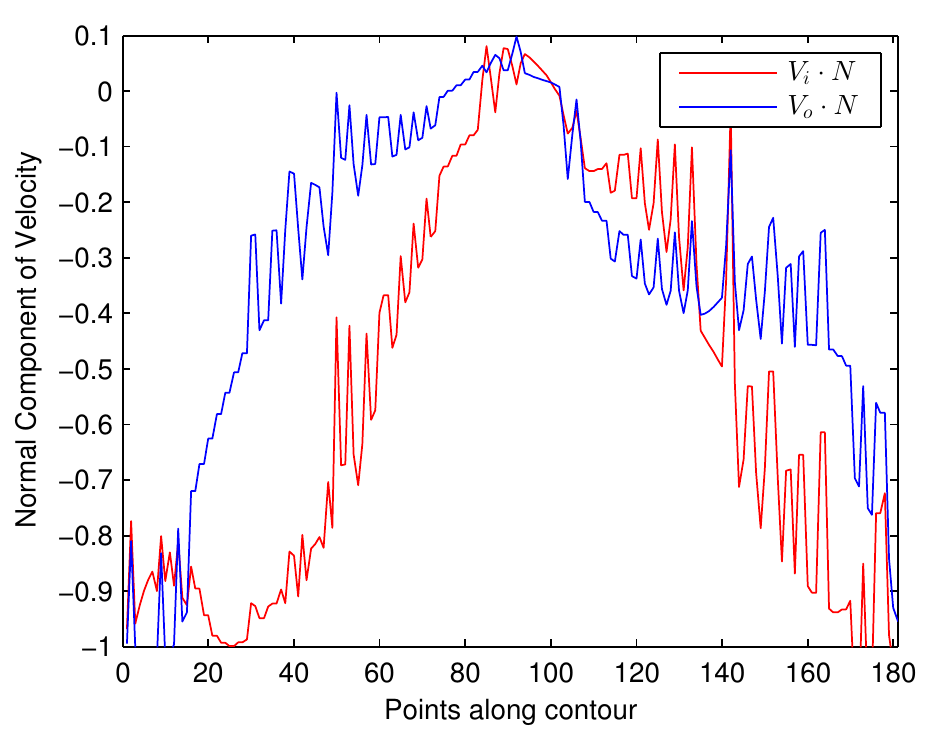} &
      \includegraphics[width=.27\linewidth]{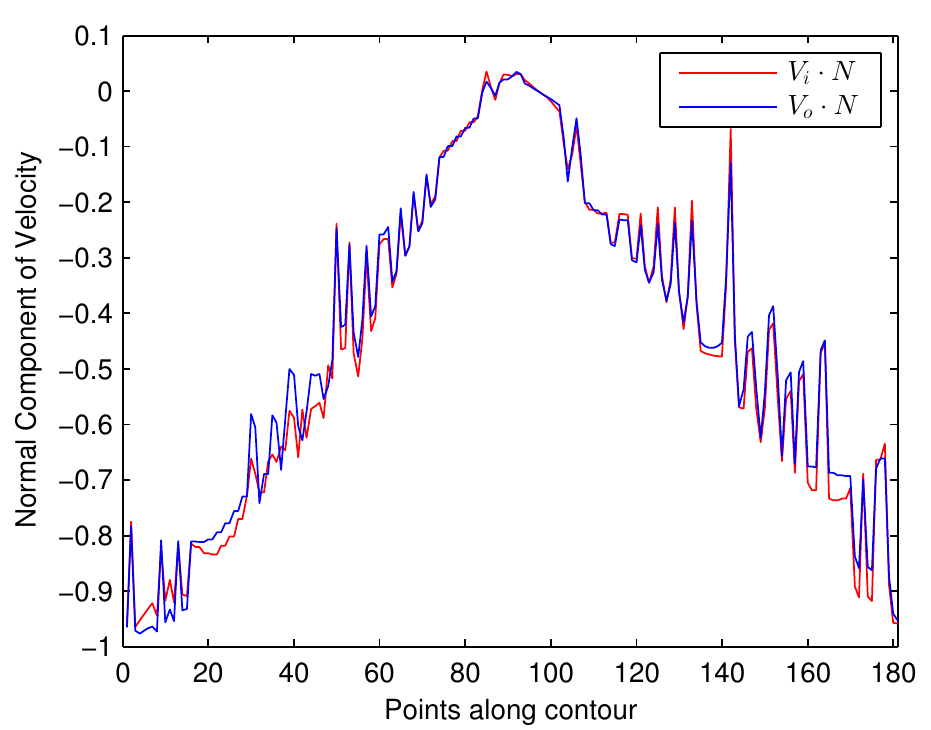} \\
      image + boundary & next image & within region optical flow & our method \\
    \end{tabular}
  }
  
  \caption{{\bf Regularization within regions alone does not
      necessarily induce matching normal motions.}  [First]: $I_1$
    with the boundary of $R$, $\partial R$, marked in red. [Second]:
    $I_2$. [Third]: Normal components of the velocities just inside
    and just outside $\partial R$ using traditional optical flow
    computation separately within $R$ and $\Omega\backslash R$.
    Notice that the normal components of the velocities inside and
    outside $\partial R$ are different, and therefore not
    physically viable.  [Fourth]: The result of our approach using
    regularization only within regions, while satisfying the normal
    motion matching constraint: normal motions along $\partial R$ are
    equal.}
\label{fig:normal_constraint_expt}
\end{figure}

\begin{figure}
  \centering
  {\small 
    \begin{tabular}{c@{}c@{}c@{}c}  
      \includegraphics[width=.24\linewidth, clip=true, trim=0 33 0 33]{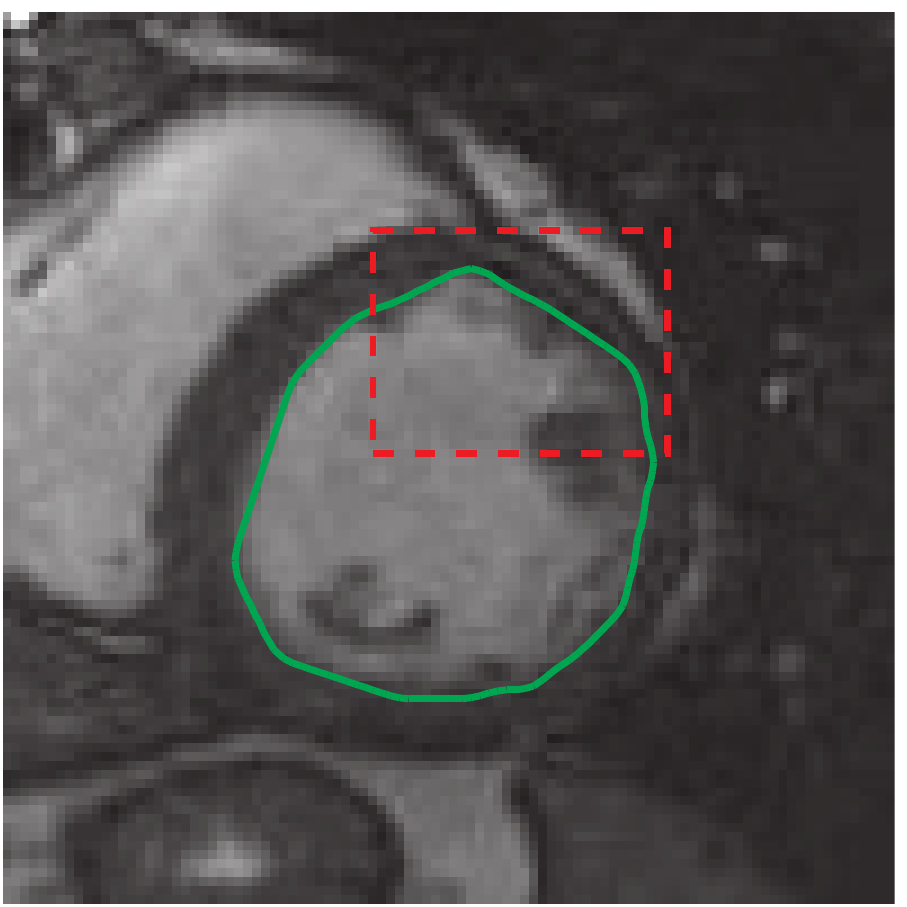} &
      \includegraphics[width=.24\linewidth]{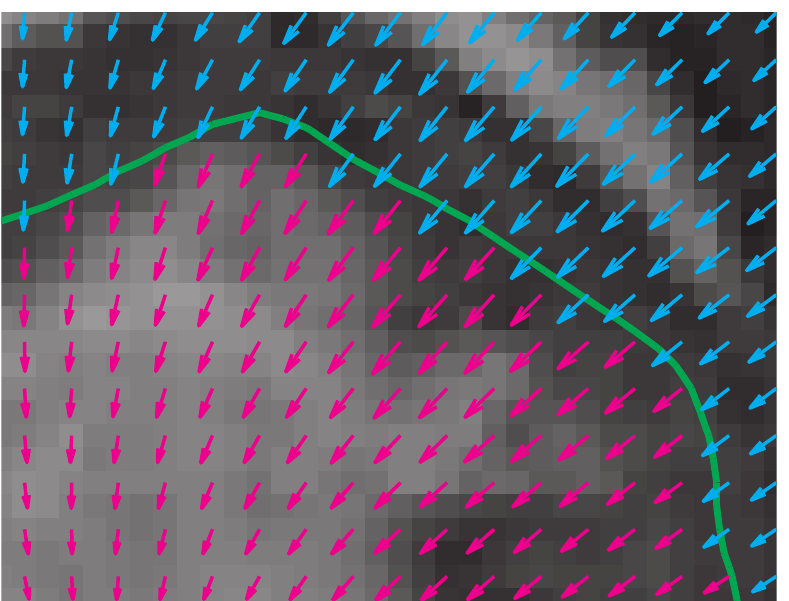} &
      \includegraphics[width=.24\linewidth]{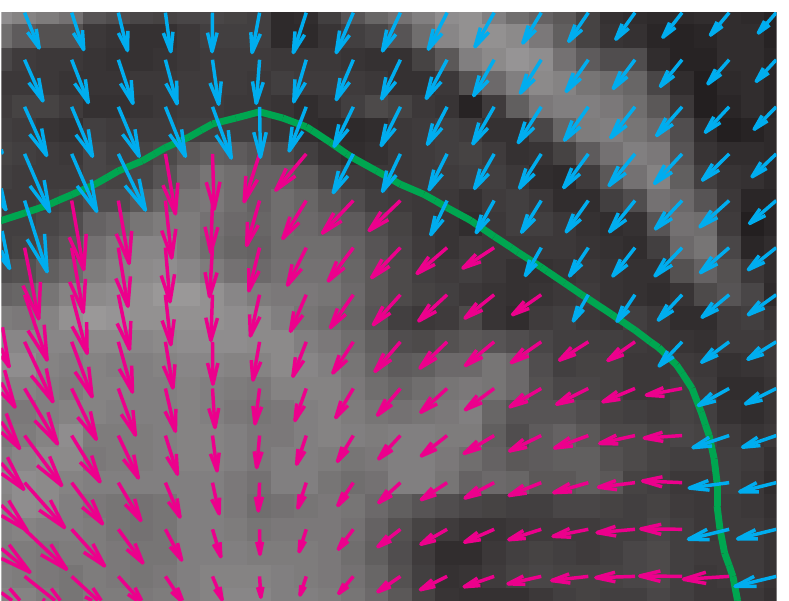} &
      \includegraphics[width=.25\linewidth,clip=true, trim=0 4 0 5]{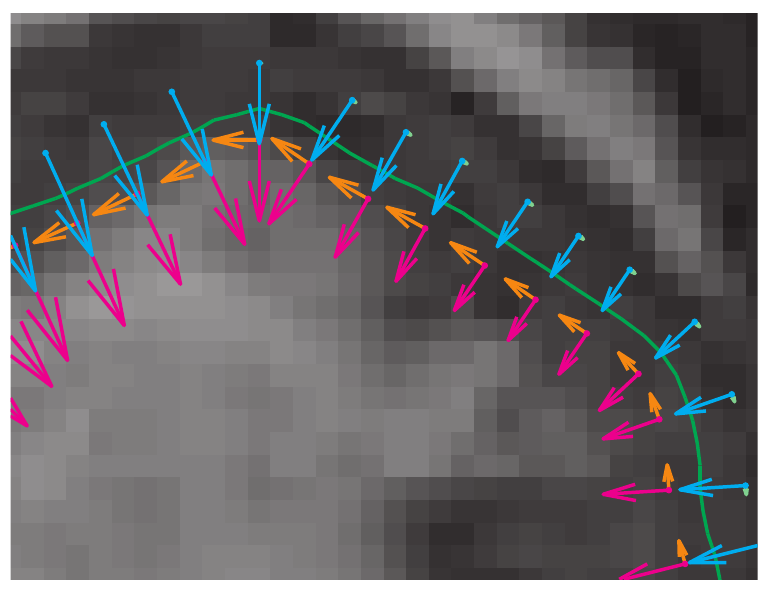}\\
      image + boundary & global optical flow & our method & 
      our method (motion decomp.)
    \end{tabular}
  }
  \caption{{\bf Why regularization within regions and matching motion
      normals constraint?} Heart is contracting. [First]: $J_0$,
    $\partial R$ in green, and the red box is zoomed in subsequent
    images. [Second]: velocity field with global regularization shows
    smoothing across the boundary. [Third]: velocity fields
    $v^{in},v^{out}$ with within region regularization and normal
    constraint shows clear difference to global
    regularization. [Fourth]: normal and tangent components of
    inside/outside velocities $v^{in},v^{out}$; notice the matching
    normal components, but discontinuity of tangent components across
    boundary (tangent component outside is nearly zero).  Outside
    boundary velocity indicates contractive behavior of myocardium,
    and tangent component inside indicates some circulation near
    boundary: cannot be captured with globally smooth regularization.}
  \label{fig:global_vel_comp}
\end{figure}

The problem above is a generalization of Horn \& Schunck optical
flow. Note that solving for the Horn \& Schunck optical flow within
each region separately does not lead to motions such that at the
interface, they have equal normal components (see
Figure~\ref{fig:normal_constraint_expt}), whereas the solution of
\eqref{eq:energy_lin} to be presented in subsequent sections does.
Note that computing Horn \& Schunck optical flow in each region
requires boundary conditions (and typically they are chosen to be
Neumann boundary conditions: $\nabla v_i \cdot N =0$ and $\nabla v_o
\cdot N = 0$ on $\partial R$). Note that replacing these boundary
conditions with the boundary constraint \eqref{eq:boundary_constraint}
does not specify a unique solution. Also, while Horn \& Schunck
optical flow computed on the whole domain $\Omega$ naturally gives a
globally smooth motion, which by default satisfies matching normals at
the interface, this is not natural for the ventricles / myocardium,
where different motions exist in the regions (see
Figure~\ref{fig:global_vel_comp}), and the motions should not be
smoothed across the regions.

In the subsections below, we show how the matching normal constraint
can be enforced approximately and exactly. This leads to PDEs for
$v_i$ and $v_o$ that are coupled with boundary conditions that are not
seen in optical flow estimation and medical image registration.

\subsection{Energy Optimization with a Soft Constraint}
The first approach to optimize \eqref{eq:energy_lin} subject to
\eqref{eq:boundary_constraint} is to construct a new energy that
includes a term that penalizes deviations away from
\eqref{eq:boundary_constraint}. This greatly favors deformations that
satisfy the constraint, but the optimization of the energy does not in
general satisfy the constraint exactly. We consider the following
energy:
\begin{equation}
  E_s(v_i, v_o; R, I) = E(v_i, v_o; R, I) + 
  \frac{\beta}{2} \int_{\partial R} (v_i\cdot N - v_o\cdot N)^2 \ud s
\end{equation}
where $\beta > 0$. Note that the space of $v_i, v_o$ is a linear
space, and the energy $E_s$ is convex and thus, any local optimum is a
global optimum.  To find the necessary conditions, we compute the
first variation of $E_s$. Let $h_i : R\to\R^2$ and $h_o :
\Omega\backslash R \to \R^2$ be perturbations of $v_i$ and $v_o$, then
the first variation in $(h_i, h_o)$ (after simplification) is
\begin{align}
  dE(v_i,v_o)\cdot (h_i,h_o) &= 
  \int_R 
  \left[ (J_1- I + \nabla I \cdot v_i) \nabla I - \alpha_i\Delta v_i
  \right] \cdot h_i \ud x \\ 
  &+ 
  \int_{\Omega\backslash R} 
  \left[ (J_1- I + \nabla I \cdot v_o) \nabla I - \alpha_o\Delta v_o \right] \cdot h_o \ud x
  \\
  &+
  \int_{\partial R} 
  \left[ 
    \alpha_i (\nabla v_i \cdot N) +
    \beta (v_i\cdot N - v_o\cdot N) N 
  \right] \cdot h_i\ud s 
  \\
  & +
  \int_{\partial R}
  \left[ 
    \alpha_o (\nabla v_o \cdot N) +
    \beta (v_i\cdot N - v_o\cdot N) N   
  \right]
  \cdot h_o \ud s
\end{align}
where the first two terms in the two boundary integrals are obtained
from integration by parts, and $\Delta$ denotes the Laplacian
operator. The necessary conditions for a minimum are obtained by
choosing $v_i,v_o$ such that $dE(v_i,v_o)\cdot (h_i,h_o)=0$ for all
$h_i,h_o$, and thus, we have
\begin{align}
  \label{eq:vin_soft_PDE}
  -\alpha_i \Delta v_i + \nabla I \nabla I^T v_i &= -(J_1-I) \nabla
  I, \quad \mbox{in } R
  \\
  \label{eq:vout_soft_PDE}
  - \alpha_o \Delta v_o + \nabla I \nabla I^T v_o &= -(J_1-I) \nabla
  I, \quad \mbox{in } \Omega\backslash R \\
  \label{eq:boundary_soft_first}
  \alpha_i \nabla v_i \cdot  N + \beta (v_i\cdot N - v_o\cdot N) N &= 0, 
  \quad \mbox{on } \partial R \\
  \label{eq:boundary_soft_last}
  \alpha_o \nabla v_o \cdot N + \beta (v_i\cdot N - v_o\cdot N) N &= 0, 
  \quad \mbox{on } \partial R
\end{align}
where ${}^T$ indicates transpose.  The last two boundary conditions
are Robin boundary conditions (these conditions are constraints on a
linear combination of the normal derivatives of the functions and the
function values on the boundary). These boundary conditions specify a
unique solution of the PDE. We note that $v_i$ and $v_o$ are linked to
each other inside and outside by the boundary conditions (unlike
separate solution of $v_i$ and $v_o$ using Neumann boundary conditions
as in traditional optical flow). Such a link between $v_i$ and $v_o$
is expected given that the normal components on the boundary are to be
close, and $v_i$ and $v_o$ are to be smooth within their respective
regions of definition.

\subsection{Energy Optimization with the Hard Constraint} 

We now show how to optimize \eqref{eq:energy_lin} subject to
\eqref{eq:boundary_constraint} by enforcing
\eqref{eq:boundary_constraint} exactly. We optimize $E$ in
\eqref{eq:energy_lin} among $v_i$ and $v_o$ that satisfy
\eqref{eq:boundary_constraint} exactly. The space of $v_i,v_o$
satisfying the normal constraint is a linear space, and the
energy is convex, and so any local optimum must be a global
optimum. Therefore, we now compute the first variation of $E$
evaluated at $(v_i,v_o)$ applied to a perturbation $h_i, h_o$ in the
permissible space (those that perturb $v_i,v_o$ so that the constraint
is satisfied). Note that the space of permissible perturbations also
satisfy the normal matching constraint: $h_i\cdot N=h_o\cdot N$ on
$\partial R$ (this is obtained by differentiating the constraint in
the direction of $h_i,h_o$). The variation is
\begin{align}
  dE(v_i,v_o)\cdot (h_i,h_o) &= 
  \int_R 
  \left[ (J_1- I + \nabla I \cdot v_i) \nabla I - \alpha_i\Delta v_i
  \right] \cdot h_i \ud x 
  \nonumber \\
  &+ 
  \int_{\Omega\backslash R} 
  \left[ (J_1- I + \nabla I \cdot v_o) \nabla I - \alpha_o\Delta v_o \right] \cdot h_o \ud x
  \nonumber\\
  &
  \label{eq:variation_E}
  +
  \int_{\partial R} \alpha_i (\nabla v_i \cdot N)\cdot h_i \ud s 
  - \int_{\partial R} \alpha_o (\nabla v_o\cdot N)\cdot h_o \ud s.
\end{align}
The above expression holds even for perturbations that are not
permissible (do not satisfy the constraint). One can decompose $h_i$
and $h_o$ on $\partial R$ into its normal and tangential components:
\begin{align}
  h_i &= \pi_N (h_i) + \pi_T (h_i)  \\
  h_o &= \pi_N (h_o) + \pi_T (h_o)
\end{align}
where $\pi_N (h) = (h\cdot N)N$ and $\pi_T(h) = h-\pi_N(h)$ for $h$
defined on $\partial R$. Note that $\pi_N(h_i)=\pi_N(h_o)$ by the
normal constraint, and thus, we will set $h^N = \pi_N(h_i)\cdot N =
\pi_N(h_o)\cdot N$. One can similarly decompose $\nabla v_i \cdot N$
and $\nabla v_o \cdot N$ into normal and tangential components on
$\partial R$. Therefore,
\begin{align}
  (\nabla v_i\cdot N) \cdot h_i &= h^N (\nabla v_i\cdot N)\cdot N + 
  \pi_T(h_i)\cdot \pi_T(\nabla v_i\cdot N) \\
  (\nabla v_o\cdot N) \cdot h_o &= h^N (\nabla v_o\cdot N)\cdot N + 
  \pi_T(h_o)\cdot \pi_T(\nabla v_o\cdot N).
\end{align}
Substituting these formulas into the variation \eqref{eq:variation_E} yields
\begin{align*}
  dE(v_i,v_o)\cdot (h_i,h_o) &= 
  \int_R 
  \left[ (J_1- I + \nabla I \cdot v_i) \nabla I - \alpha_i\Delta v_i \right] \cdot h_i \ud x \\
  &+ 
  \int_{\Omega\backslash R} 
  \left[ (J_1- I + \nabla I \cdot v_o) \nabla I - \alpha_o\Delta v_o \right] \cdot h_o \ud x
  \\
  &+
  \int_{\partial R} h^N 
  \left[ 
    \alpha_i (\nabla v_i\cdot N)\cdot N - 
    \alpha_o (\nabla v_o\cdot N)\cdot N
  \right]
  \ud s \\
  &+\int_{\partial R} \alpha_i \pi_T(h_i)\cdot \pi_T(\nabla v_i\cdot N) \ud s
  - \int_{\partial R} \alpha_o \pi_T(h_o)\cdot \pi_T(\nabla v_o\cdot N)  \ud s.
\end{align*}
Since the $h^N$ and $\pi_T(h_i),\,\pi_T(h_o)$ may be chosen
independently and arbitrarily, the necessary conditions for an optimum
are
\begin{align}
  \label{eq:vin_hard_PDE}
  -\alpha_i \Delta v_i + \nabla I \nabla I^T v_i &= -(J_1-I) \nabla
  I, \quad \mbox{in } R
  \\
  \label{eq:vout_hard_PDE}
  - \alpha_o \Delta v_o + \nabla I \nabla I^T v_o &= -(J_1-I) \nabla
  I, \quad \mbox{in } \Omega\backslash R \\
  \label{eq:boundary_hard_first}
  \alpha_i (\nabla v_i\cdot N)\cdot N &= 
  \alpha_o (\nabla v_o\cdot N)\cdot N, \quad \mbox{on } \partial R \\
  \nabla v_i\cdot N &= \pi_N( \nabla v_i\cdot N ), 
  \quad \mbox{on } \partial R \\
  \nabla v_o\cdot N &= \pi_N( \nabla v_o\cdot N ), 
  \quad \mbox{on } \partial R \\
  \label{eq:boundary_hard_last}
  v_i\cdot N &= v_o\cdot N, \quad \mbox{on } \partial R
\end{align}
The above PDE is uniquely specified, and thus the solution specifies a
global optimum. The boundary conditions indicate that the normal
derivatives of $v_i, v_o$ only have normal components, and the normal
components of the normal derivative of $v_i$ and $v_o$ differ by a
scalar factor $\alpha_o/\alpha_i$. Note that, like the case of
enforcing the normal continuity constraint via the soft penalty, $v_i$
and $v_o$ are related to by the boundary conditions, which enforce the
normal continuity constraint exactly while regularizing only within
regions $R$ and $\Omega\backslash R$ separately.

\subsection{Numerical Solution for Infinitesimal Deformation Estimation}
\label{subsec:numerics_velocity}

The operators on the left-hand side of \eqref{eq:vin_hard_PDE} and
\eqref{eq:vout_hard_PDE} (and similarly \eqref{eq:vin_soft_PDE} and
\eqref{eq:vout_soft_PDE} in the previous subsection) that act on $v_i$
and $v_o$ are (with the given boundary conditions) positive
semi-definite, and thus, one may use the conjugate gradient algorithm
for a fast numerical solution. This property is verified in
Appendix~\ref{app:positive_definiteness}.

Since the boundary conditions are not standard of the PDE used in the
medical imaging community, we now show one possible scheme for the
numerical discretization of the PDE in the previous sections. We apply
a finite difference discretization (although higher accuracy may be
obtained with a finite element method). Consider a pixelized regular
grid, we apply the standard finite difference approximation for the
Laplacian:
\begin{align}
  \Delta v_i(x) &= \sum_{y\sim x, y\in R} (v_i(y)-v_i(x)) + 
  \sum_{y\sim x, y\in \Omega\backslash R} (v_i(y)-v_i(x)) \\
  \Delta v_o(x) &= \sum_{y\sim x, y\in \Omega\backslash R} (v_o(y)-v_o(x)) + 
  \sum_{y\sim x, y\in R} (v_o(y)-v_o(x)),
\end{align}
where $y\sim x$ indicates that $y$ is a four-neighbor of $x$. Note
that $v_i(y)$ is not defined for $y\in \Omega\backslash R$ (also,
$v_o(y)$ is not defined for $y\in R$). We now derive extrapolation
formulas for these quantities by discretizing the boundary conditions.

We consider discretization of
\eqref{eq:boundary_hard_first}-\eqref{eq:boundary_hard_last}. Let
$x\in R$ and $y\in \Omega\backslash R$.  Applying a one-sided first
order difference to approximate $\nabla v_i(x) \cdot N$ and $\nabla
v_o(y) \cdot N$, we find
\begin{align*}
  \alpha_i ( v_i(y) - v_i(x) ) + 
  \beta \pi_{N} ( v_i(x) ) - \beta \pi_{N} ( v_o(x) )
  &= 0 \\
  \alpha_o ( v_o(y) - v_o(x) ) + 
  \beta \pi_{N} ( v_i(y) ) - \beta \pi_{N} ( v_o(y) )  
  &= 0.
\end{align*}
where $N$, the outward normal can be approximated simply by the unit
vector pointing from $x$ to $y$, or more accurately with a level set
representation of the region $R$, the gradient of the level set
function. We employ the latter approximation in determining $\pi_N$.

Solving for $v_i(y)$ and $v_o(x)$ in terms of $v_o(y)$ and $v_i(x)$, one obtains
\begin{align*}
  v_i(y) &= v_i(x) + \frac{\beta(\beta-\alpha_o)}{\beta^2-\alpha_i\alpha_o} 
  \pi_N( v_o(y)-v_i(x) ) \\
  v_o(x) &= v_o(y) - \frac{\beta(\beta-\alpha_i)}{\beta^2-\alpha_i\alpha_o} 
  \pi_N ( v_o(y) - v_i(x) ).
\end{align*}
Let $v=(v_i,v_o)$ be the velocity on $\Omega$ and $A_s$ be the
operator on the left hand side of \eqref{eq:vin_soft_PDE} and
\eqref{eq:vout_soft_PDE}, then the discretization is
\begin{multline}
  A_s v(x) = \\
  \begin{cases}
    -\alpha_i \sum_{y\sim x, y\in R} (v(y)-v(x)) + 
    \frac{\alpha_i \beta(\beta-\alpha_o)}{\beta^2-\alpha_i\alpha_o}
    \sum_{y\sim x, y\in R^c}  \pi_N ( v(y)-v(x) ) + 
    \nabla_i I(x) \nabla_i I(x)^Tv(x) & x\in R \\
    -\alpha_o \sum_{y\sim x, y\in R^c} (v(y)-v(x)) + 
    \frac{\alpha_o \beta(\beta-\alpha_i)}{\beta^2-\alpha_i\alpha_o}
    \sum_{y\sim x, y\in R^c}  \pi_N ( v(y)-v(x) ) + 
    \nabla_o I(x) \nabla_o I(x)^Tv(x) & x\in R^c
  \end{cases},
\end{multline}
where $\nabla_i$ and $\nabla_o$ are gradient operators approximated
with central differences for interior points of $R$ and $R^c =
\Omega\backslash R$, respectively, and one-sided differences are
applied at the boundary points so that differences do not cross the
boundary. One then solves the system below using the conjugate
gradient algorithm:
\begin{equation}
  A_s v(x) = -(J_1(x)-I(x)) 
  \begin{cases}
    \nabla_i I(x) & x\in R\\
    \nabla_o I(x) & x\in R^c
  \end{cases}.
\end{equation}

Next, by similar methodology, one can discretize the PDE
\eqref{eq:vin_hard_PDE}-\eqref{eq:boundary_hard_last}. The
discretization of the boundary conditions are
\begin{align*}
  \alpha_i ( v_i(y)-v_i(x) ) \cdot N &= \alpha_o ( v_o(y)-v_o(x) ) \cdot N\\
  v_i(y)-v_i(x) &= \pi_N ( v_i(y)-v_i(x) ) \\
  v_o(y)-v_o(x) &= \pi_N ( v_o(y)-v_o(x) )\\
  v_i(y)\cdot N  &= v_o(x)\cdot N
\end{align*}
for $y\in R^c$ and $x\in R$. Since $v_i(y)$ and $v_o(x)$ are not
defined, we derive extrapolation formulas for these quantities by
solving the above system for $v_i(y)$ and $v_o(x)$ in terms of
$v_i(x)$ and $v_o(y)$; this yields
\begin{align}
  v_i(y) &= v_i(x) + \frac{\alpha_o}{\alpha_i+\alpha_o} \pi_N( v_o(y) -v_i(x) ) \\
  v_o(x) &= v_o(y) - \frac{\alpha_i}{\alpha_i+\alpha_o} \pi_N( v_o(y) -v_i(x) ).
\end{align}
The discretization of the operators on the left hand side of
\eqref{eq:vin_hard_PDE} and \eqref{eq:vout_hard_PDE}, which we denote
$A_h$ is then
\begin{multline}
  A_h v(x) = \\
  \begin{cases}
    -\alpha_i \sum_{y\sim x, y\in R} (v(y)-v(x)) + 
    \frac{\alpha_i\alpha_o}{\alpha_i+\alpha_o}
    \sum_{y\sim x, y\in R^c}  \pi_N ( v(y)-v(x) ) + 
    \nabla_i I(x) \nabla_i I(x)^Tv(x) & x\in R \\
    -\alpha_o \sum_{y\sim x, y\in R^c} (v(y)-v(x)) + 
    \frac{\alpha_i\alpha_o}{\alpha_i+\alpha_o}
    \sum_{y\sim x, y\in R^c}  \pi_N ( v(y)-v(x) ) + 
    \nabla_o I(x) \nabla_o I(x)^Tv(x) & x\in R^c
  \end{cases},
\end{multline}
and then the solution for $v=(v_i,v_o)$ is obtained by solving
\begin{equation}
  A_h v(x) = -(J_1(x)-I(x)) 
  \begin{cases}
    \nabla_i I(x) & x\in R\\
    \nabla_o I(x) & x\in R^c
  \end{cases}
\end{equation}
using conjugate gradient.

We note that the numerical solution for our method, of within region
regularization along with the normal constraint, has similar
computational cost as global regularization (with the traditional Horn
\& Schunck method). The operators $A_s$ and $A_h$ slightly differ from
the operator in global regularization. The simple modification as well
as fast computational cost make our method an easy and costless
alternative to traditional global regularization.

We note that both the hard and soft constraint solution lead to
boundary conditions that are not traditional, and both are solved
using a similar numerical scheme (in the next section). In the
experiments we use the hard-constraint formulation, and the soft
version is presented for completeness and to show that the soft
version would not lead to any easier formulation or numerical scheme.

\section{Larger Deformation Estimation and Shape Tracking}
In this section, we consider the case of non-infinitesimal
deformations between frames, which is typical in realistic MRI
sequences. We derive a simple technique for the registration the
satisfies the properties \eqref{eq:warp_vel} and
\eqref{eq:warp_decomp}.

This is accomplished by using the results of the previous section to
estimate an initial infinitesimal deformation $v_0$ between the two
given images $J_0$ and $J_1$, then $J_0$ and the region $R$ are
deformed infinitesimally by $v_0$, then the process is repeated on the
deformed region and deformed image until convergence. The accumulated
warp $w$ can be computed easily, but in tracking, the deformed region
$w(R)$ is of primary interest.

We now put the simple scheme mentioned above into a PDE formulation.
This formulation estimates $R_{\tau}$, $\phi_{\tau}^{-1}$, and
$w^{-1}$ defined in \eqref{eq:warp_vel} and \eqref{eq:warp_decomp}. To
do this, one solves for the incremental deformation $v_{\tau}$ by one
of the methods presented in the previous section, the image $J_0$ is
warped by the accumulated warp $\phi_{\tau}^{-1}$, and the procedure
is repeated, but this time solving for the velocity to deform
$J_0\circ \phi_{\tau}^{-1}$. This procedure is summarized below:
\begin{align}
  \label{eq:vel_solve}
  v_{\tau} &= \operatorname*{arg\,min}_v E(v; R_{\tau},  I_{\tau})
  \\
  \label{eq:warp_evolution}
  \partial_{\tau} \phi_{\tau}^{-1} &= -\nabla \phi_{\tau}^{-1} \cdot
  v_{\tau}, \, \phi_0^{-1}(x) = x\\
  \label{eq:region_evolution}
  \partial_{\tau} \Psi_{\tau} &= -\nabla \Psi_{\tau} \cdot
  v_{\tau}, \, \Psi_0(x) = \mbox{dist}_R(x), \, 
  R_{\tau} = \{ \Psi_{\tau} \leq 0 \} \\
  \label{eq:appearance_evolution}
  I_{\tau} &= J_0 \circ \phi_{\tau}^{-1}
\end{align}
where $\mbox{dist}_R$ is the signed distance function of $R$,
$R_{\tau}=\phi_{\tau}(R)$ is the region formed by flowing $R$ along
the velocity field for time $\tau$, and $E$ is defined in
\eqref{eq:energy_lin}.  The solution of \eqref{eq:energy_lin} is
determined by one of the methods presented in
Section~\ref{sec:infinitesimal_deformation} (we use the hard
constraint formulation in the experiments).  The region $R_{\tau}$ is
represented by a level set function $\Psi_{\tau} : \Omega \to \R$,
which makes the computation of the region $R_{\tau}$ convenient and
have sub-pixel accuracy, although the level set is not required
($R_{\tau}$ may be directly computed from $R$ and
$\phi^{-1}_{\tau}$). The level set function satisfies a transport
equation shown in \eqref{eq:region_evolution}. The backward map
$\phi_{\tau}^{-1}$ satisfies a transport PDE: the identity map is
transported along integral curves of $v_{\tau}$ to determine
$\phi_{\tau}^{-1}$. While in tracking, only $R_{\tau}$ is desired, the
backward map is computed to aid in accurate numerical computation of
$I_{\tau}$ \eqref{eq:appearance_evolution}, which is required to
estimate $v_{\tau}$ \eqref{eq:vel_solve}. At the time of convergence
of the region $R_{\tau}$, $T$, $I_T$ approximates $J_1$, and the
registration between $J_1$ to $J_0$ is $w^{-1} = \phi_T^{-1}$.

Note that integrating a sufficiently smooth vector field $v_{\tau}$ to
form $\phi_{\tau}$ as in \eqref{eq:warp_evolution} (and
\eqref{eq:warp_vel}) guarantees that $\phi_{\tau}$ is a diffeomorphism
within $R$ and $\Omega\backslash R$ (see classical results in
\cite{ebin1970groups}). Note that since $v_{\tau}$ in each region
$R_{\tau}$ and $\Omega \backslash R_{\tau}$ are solutions of Poisson
equations, $v_{\tau}$ is differentiable with Sobolev $H^2$ regularity
in each region, and so sufficiently smooth.

\subsection{Tracking Multiple Regions}
\label{subsec:multiple_regions}

\begin{figure}
\centering
\includegraphics[trim=0 75 0 375,clip,totalheight=1.8in]{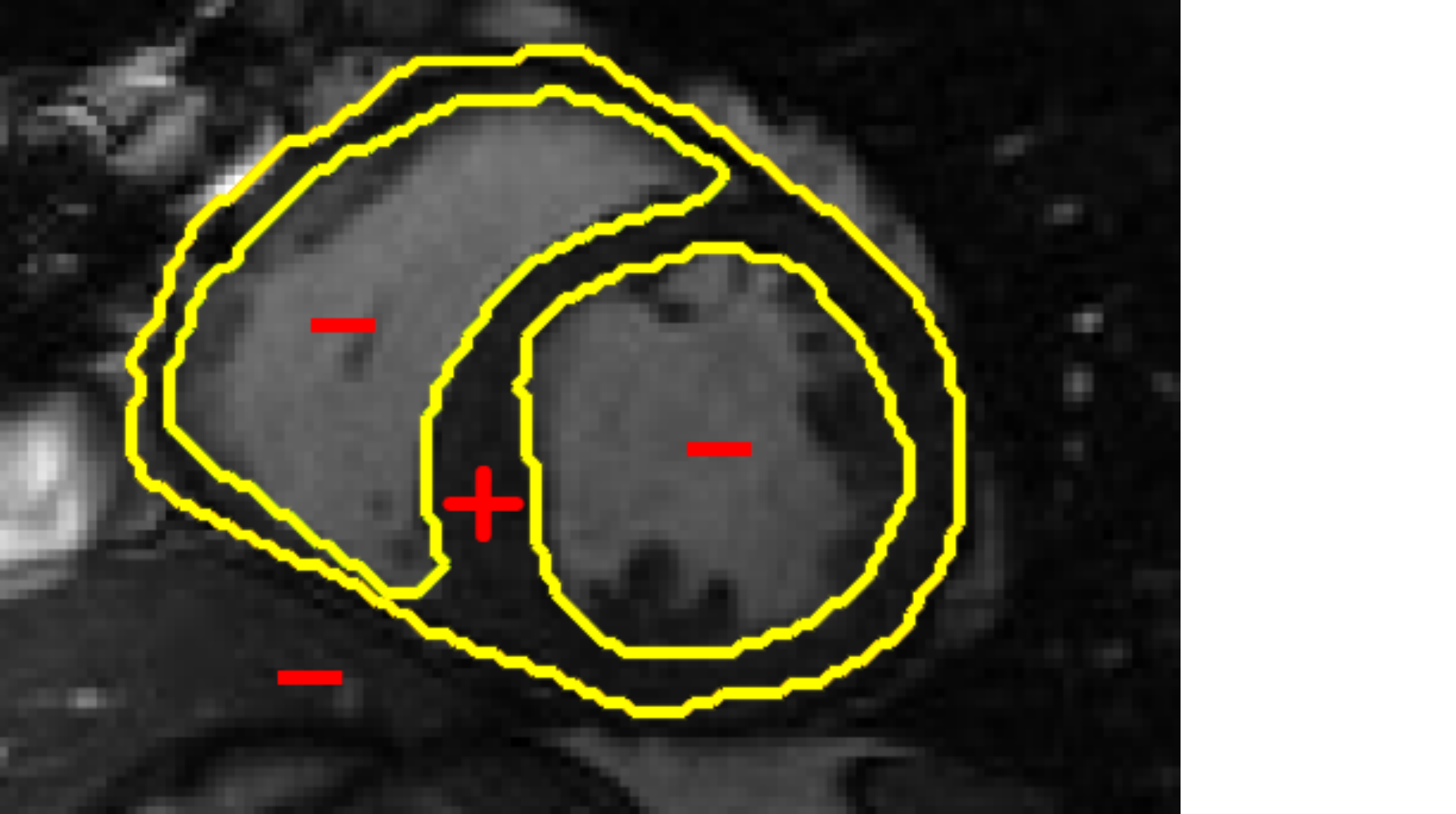}
\caption{Representation of the right and left ventricles and
  myocardium with a single level set function (sign of level set
  function indicated in red in the regions bounded by yellow
  contours).}
\label{fig:heart_levelset_rep}
\end{figure}

In cardiac image analysis, there are multiple structures (the right
and left ventricles, and myocardium) that all useful and should be
segmented. Our method is easily adaptable to this case. Indeed,
computation of $v_{\tau}$ in Section~\ref{subsec:numerics_velocity}
can be readily generalized.  In general, multiple level sets should be
used to represent multiple regions.  However, in our case of interest
(ventricles and surrounding epicardium), the regions form a rather
simple topology (see Figure~\ref{fig:heart_levelset_rep}), and all
regions can be represented using a single level set.

While theoretically $\phi_{\tau}$ for each $\tau$ will be an
invertible/onto map in each individual region, and thus regions cannot
change topology, numerically, between close by structures,
merging/splitting may occur. Since we know that, in our application of
interest, there is no such topology change, we enforce a hard topology
constraint, that topology must not change during the level set
evolution. This is now standard with level sets using discrete
topology preserving techniques \cite{han2003topology}. The original
level set evolution is augmented with a step that looks for
\emph{non-simple points} that change sign in a level set update, i.e.,
locations of topology change. Such points are not allowed to change
sign, and this preserves topology. Non-simple points are easily
detected with local pixel-wise operations, and this makes
implementation easy and the technique adds very little computational
cost. The reader is referred to \cite{han2003topology} for details on
the criteria for simple points.

\section{Experiments}
\label{sec:experiments}

This section consists of four sets of experiments. The first two
experiments are examples that illustrate and verify that our technique
works as expected. The third and fourth experiments are the core
experiments that show the main motivation of our algorithm: as a
technique to improve the prediction step in interactive segmentation
algorithms, which are predominantly used in commercial applications
for cardiac analysis. These last experiments thus compares our
technique to the popular and recent cardiac image segmentation
software Medviso \cite{heiberg2010design,sjogren2012semi}.

\comment{
The last
experiment shows the use of our algorithm on ventricles and epicardium
segmentation (multiple regions) all at once as described in
Section~\ref{subsec:multiple_regions}. 

The first
experiment is a synthetic experiment that illustrates our model and
its advantage over global regularization. The second experiment
compares segmentation of the left and right ventricles using both
within region regularization with the normal constraint with global
regularization. 
}

\subsection{Synthetic Experiment}

We start by verifying our registration and tracking technique on a
synthetic sequence designed to mimic the piecewise deformation with
matching normals for which our technique is designed. We consider a
sequence composed of images with two textures (one for the object, and
the other for the background). Textures are needed so that optical flow can be
determined.  The region of interest is the disc, and it along with the
background contracts. In addition to the contraction, there is small
rotational motions of the disc and the background in opposite
directions. The sequence is constructed so that the normal component
across the boundary matches in both regions. Note that this causes the
true flow to be non-smooth across the boundary. Ground truth
registration between consecutive images are known, and the deformation
is not infinitesimal.

\begin{figure}
  \centering
  {\small
    \begin{tabular}{c@{}c@{}c@{}c}
      image & next image & color code & ground truth \\
      \includegraphics[width=.125\linewidth,clip=true, trim=4 4 8
      8]{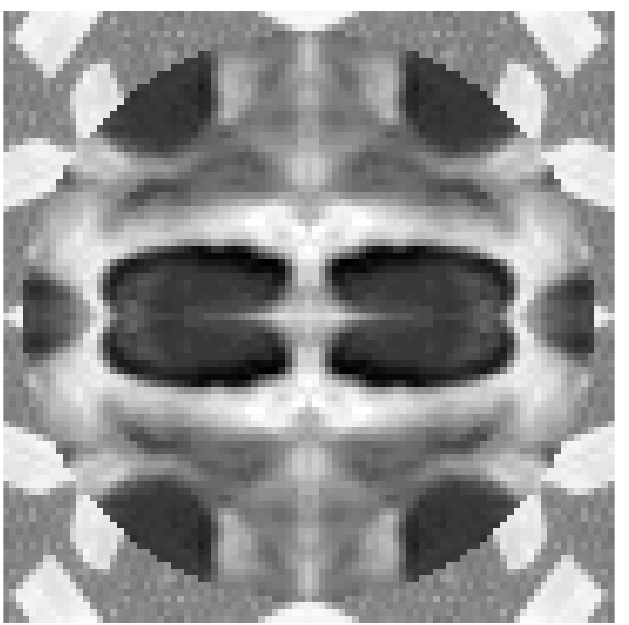} &
      \includegraphics[width=.125\linewidth,clip=true, trim=4 4 8
      8]{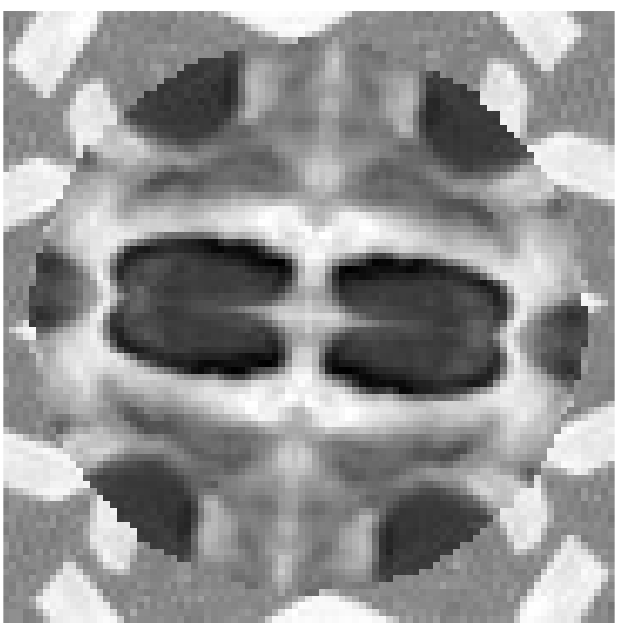} &
      \includegraphics[width=.125\linewidth,clip=true, trim=4 4 8
      8]{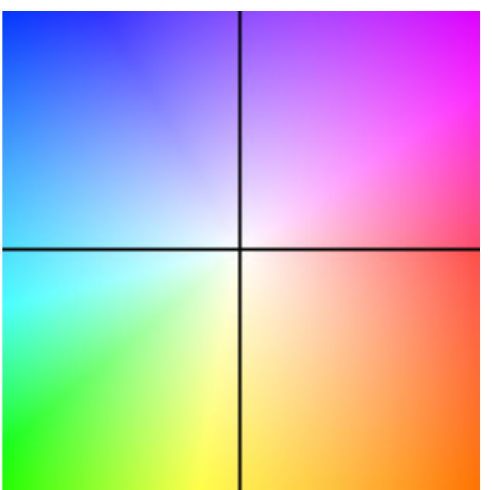} & 
      \includegraphics[width=.125\linewidth,clip=true, trim=4 4 8
      8]{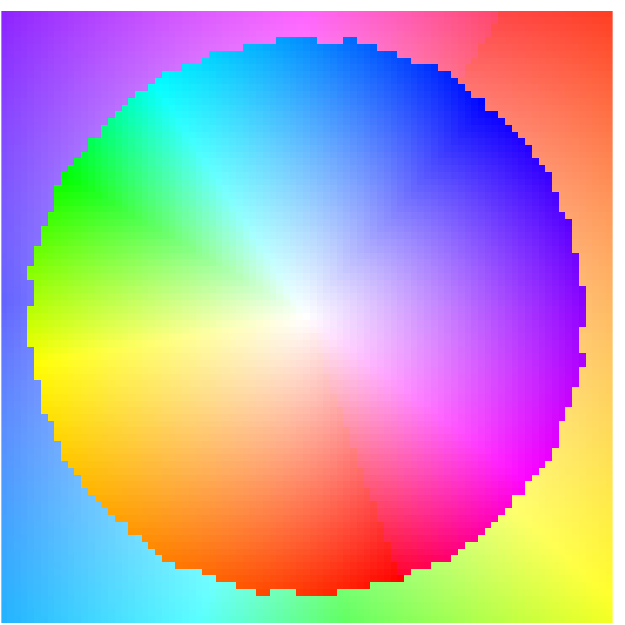} \\
      \includegraphics[width=.125\linewidth,clip=true, trim=4 4 8
      8]{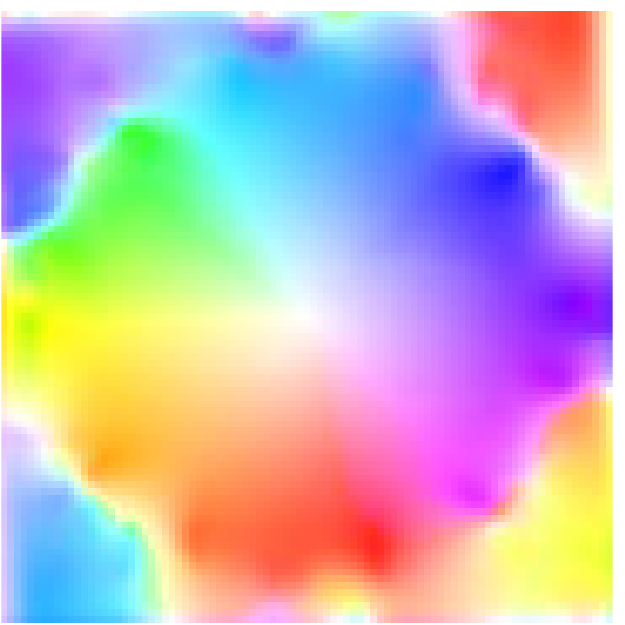} &
      \includegraphics[width=.125\linewidth,clip=true, trim=4 4 8 8]{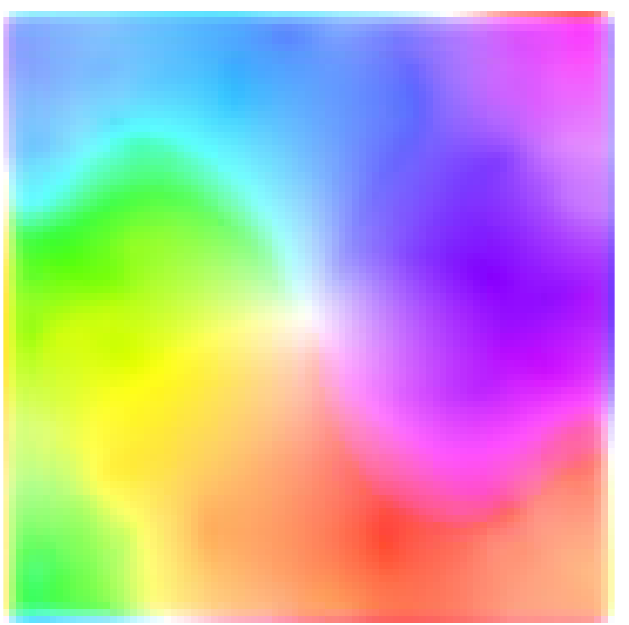} &
      \includegraphics[width=.125\linewidth,clip=true, trim=4 4 8 8]{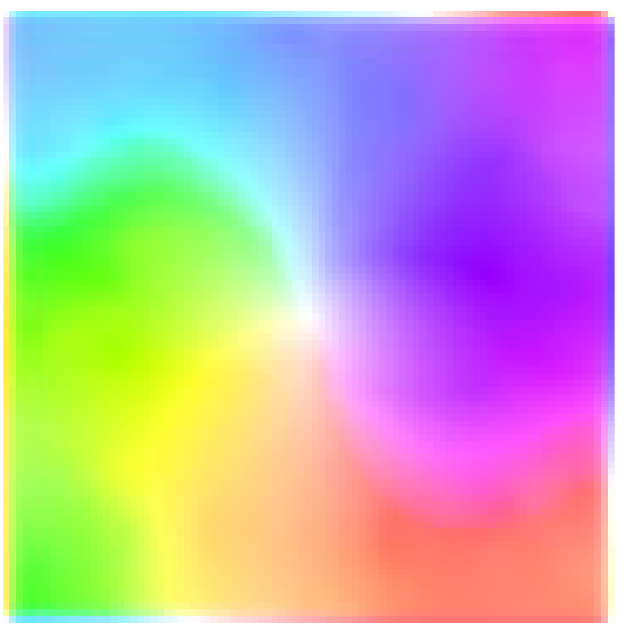} &
      \includegraphics[width=.125\linewidth,clip=true, trim=4 4 8
      8]{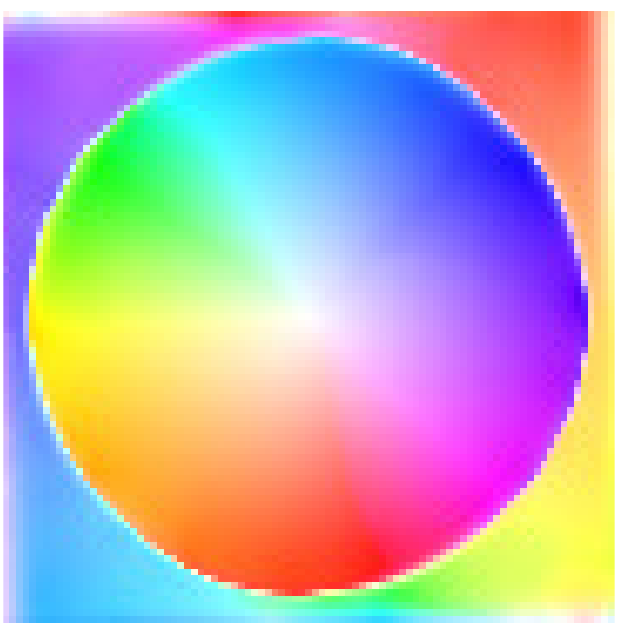}\\
      global, $\alpha=3$ & global, $\alpha=20$ & global, 
      $\alpha=50$ & our method \\
    \end{tabular}
  }
  \caption{{\bf Global vs. Physically Motivated Registration on
      Synthetic Images.} [Top]: First two frames of the sequence,
    optical flow color map, and ground truth registration between
    images (two regions of differing motion and matching normals on
    the boundary).  [Bottom]: Optical flow computed using global
    regularization $\alpha=\{3,20,50\}$, and by the proposed method
    with smoothness $\alpha_i=\alpha_o=3$. The proposed method
    recovers the true registration while no amount of global
    regularization recovers the true registration.}
  \label{fig:synthetic_a}
\end{figure}

The first row of Figure~\ref{fig:synthetic_a} shows the first two
frames of the synthetic sequence, the optical flow color code (whose
color indicates direction and intensity of color indicates magnitude)
and the ground truth optical flow for the first two frames. The second
row shows registrations computed by global regularization with
smoothness $\alpha_i=\alpha_o = \{3,20,50\}$ and the proposed method
with $\alpha_i=\alpha_o=3$. Notice that global regularization smooths
across the boundary, thus mixing inhomogeneous motions, leading to an
inaccurate registration. Our proposed registration, which does not
smooth across the boundary while satisfying the physical constraint,
is able to accurately recover the true registration.

\begin{figure}
\centering
{\small
\begin{tabular}{c@{}c@{}c@{}c}
  initial & \multicolumn{3}{c}{tracked (red-algorithm, yellow-ground truth)} \\
  \rotatebox{90}{global $\alpha\!=\!3$}
  \includegraphics[width=.12\linewidth,clip=true, trim=4 4 8
  8]{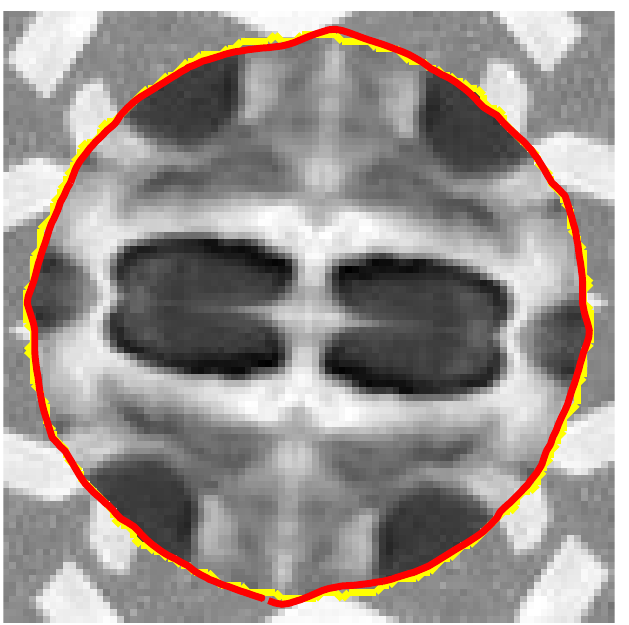} &
  \includegraphics[width=.12\linewidth,clip=true, trim=4 4 8
  8]{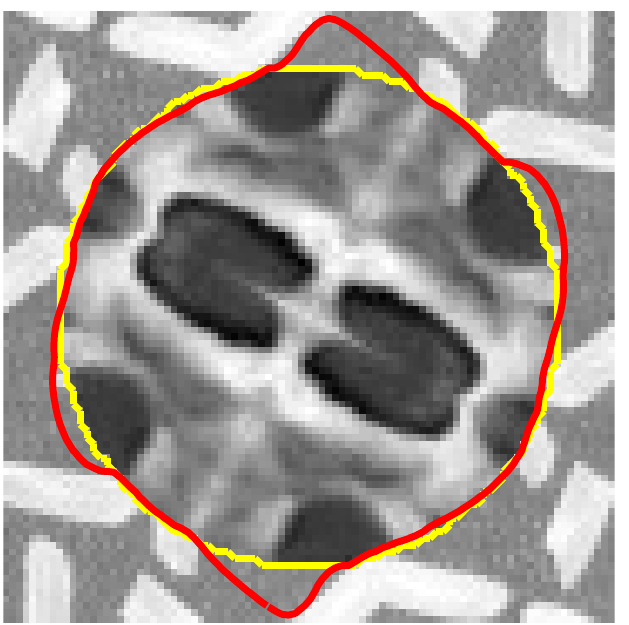} & 
  \includegraphics[width=.12\linewidth,clip=true, trim=4 4 8
  8]{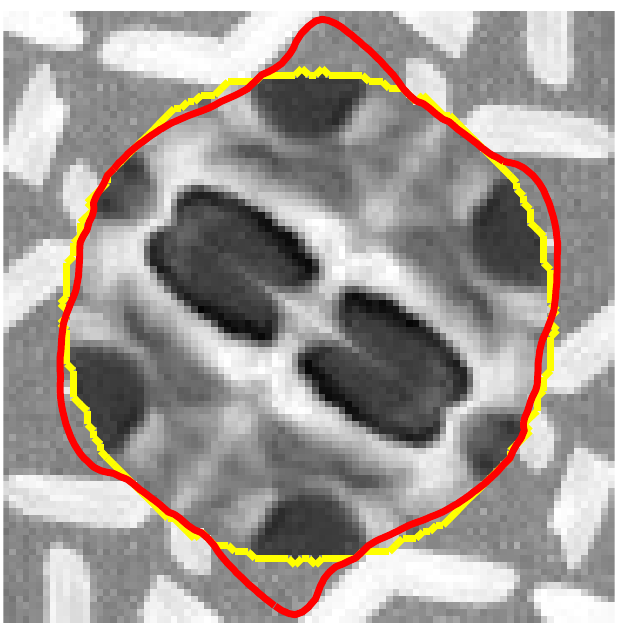} &
  \includegraphics[width=.12\linewidth,clip=true, trim=4 4 8
  8]{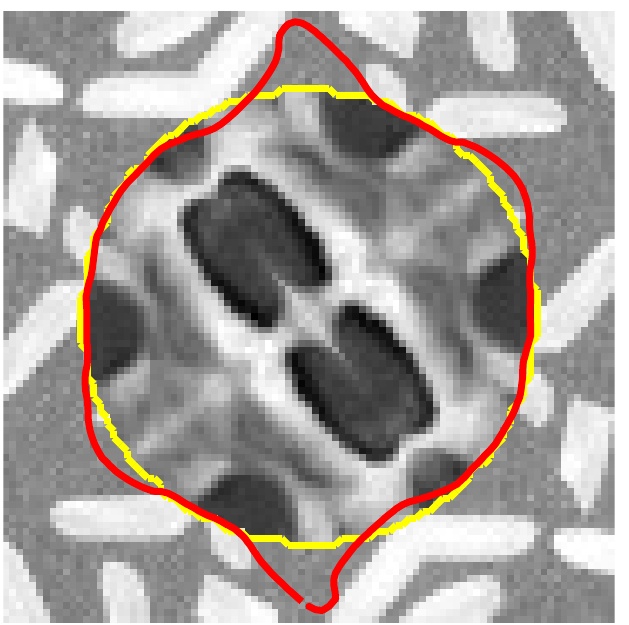}\\
  \rotatebox{90}{global $\alpha\!=\!20$}
  \includegraphics[width=.12\linewidth,clip=true, trim=4 4 8 8]{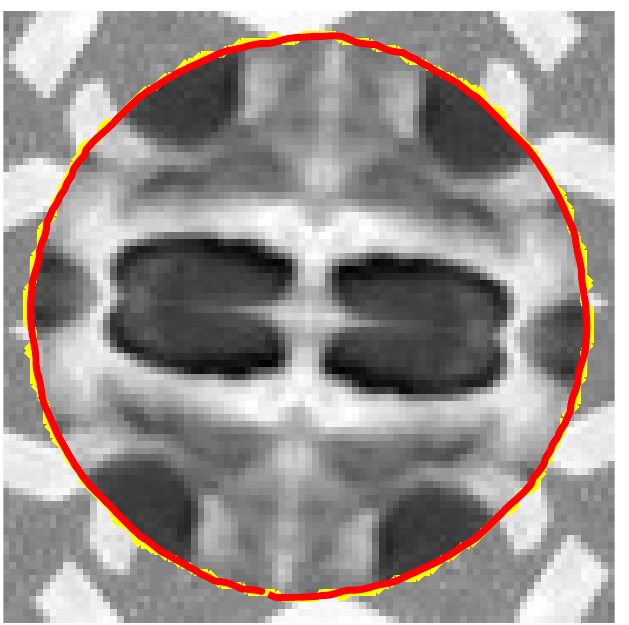} &
  \includegraphics[width=.12\linewidth,clip=true, trim=4 4 8
  8]{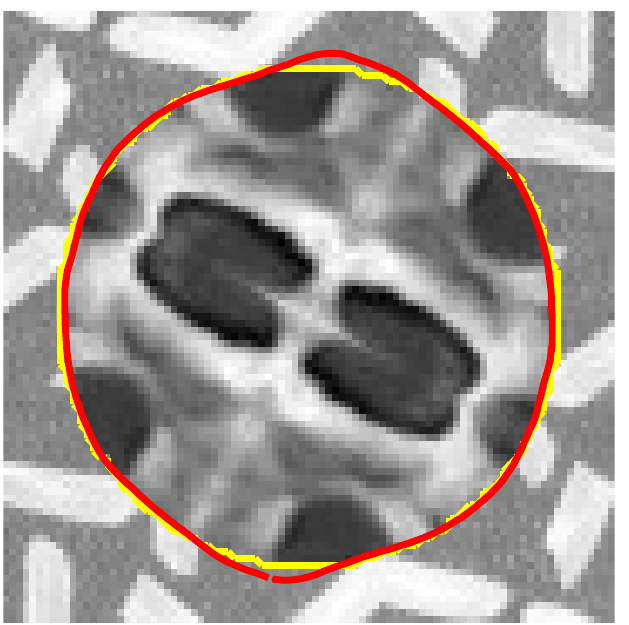} &
  \includegraphics[width=.12\linewidth,clip=true, trim=4 4 8
  8]{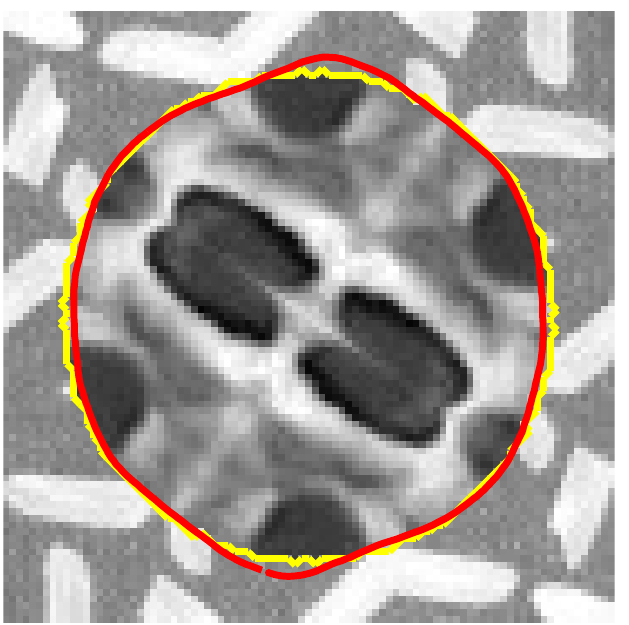} &
  \includegraphics[width=.12\linewidth,clip=true, trim=4 4 8
  8]{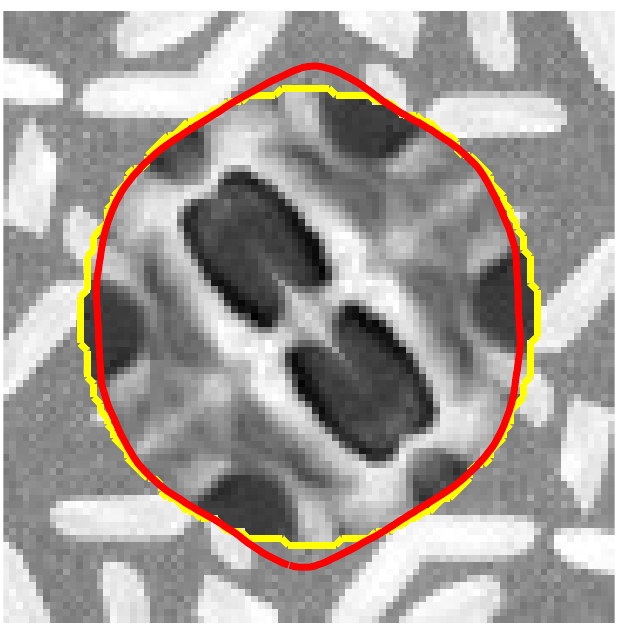}\\
  \rotatebox{90}{global $\alpha\!=\!50$}
  \includegraphics[width=.12\linewidth,clip=true, trim=4 4 8
  8]{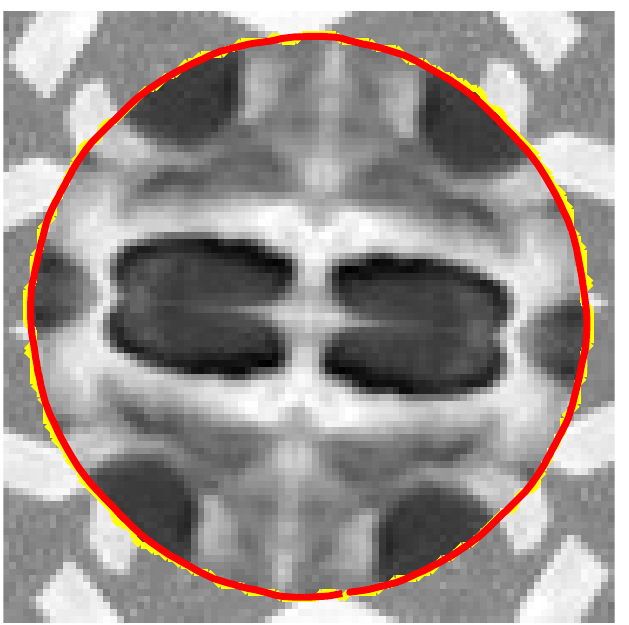} &
  \includegraphics[width=.12\linewidth,clip=true, trim=4 4 8
  8]{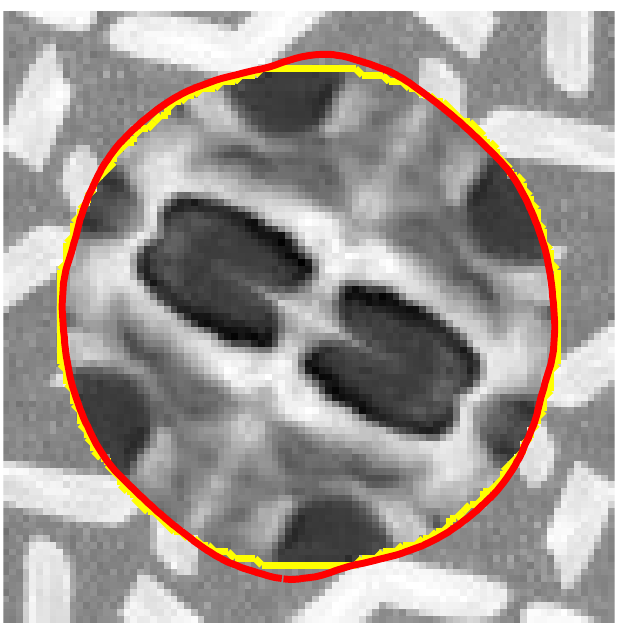} &
  \includegraphics[width=.12\linewidth,clip=true, trim=4 4 8
  8]{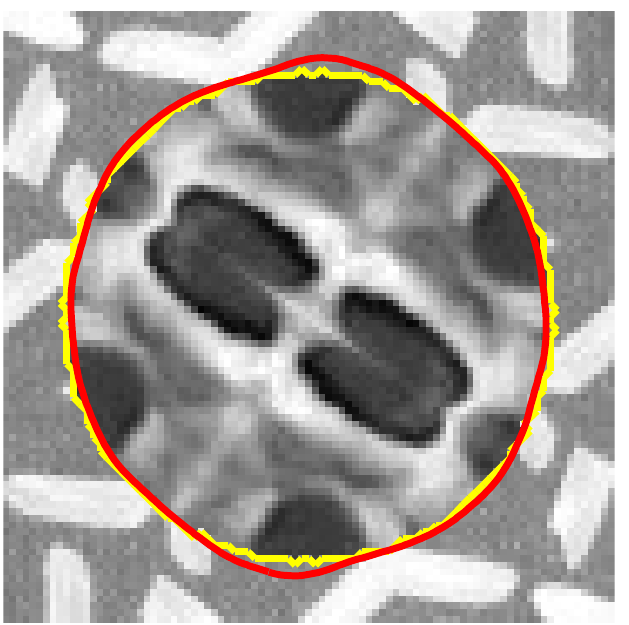} &
  \includegraphics[width=.12\linewidth,clip=true, trim=4 4 8
  8]{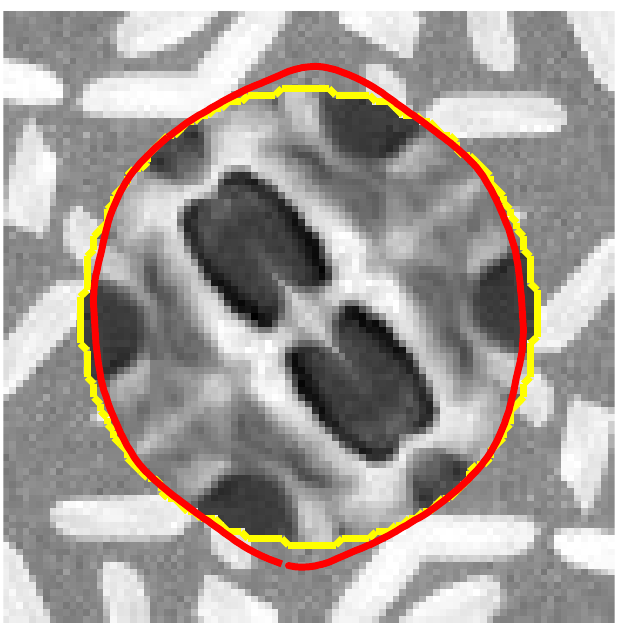}\\
  \rotatebox{90}{\,\,\,our method}
  \includegraphics[width=.12\linewidth,clip=true, trim=4 4 8
  8]{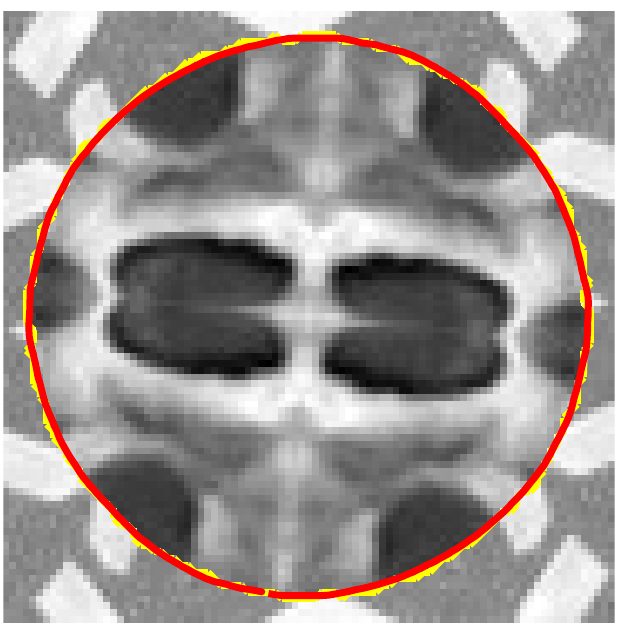} &
  \includegraphics[width=.12\linewidth,clip=true, trim=4 4 8
  8]{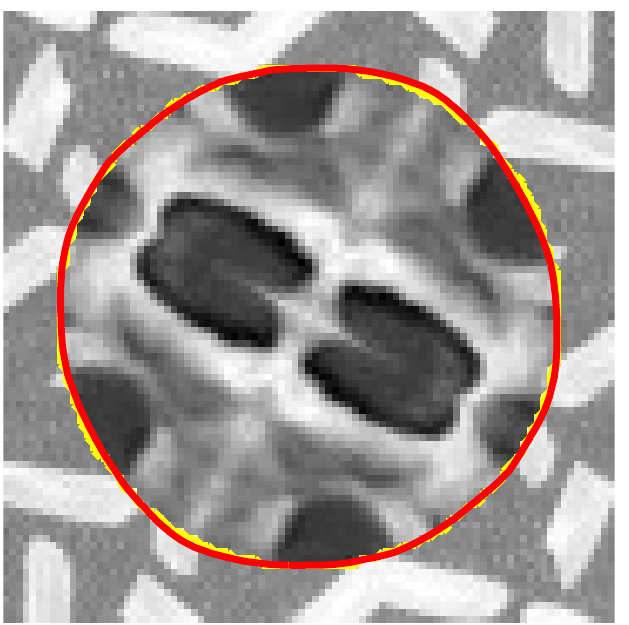} &
  \includegraphics[width=.12\linewidth,clip=true, trim=4 4 8
  8]{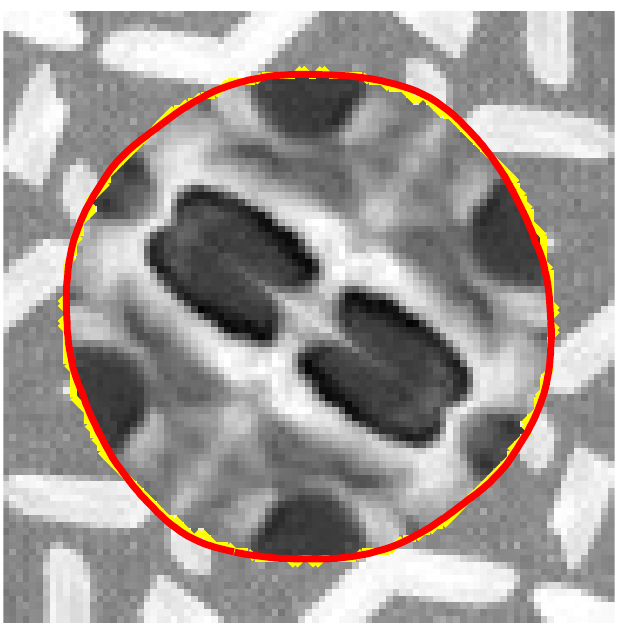} &
  \includegraphics[width=.12\linewidth,clip=true, trim=4 4 8
  8]{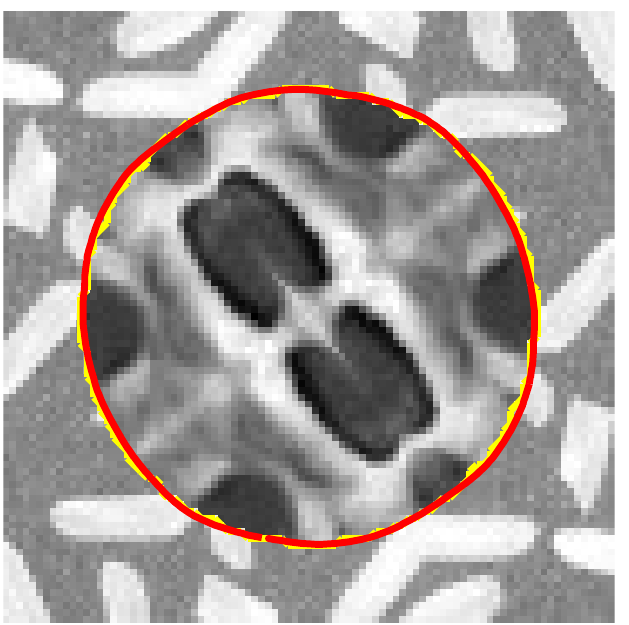}
\end{tabular}
}
\caption{{\bf Global vs. Physically Motivated Tracking on Synthetic
    Sequence}.  [First three rows]: Region tracking using global
  regularization with $\alpha = \{3,20,50\}$. [Last row]: proposed
  method. Since the proposed technique is able to better capture the
  underlying registration than global regularization, it also has
  higher accuracy in segmentation of the circular object.}
\label{fig:synthetic_b}
\end{figure}

Figure~\ref{fig:synthetic_b} displays the results of tracking the
whole synthetic sequence of 10 frames (only 4 are shown) with
registration that uses global regularization and our method. The first
three rows display the result of tracking using global regularization
with smoothness $\alpha_i=\alpha_o = \{3,20,50\}$. Notice that global
regularization of the deformation with small global regularization
($\alpha_i=\alpha_o=3$) leads to less smoothing across the boundary,
but an inaccurate segmentation due to small regularization which traps
the contour in small scale structures. Larger global regularization
smooths more across the boundary leading to an inaccurate
registration. The segmentation improves, but the boundary is still not
captured accurately. No amount of global regularization is able to
detect an accurate boundary. Finally, our method (last row), which
smooths within regions while simultaneously satisfying the normal
matching constraint is able to capture both an accurate registration
and segmentation.

\comment{
The region by
chance becomes more accurate in this example since increasing the
regularization averages the deformation from the background (which is
uniformly inward) and thus mathces the motion of the
boundary. However, even in this case, the results of the region
detected in global regularization are less accurate than that of the
proposed method (Figure~\ref{fig:synthetic1}).
}
\comment{
The next synthetic sequence (Figure~\ref{fig:synthetic2}) is similar
to the first, but in this case, the background also rotates in an
opposite direction as the inside. The whole image also contracts
inward, and the normal term of the inside and outside deformation
match. Notice that in this case, where the deformations inside and
outside highly differ, global regularization results in a highly
inaccurate deformation estimation (no matter the amount of regularity)
and a highly inaccruate region detection compared with the proposed
piecewise regularization.
}

\subsection{Ventricle Segmentation: Comparison of Three Registration
  Schemes}

\begin{figure}
  \centering
  {\small
    
  \begin{tabular}{c@{}c@{}c@{}c@{}c@{}c}
    initial & \multicolumn{4}{c}{ventricle tracked (red - algorithm
      result, yellow - ground truth)} \\
    \rotatebox{90}{\,\,region registration}
    \includegraphics[width=.19\linewidth,clip,trim=25 20 25
    50]{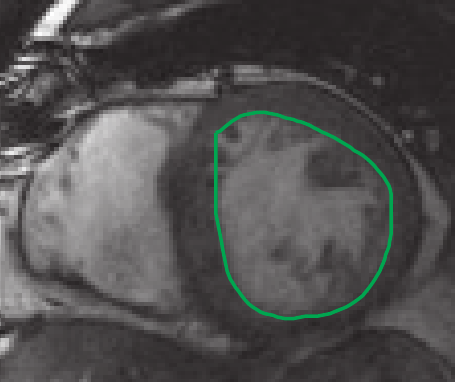} &
    \includegraphics[width=.19\linewidth,clip,trim=25 20 25
    50]{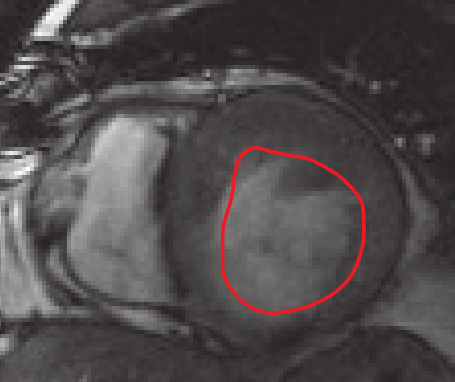} &
    \includegraphics[width=.19\linewidth,clip,trim=25 20 25
    50]{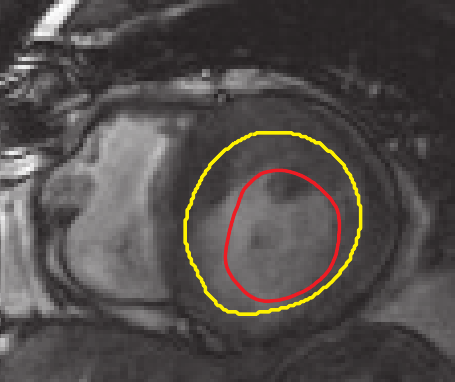} &
    \includegraphics[width=.19\linewidth,clip,trim=25 20 25
    50]{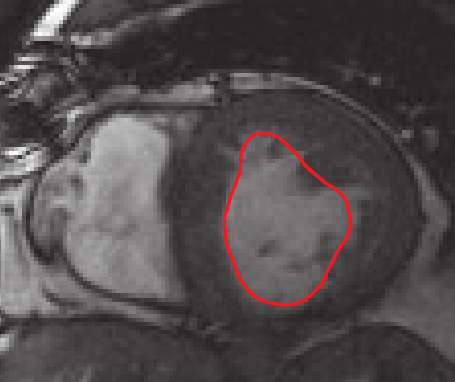} &
    \includegraphics[width=.19\linewidth,clip,trim=25 20 25
    50]{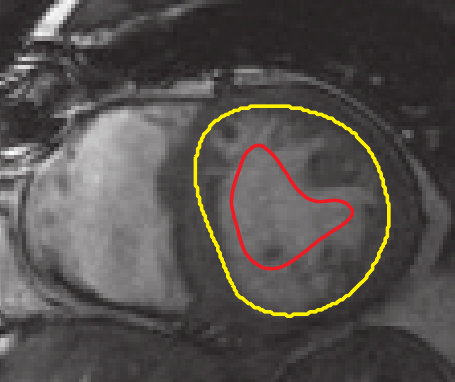}\\
    \rotatebox{90}{\,\,global smoothing}
    \includegraphics[width=.19\linewidth,clip,trim=25 20 25
    50]{compare_global/SC-HF-I-07-IM-0021-vb} &
    \includegraphics[width=.19\linewidth,clip,trim=25 20 25
    50]{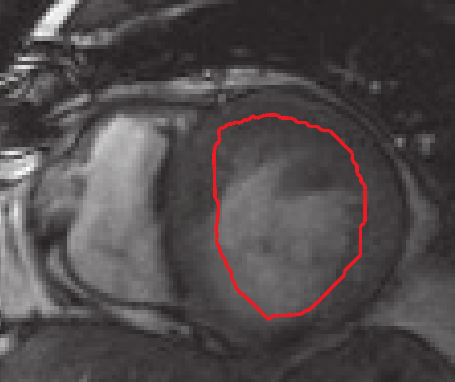} &
    \includegraphics[width=.19\linewidth,clip,trim=25 20 25
    50]{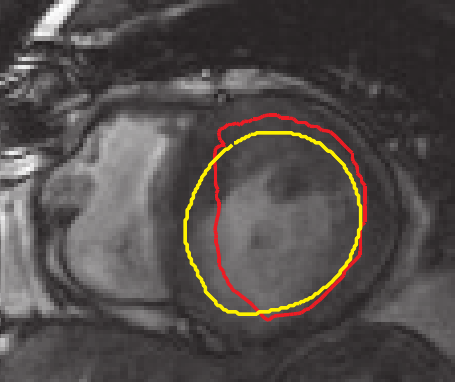} &
    \includegraphics[width=.19\linewidth,clip,trim=25 20 25
    50]{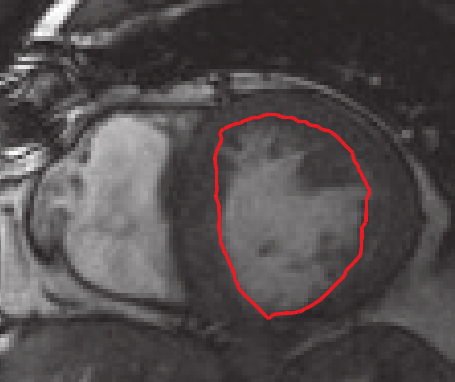} &
    \includegraphics[width=.19\linewidth,clip,trim=25 20 25 50]{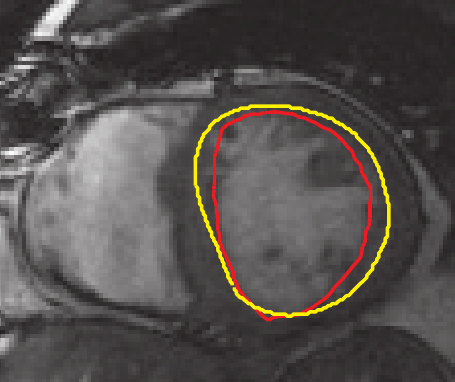}\\
    \rotatebox{90}{\quad\quad our method}
    \includegraphics[width=.19\linewidth,clip,trim=25 20 25
    50]{compare_global/SC-HF-I-07-IM-0021-vb} &
    \includegraphics[width=.19\linewidth,clip,trim=25 20 25
    50]{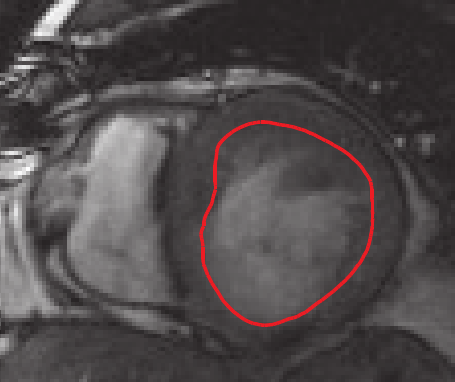} &
    \includegraphics[width=.19\linewidth,clip,trim=25 20 25
    50]{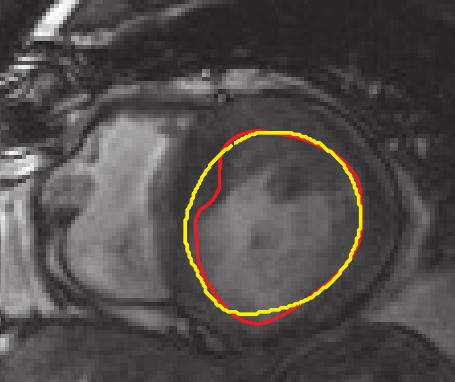} &
    \includegraphics[width=.19\linewidth,clip,trim=25 20 25
    50]{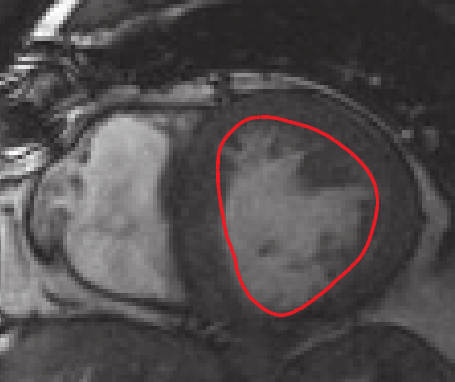} &
    \includegraphics[width=.19\linewidth,clip,trim=25 20 25 50]{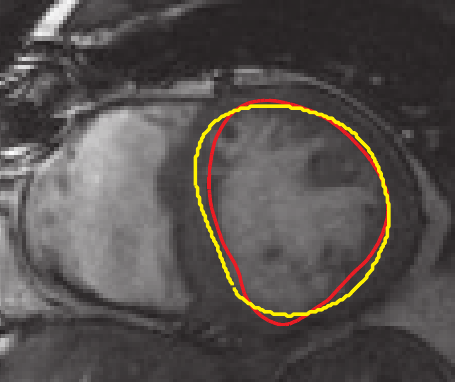}
  \end{tabular}
  }
  \caption{{\bf Illustrative Differences of Registration Schemes on the LV.}  [Top]:
    registering only the inside of the ventricle leads to inaccurate
    segmentation in subsequent frames. [Middle]: registering the
    whole image with global regularization smooths motion across
    different structures and leads to inaccurate
    segmentation. [Bottom]: registering the entire image with proposed
    technique leads to the most accurate segmentation.  Green:
    initialization, red: algorithm result, yellow: ground truth. }
  \label{fig:registrations_LV}
\end{figure}

\begin{figure}
  \centering
  {\small
  \begin{tabular}{c@{}c@{}c@{}c@{}c@{}c}
    initial & \multicolumn{4}{c}{ventricle tracked (red - algorithm
      result, yellow - ground truth)} \\
    \rotatebox{90}{\,\,region registration}
    \includegraphics[width=.19\linewidth,clip,trim=25 30 25
    40]{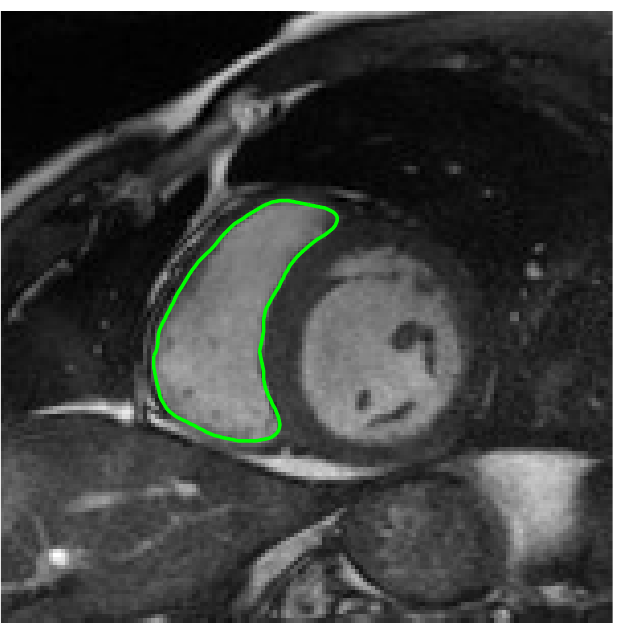} &
  \includegraphics[width=.19\linewidth,clip,trim=25 30 25
  40]{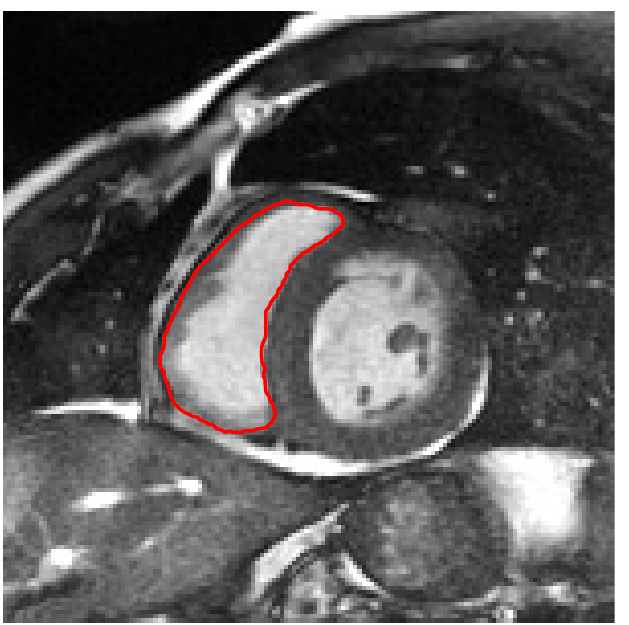} &
  \includegraphics[width=.19\linewidth,clip,trim=25 30 25
  40]{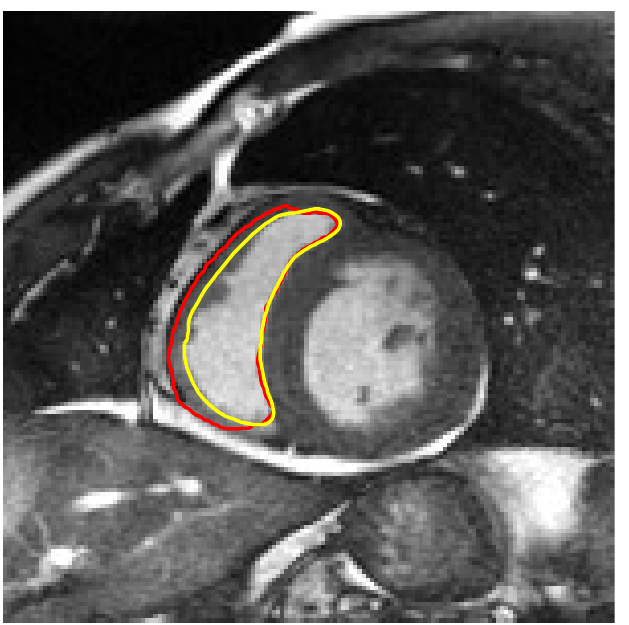} &
  \includegraphics[width=.19\linewidth,clip,trim=25 30 25
  40]{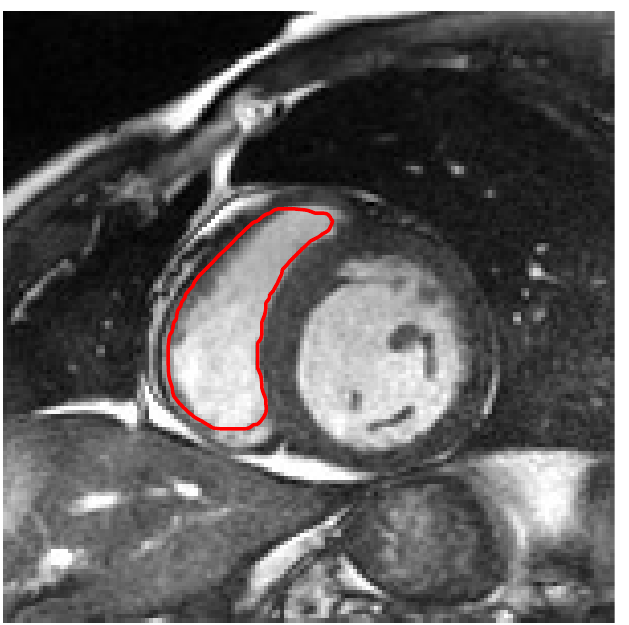} &
  \includegraphics[width=.19\linewidth,clip,trim=25 30 25
  40]{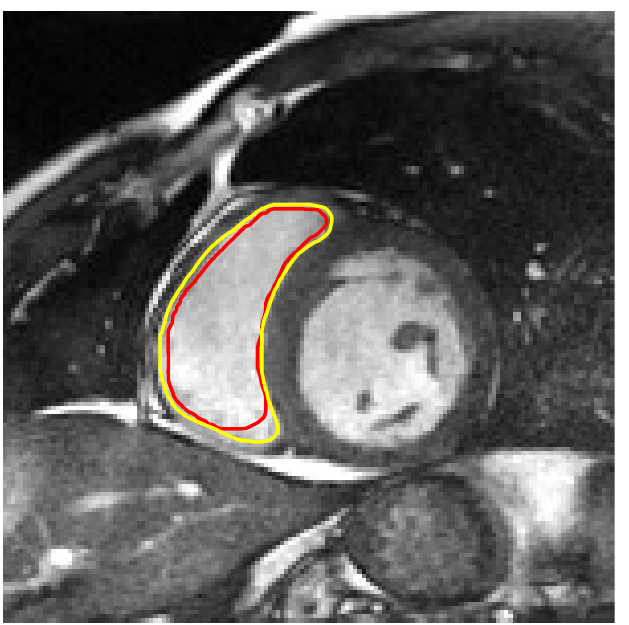}\\
    \rotatebox{90}{\,\,global smoothing}
    \includegraphics[width=.19\linewidth,clip,trim=25 30 25 40]{compare_global/SC-HF-I-06-IM-0041} &
    \includegraphics[width=.19\linewidth,clip,trim=25 30 25 40]{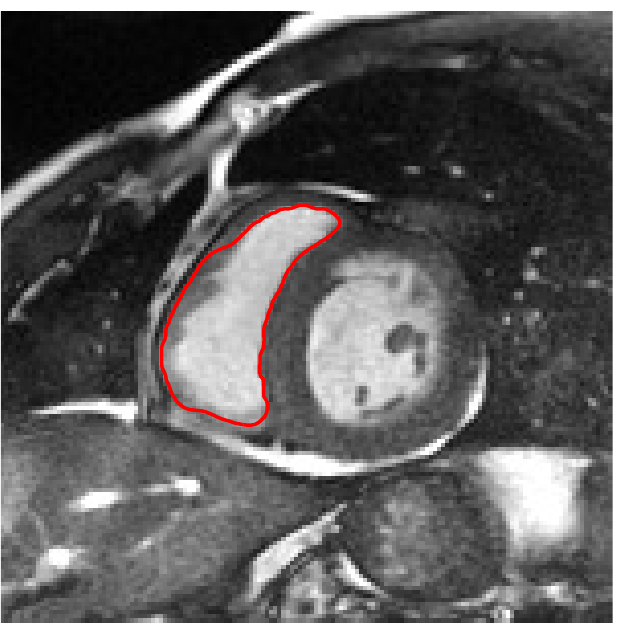} &
    \includegraphics[width=.19\linewidth,clip,trim=25 30 25 40]{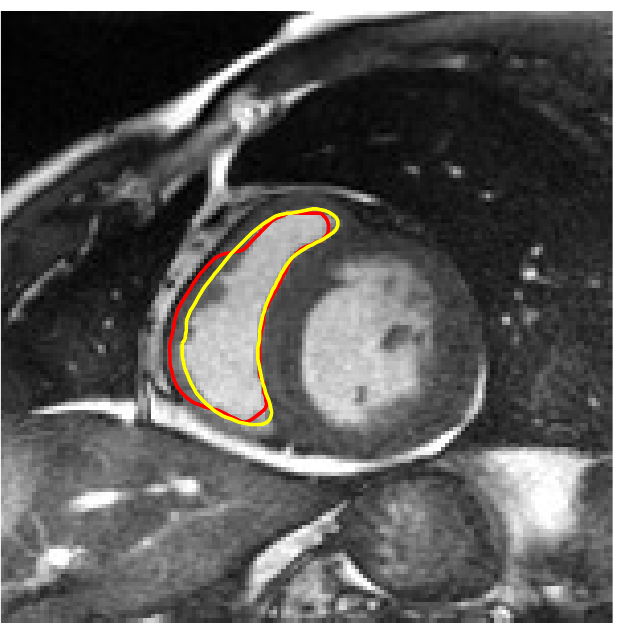} &
  \includegraphics[width=.19\linewidth,clip,trim=25 30 25
  40]{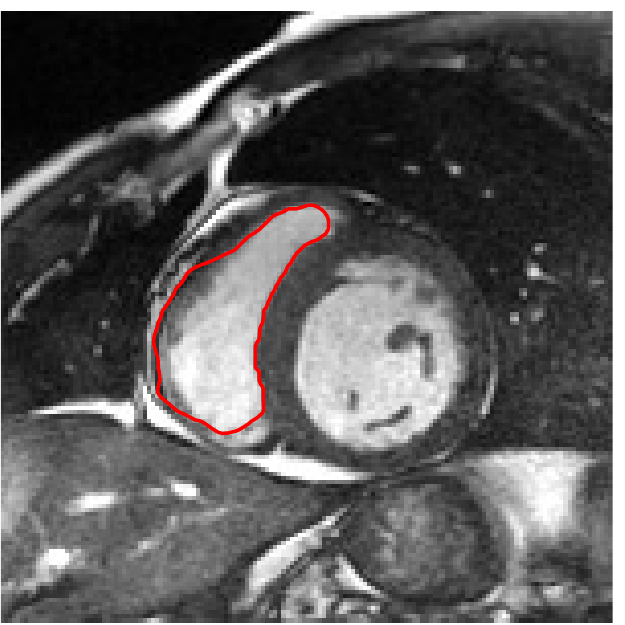} &
  \includegraphics[width=.19\linewidth,clip,trim=25 30 25
  40]{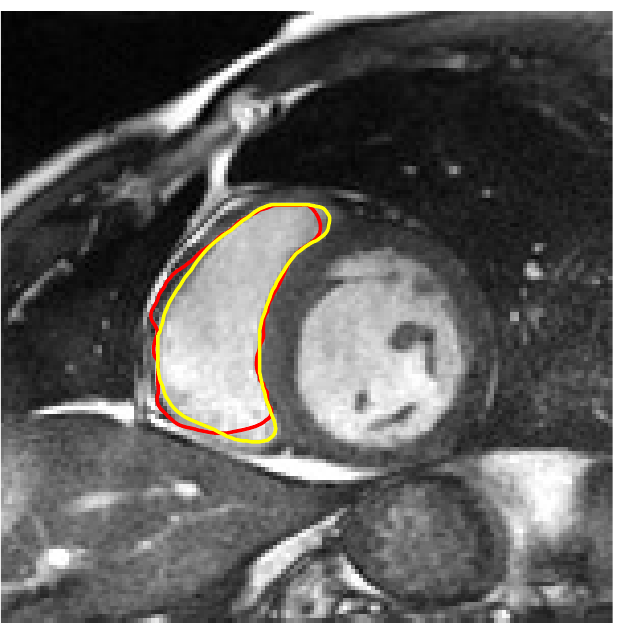}\\
      \rotatebox{90}{\quad\quad our method}
  \includegraphics[width=.19\linewidth,clip,trim=25 30 25
  40]{compare_global/SC-HF-I-06-IM-0041} &
  \includegraphics[width=.19\linewidth,clip,trim=25 30 25
  40]{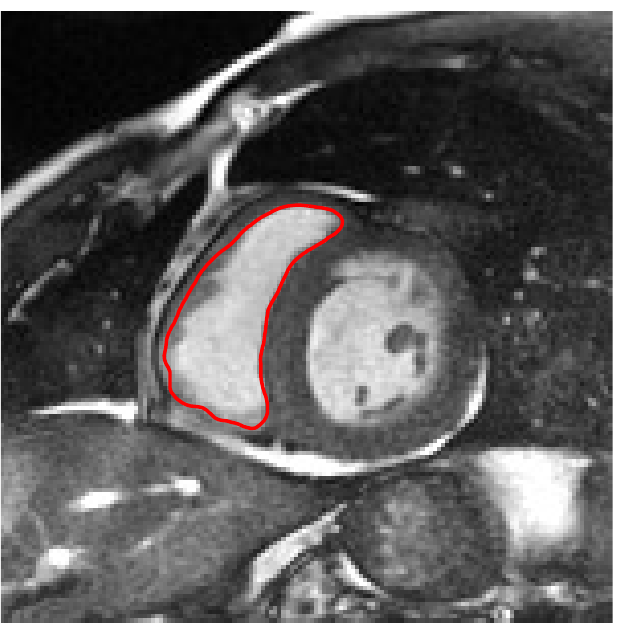} &
  \includegraphics[width=.19\linewidth,clip,trim=25 30 25
  40]{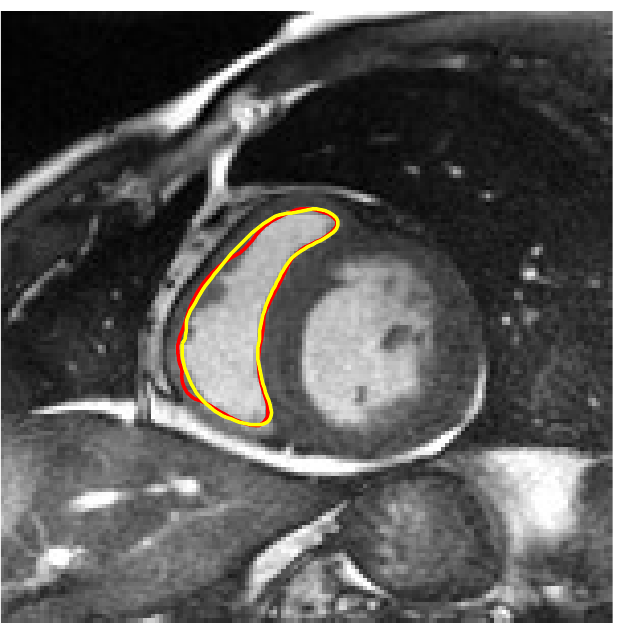} &
  \includegraphics[width=.19\linewidth,clip,trim=25 30 25
  40]{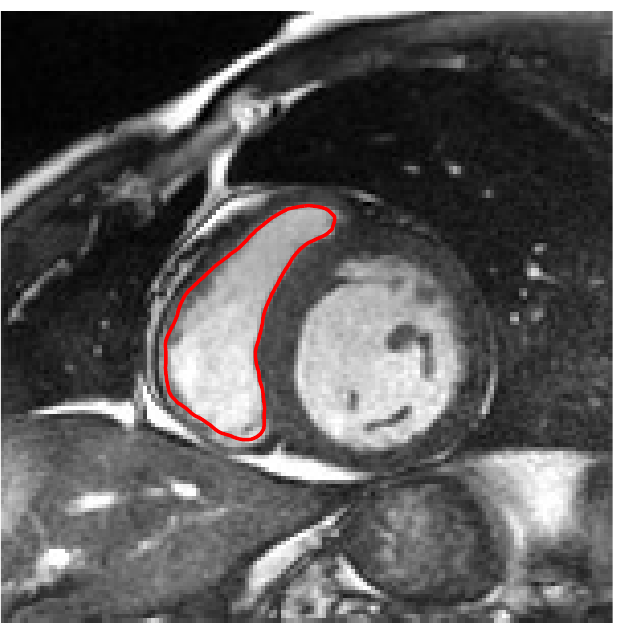} &
  \includegraphics[width=.19\linewidth,clip,trim=25 30 25 40]{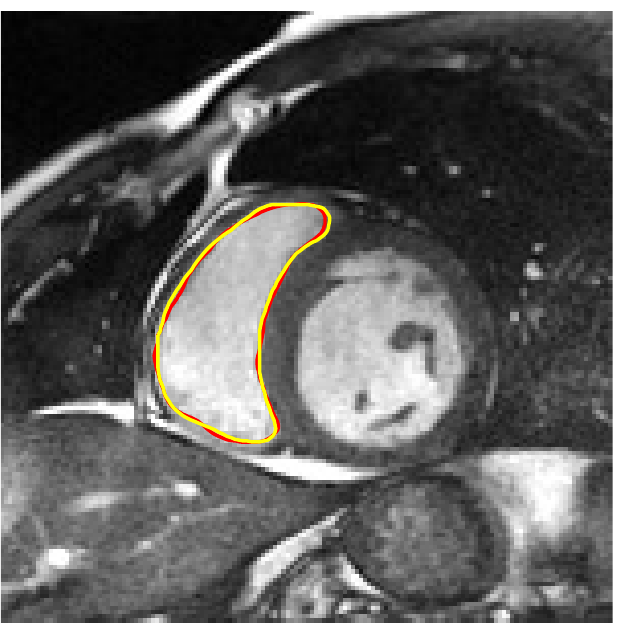}
\end{tabular}
}
\caption{ {\bf Illustrative Differences of Registration Schemes on the
    RV.}  [Top]: registering only the inside of the ventricle leads to
  inaccurate segmentation in subsequent frames. [Middle]: registering
  the whole image with global regularization smooths motion across
  different structures and leads to inaccurate segmentation. [Bottom]:
  registering the entire image with proposed technique leads to the
  most accurate segmentation.  Green: initialization, red: algorithm
  result, yellow: ground truth.}
\label{fig:registrations_RV}
\end{figure}

In this experiment, we focus on real cardiac MRI data and compare
registration methods used for segmentation of the LV and RV.  We
visually compare the tracking results given by our method to (M1)
registration of only the interior of current estimate of the ventricle
to a subset of next image (to show whole image registration is
needed), and to (M2) standard full image registration with global
smoothness. M1 is achieved by computing just the inside velocity with
Neumann boundary conditions on $\partial R_{\tau}$ (normal constraint
does not apply in M1). The best results with respect to ground truth
are chosen by choosing the optimal parameter $\alpha$ in all
methods. Results on LV and RV tracking for a full cardiac cycle are
given in Figure~\ref{fig:registrations_LV} for the LV and
Figure~\ref{fig:registrations_RV} for the RV.  Registering only the
organ (M1) results in errors (as the background registration is
helpful in restricting undesirable registrations of the foreground).
Globally smooth registration (M2) smooths motion from irrelevant
background structures into the ventricles, which results in drifting
from the desired boundary. Our method, which smooths within regions
with satisfying the physical constraint, is able to achieve the most
accurate results.

\comment{
We have also quantified the differences of our approach and global
regularization in segmentation on the MICCAI LV and RV Datasets. On
the LV dataset, the metric used is APD (average perpendicular
distance), and our approach has 
}

%/*******************************Experiments******************************************/

\subsection{LV and RV Segmentation: Quantitative Comparison to
  Commercial Software}

We show experiments demonstrating the main use of our algorithm: in
improving the prediction step of interactive segmentation methods. We
show that less interaction is needed with our approach than a recent
and widely used commercial cardiac segmentation software, Segment
from Medviso \cite{heiberg2010design,sjogren2012semi}.  We perform
quantitative assessment of the tracking performance of our method and
compare it to Medviso. The evaluation was carried out on publicly
available data sets, the MICCAI Left Ventricle Dataset \cite{miccai09}
and the MICCAI Right Ventricle Dataset \cite{miccai2012RV}.  The
validation dataset from \cite{miccai09} consists of 15 sets of cardiac
cine-MRI images. Each set contains 6 to 20 2D slices from a 3D image,
with each slice having 20 images of the cardiac phases. Similarly, the
data set \cite{miccai2012RV} contains 16 sets of cardiac cine-MRI
images, each containing about 10 slices of 20 phases each. These data
sets contain ground truth segmentations for left and right ventricles
respectively (unfortunately ground truth for both the LV and RV is not
available on a single dataset that we are aware of).  Both methods
start with the same initially correct segmentation, and subsequent
frames are segmented via propagation.  No manual interaction is used
as we wish to show that our method would require less interaction.
The regularity parameter $\alpha = \alpha_i = \alpha_o$ in our method
is found by choosing $\alpha$ so that the results are closest to
ground truth in a few training cases. The same parameter is then used
for all other cases.

\begin{figure}
  \centering
  {\small
  \begin{tabular}{c@{}c@{}c@{}c@{}c}
    initial & \multicolumn{4}{c}{ventricle tracked (red - algorithm
      result, yellow - ground truth)} \\
    \rotatebox{90}{\quad\quad Medviso}
    \includegraphics[width=.19\linewidth,clip=true, trim=30 40 20
    40]{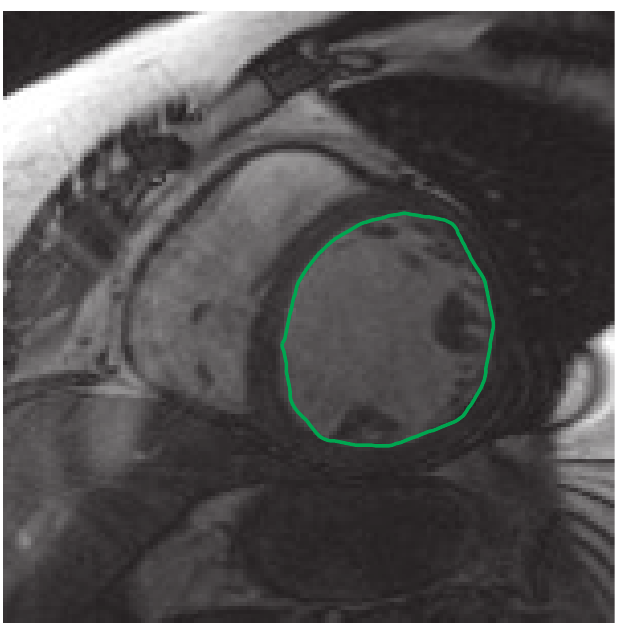} &
    \includegraphics[width=.19\linewidth,clip=true, trim=30 40 20
    40]{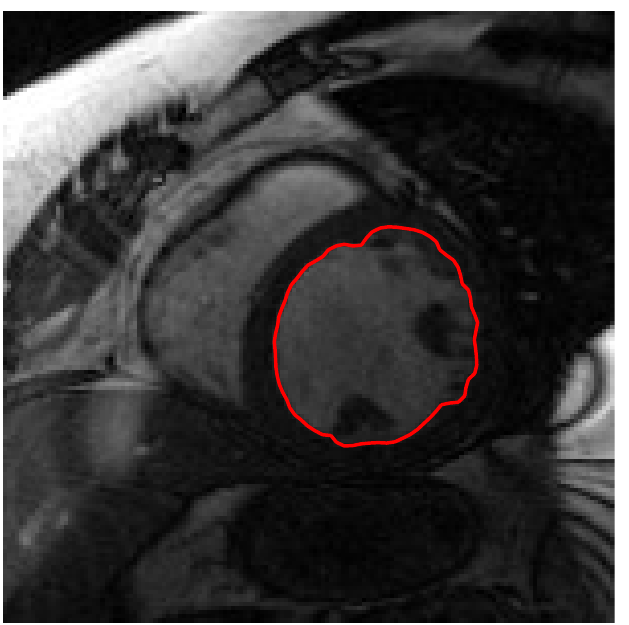} &
  \includegraphics[width=.19\linewidth,clip=true, trim=30 40 20
  40]{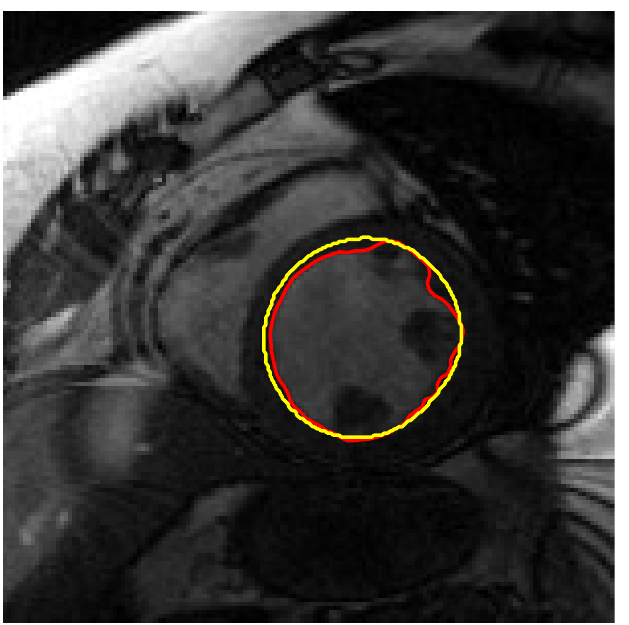} &
  \includegraphics[width=.19\linewidth,clip=true, trim=30 40 20
  40]{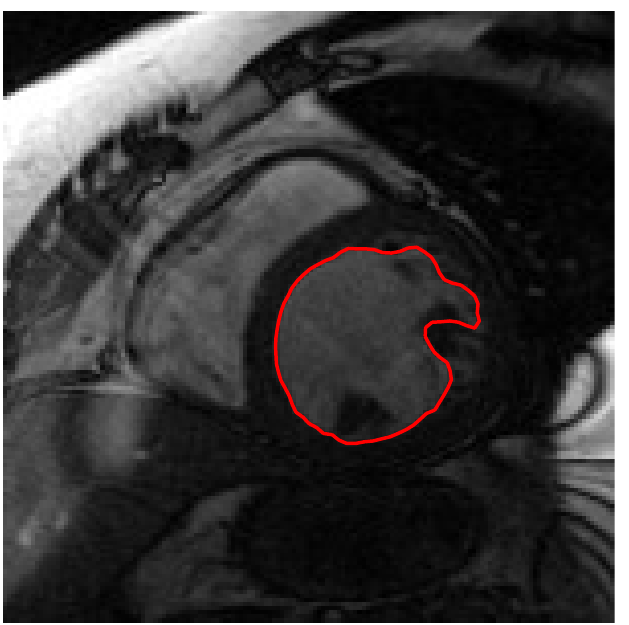} &
  \includegraphics[width=.19\linewidth,clip=true, trim=30 40 20
  40]{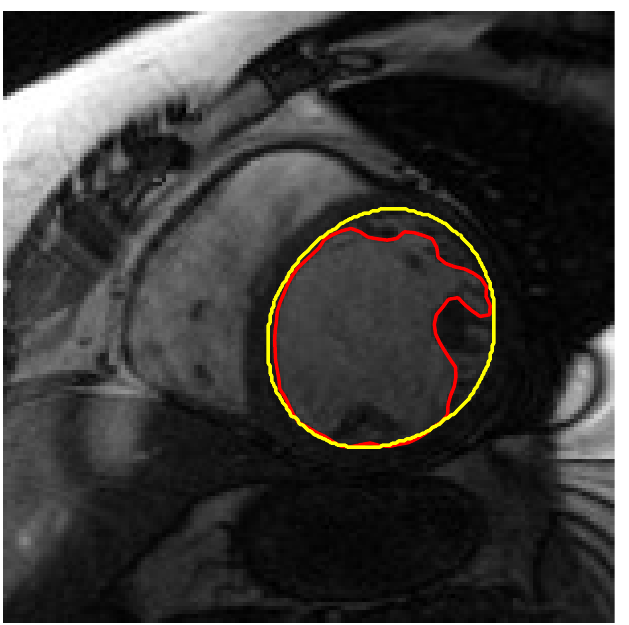} \\
  \rotatebox{90}{\quad\, our method}
  \includegraphics[width=.19\linewidth,clip=true, trim=30 40 20
  40]{compare_gt/SC-HF-NI-12-IM-0081} &
  \includegraphics[width=.19\linewidth,clip=true, trim=30 40 20
  40]{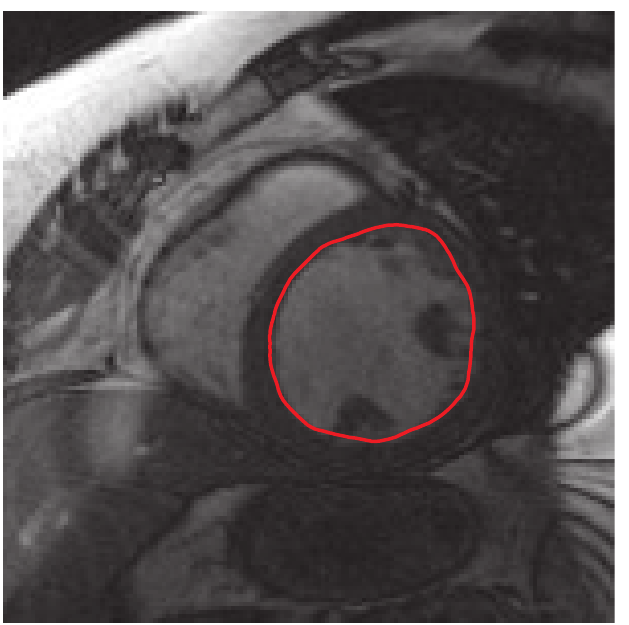} &
  \includegraphics[width=.19\linewidth,clip=true, trim=30 40 20
  40]{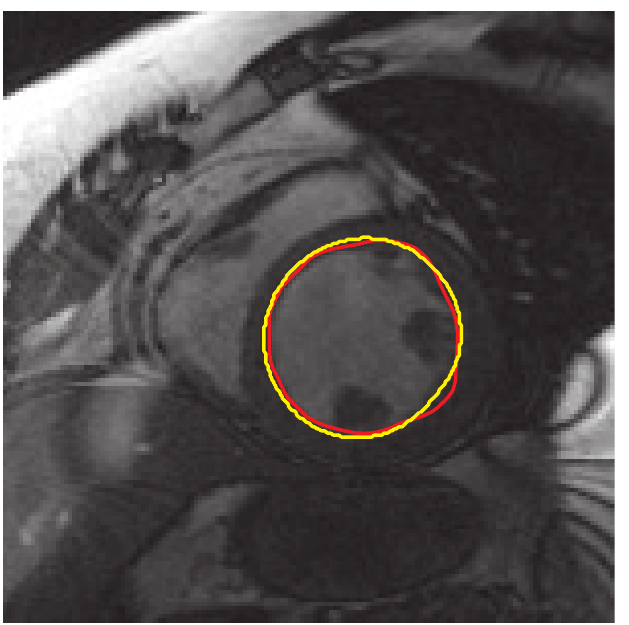} &
  \includegraphics[width=.19\linewidth,clip=true, trim=30 40 20
  40]{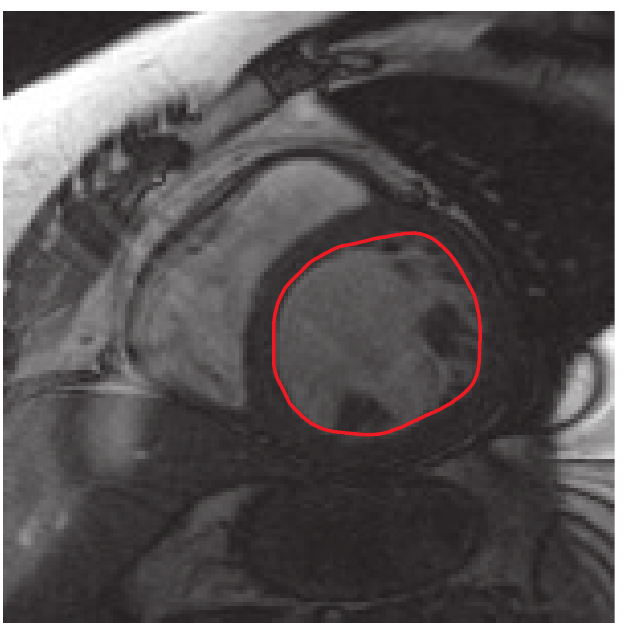} &
  \includegraphics[width=.19\linewidth,clip=true, trim=30 40 20
  40]{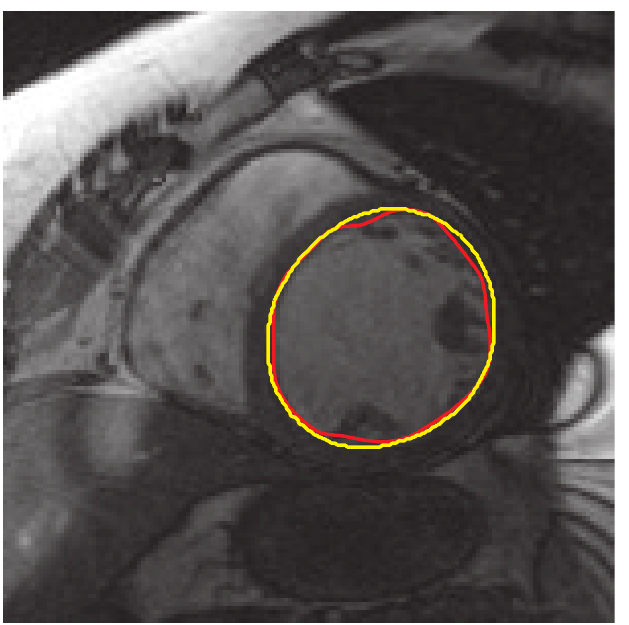} \\
  \rotatebox{90}{\quad\quad Medviso}
  \includegraphics[width=.19\linewidth,clip=true, trim=40 40 40
  40]{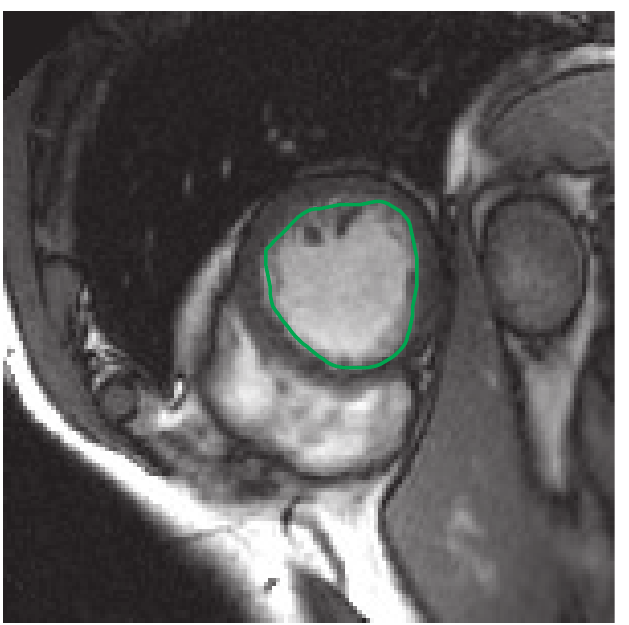} &
  \includegraphics[width=.19\linewidth,clip=true, trim=40 40 40
  40]{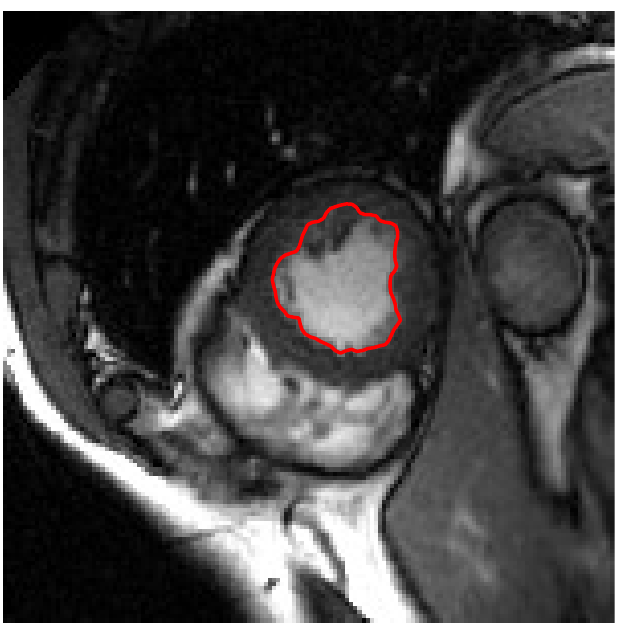} &
  \includegraphics[width=.19\linewidth,clip=true, trim=40 40 40
  40]{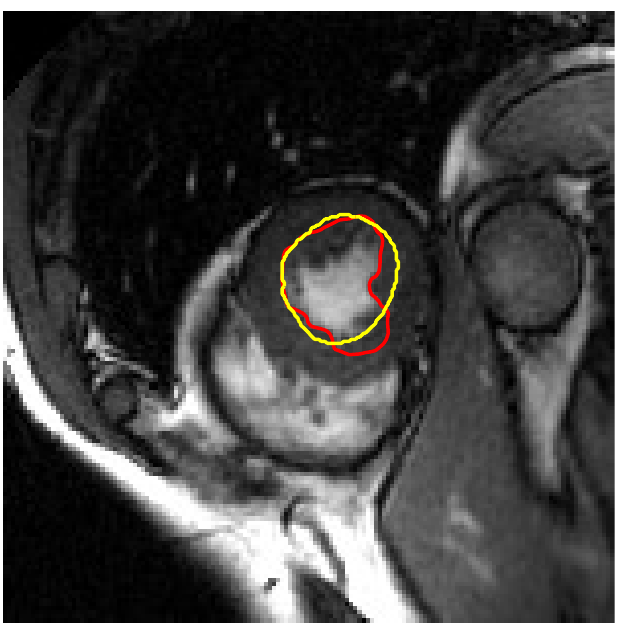} &
  \includegraphics[width=.19\linewidth,clip=true, trim=40 40 40
  40]{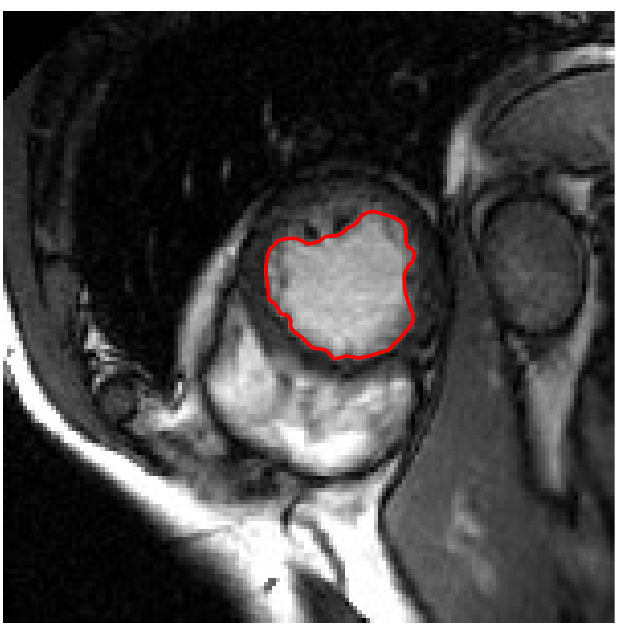} &
  \includegraphics[width=.19\linewidth,clip=true, trim=40 40 40
  40]{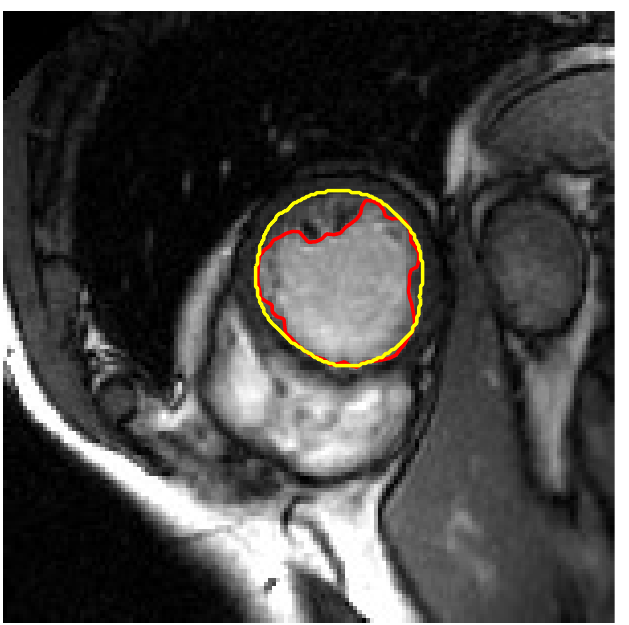} \\
  \rotatebox{90}{\quad\, our method}
  \includegraphics[width=.19\linewidth,clip=true, trim=40 40 40
  40]{compare_gt/SC-N-09-IM-0081} &
  \includegraphics[width=.19\linewidth,clip=true, trim=40 40 40
  40]{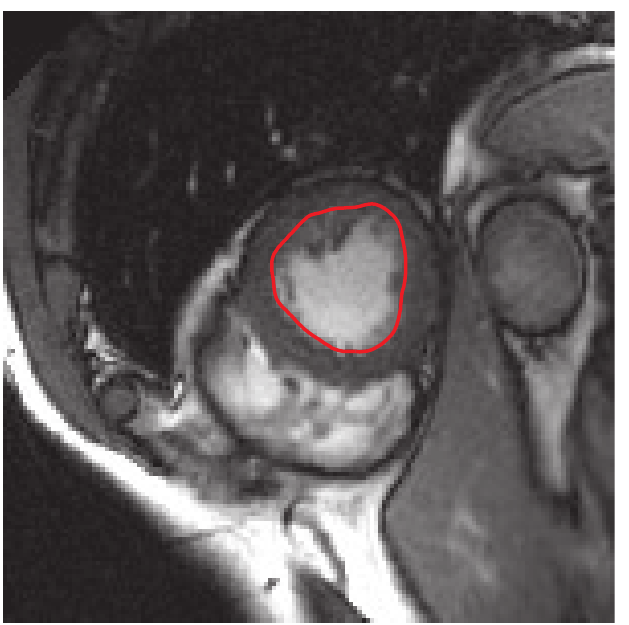} &
  \includegraphics[width=.19\linewidth,clip=true, trim=40 40 40
  40]{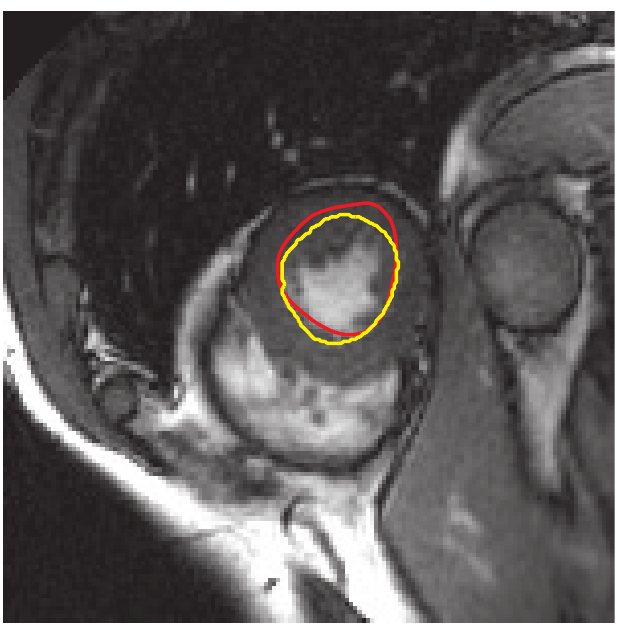} &
  \includegraphics[width=.19\linewidth,clip=true, trim=40 40 40
  40]{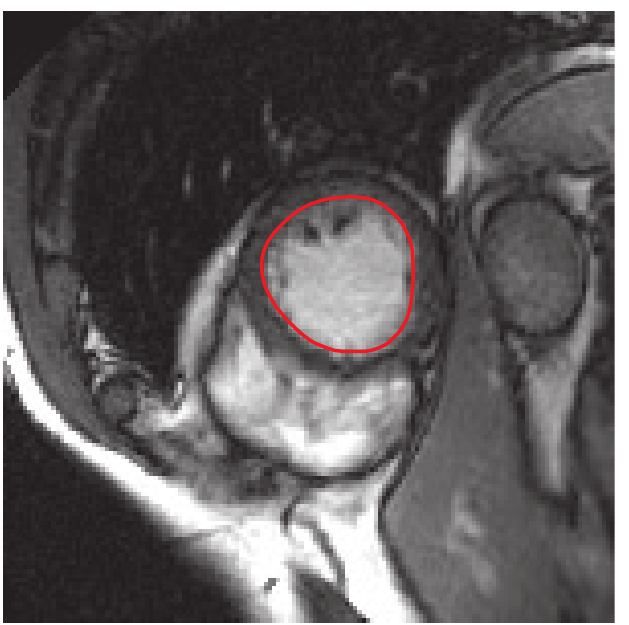} &
  \includegraphics[width=.19\linewidth,clip=true, trim=40 40 40
  40]{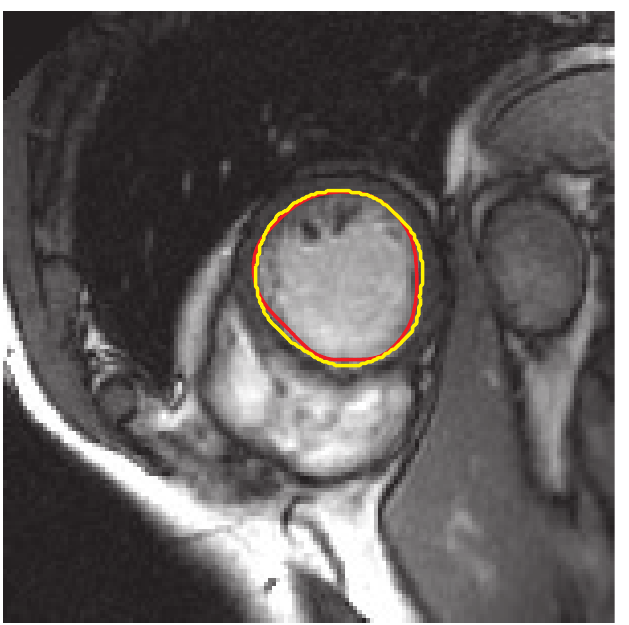}
\end{tabular}
}
\caption{{\bf Comparison on Tracking the Left Ventricle}. Two sample
results on the MICCAI LV Dataset of full cardiac cycles (only 5 out of
20 images are shown) of the proposed algorithm and Medviso.  The
ground truth when available is shown in yellow is superimposed when
available, and the red contour is the result obtained by the indicated
algorithm. Visual results indicate that our algorithm is better able
to handle non-homogenous appearance, and is thus more accurate.}
\label{fig:interactive_LV_visual}
\end{figure}

\begin{figure}
  \centering
  {\small
  \begin{tabular}{c@{}c@{}c@{}c@{}c}
    initial & \multicolumn{4}{c}{ventricle tracked (red - algorithm
      result, yellow - ground truth)} \\
    \rotatebox{90}{\quad\quad Medviso}
    \includegraphics[width=.19\linewidth,clip=true, trim=30 60 20
    30]{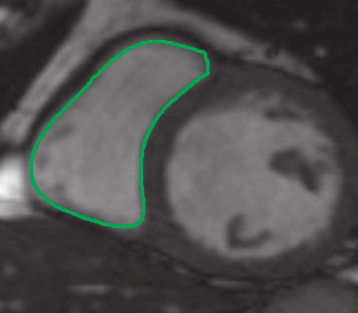} &
  \includegraphics[width=.19\linewidth,clip=true, trim=30 60 20
  30]{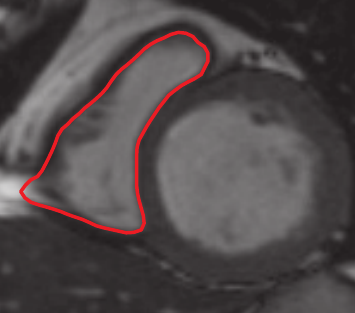} &
  \includegraphics[width=.19\linewidth,clip=true, trim=30 60 20
  30]{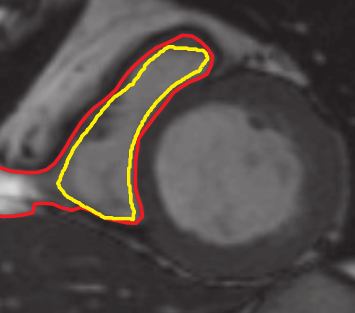} &
  \includegraphics[width=.19\linewidth,clip=true, trim=30 60 20
  30]{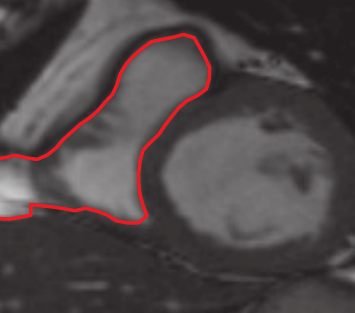} &
  \includegraphics[width=.19\linewidth,clip=true, trim=30 60 20
  30]{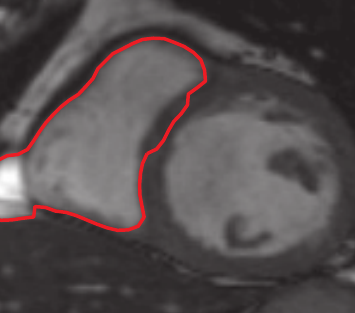} \\
  \rotatebox{90}{\quad\,\, our method}
  \includegraphics[width=.19\linewidth,clip=true, trim=30 60 20
  30]{compare_gt/1-IM-0100} &
  \includegraphics[width=.19\linewidth,clip=true, trim=30 60 20
  30]{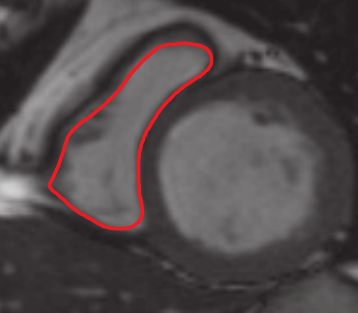} &
  \includegraphics[width=.19\linewidth,clip=true, trim=30 60 20
  30]{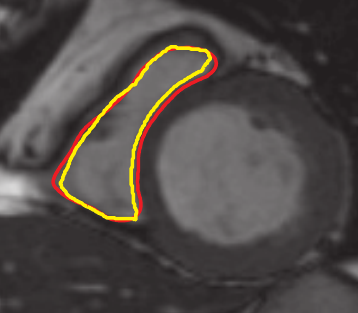} &
  \includegraphics[width=.19\linewidth,clip=true, trim=30 60 20
  30]{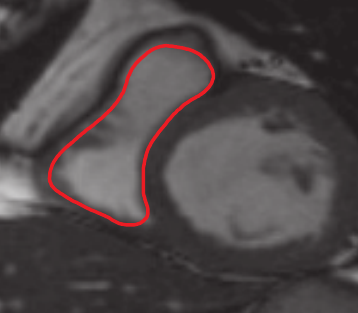} &
  \includegraphics[width=.19\linewidth,clip=true, trim=30 60 20
  30]{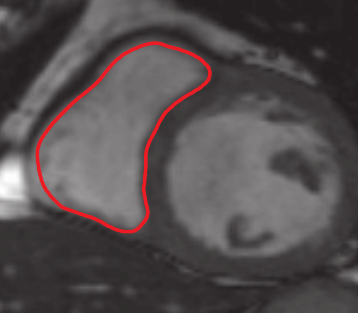} \\
  \rotatebox{90}{\quad\quad Medviso}
  \includegraphics[width=.19\linewidth,clip=true, trim=30 50 20
  30]{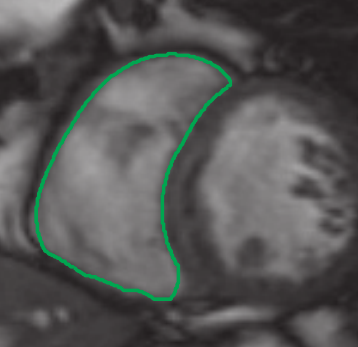} &
  \includegraphics[width=.19\linewidth,clip=true, trim=30 50 20
  30]{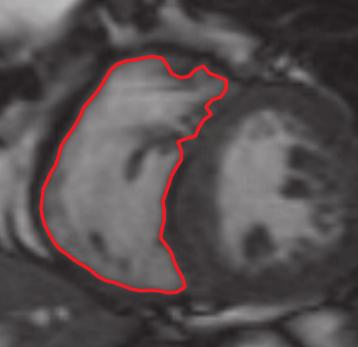} &
  \includegraphics[width=.19\linewidth,clip=true, trim=30 50 20
  30]{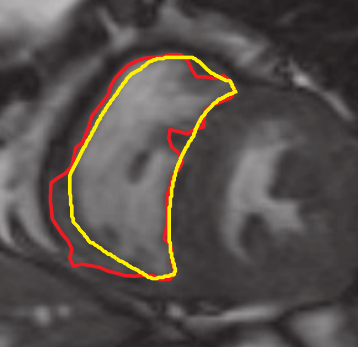} &
  \includegraphics[width=.19\linewidth,clip=true, trim=30 50 20
  30]{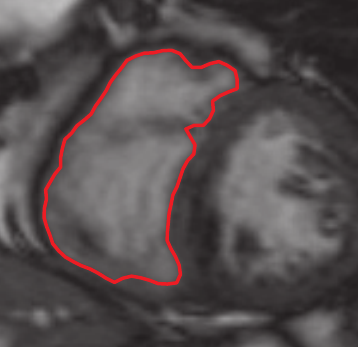} &
  \includegraphics[width=.19\linewidth,clip=true, trim=30 50 20
  30]{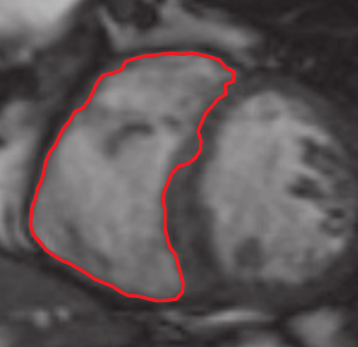} \\
  \rotatebox{90}{\quad\,\, our method}
  \includegraphics[width=.19\linewidth,clip=true, trim=30 50 20
  30]{compare_gt/5-IM-0100} &
  \includegraphics[width=.19\linewidth,clip=true, trim=30 50 20
  30]{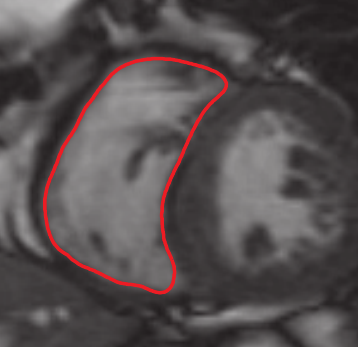} &
  \includegraphics[width=.19\linewidth,clip=true, trim=30 50 20
  30]{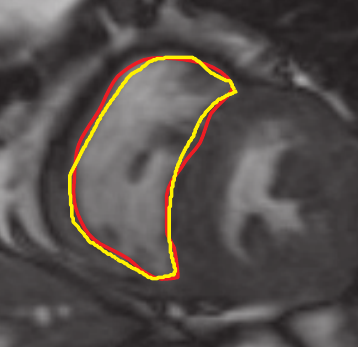} &
  \includegraphics[width=.19\linewidth,clip=true, trim=30 50 20
  30]{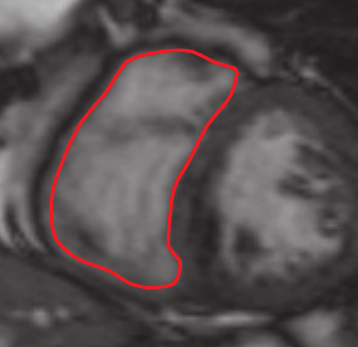} &
  \includegraphics[width=.19\linewidth,clip=true, trim=30 50 20 30]{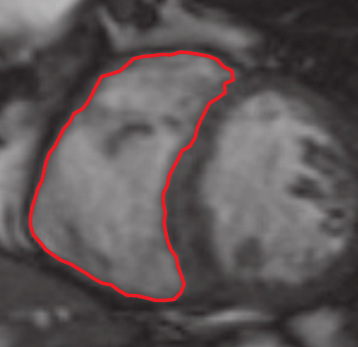}
\end{tabular}
}
\caption{{\bf Comparison on Tracking the Right Ventricle} Two sample
  results on the MICCAI RV Dataset of full cardiac cycles (only 5 out
  of 20 images are shown) of the proposed algorithm and Medviso.  The
  ground truth when available is shown in yellow is superimposed when
  available, and the red contour is the result obtained by the
  indicated algorithm. Visual results indicate that our method is less
  susceptible to clutter nearby the RV, and thus is able to capture
  the RV more accurately.}
\label{fig:interactive_RV_visual}
\end{figure}

Figures~\ref{fig:interactive_LV_visual} and
\ref{fig:interactive_RV_visual} shows some sample tracking results of
the proposed method and Medviso on full cardiac cycles of two
different cases on both the LV dataset and the RV dataset. The ground
truth (yellow) is superimposed when available. A summary of the
results on the entire datasets is shown in Table~\ref{table:results1}.
The accuracy with respect to ground truth is measured using average
perpendicular distance (APD) and dice metric (DM) for left ventricle,
and Hausdorff distance (HD) and DM for the right ventricle. These
metrics are chosen since they are the standard ones used on these
datasets.  Both qualitative and quantitative results show that our
proposed method leads to more accurate segmentation of the ventricles
and thus leads to less interaction than segmentation propagation
schemes in than Medviso.

\begin{table}
  \centering
  \begin{tabular}{lcccc}
    \hline
    &  \multicolumn{2}{c}{MICCAI LV} & \multicolumn{2}{c}{MICCAI RV} \\
    \hline
    mean $\pm$ std &  APD & DM & HD & DM\\
    \hline
    Ours   &  2.39$\pm$ .31 & .88$\pm$.02 & 6.72 $\pm$ 3.22 & .83$\pm$.15\\
    %Global &  2.47$\pm$ .35 & .88$\pm$.03 & 7.13 $\pm$ 3.64 & .82$\pm$.16\\
    MedViso &  4.68$\pm$ 1.2 & .78$\pm$.09 & 15.19 $\pm$ 6.08 & .73$\pm$.19\\
    \hline
  \end{tabular}

  \caption{{\bf Quantitative Evaluation} on the MICCAI LV validation
    \cite{miccai09} and MICCAI RV training \cite{miccai2012RV} datasets.
    Low APD/HD and high DM indicate good matches.}
  \label{table:results1}
\end{table}

\subsection{Multiple Region Segmentation: Full Heart Segmentation}

We now demonstrate our approach in performing challenging full heart
segmentation: segmentation of the ventricles and epicardium all in one
shot. Both the RV and epicardium are especially challenging as the
contrast of the RV and background is subtle in comparison to the LV,
and the myocardium wall near parts of the RV is very thin.  We are not
aware of another interactive method that is able to segment all
structures, and so we compare to Medviso even though the method is not
specifically tailored to the myocardium, but the method is generic and
is able to propagate a segmentation. Further, Medviso does not segment
multiple regions all at once and thus we perform separate segmentation
of the LV, RV and epicardium. Since ground truth is not available for
the outer wall of the myocardium in any standard dataset that we aware
of, we show visual comparison.

\begin{figure}
\begin{tikzpicture}[auto,node distance=1in]
  \node(one){
    \includegraphics[width=.18\linewidth,clip=true, trim=10 10 1 50]{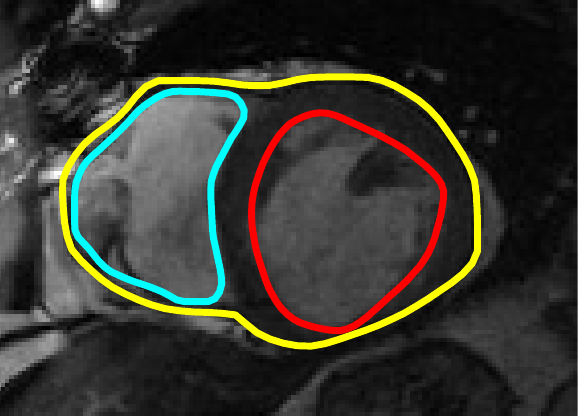}
    \includegraphics[width=.18\linewidth,clip=true, trim=10 10 1 50]{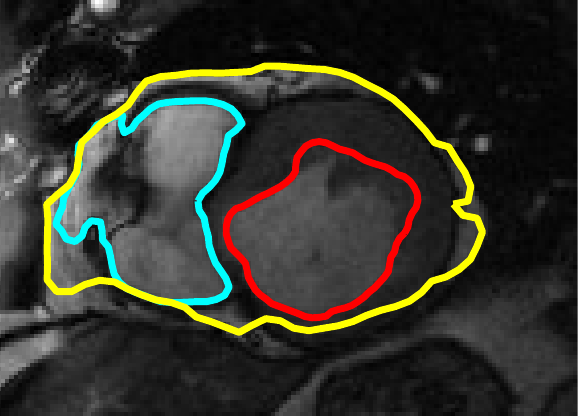}
    \includegraphics[width=.18\linewidth,clip=true, trim=10 10 1 50]{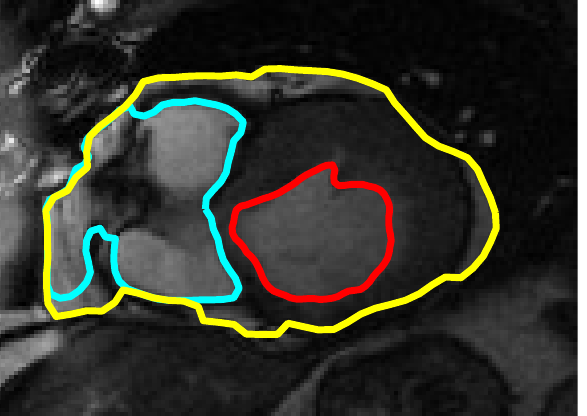}
    \includegraphics[width=.18\linewidth,clip=true, trim=10 10 1 50]{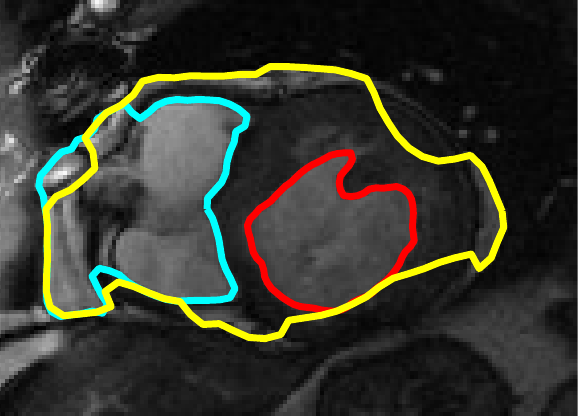}
    \includegraphics[width=.18\linewidth,clip=true, trim=10 10 1 50]{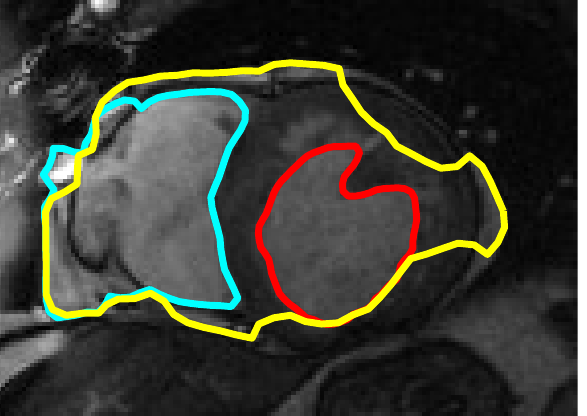}
  };
	\node[below of=one](two){
    \includegraphics[width=.18\linewidth]{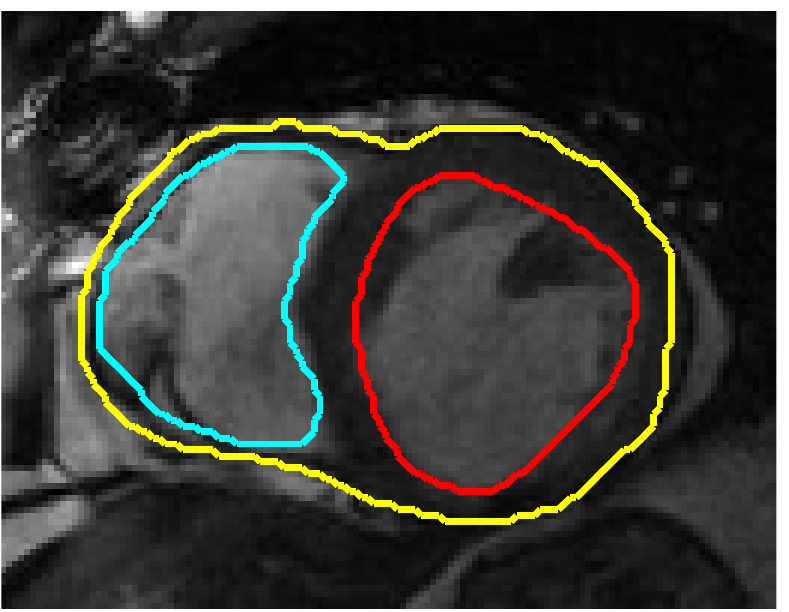}
    \includegraphics[width=.18\linewidth]{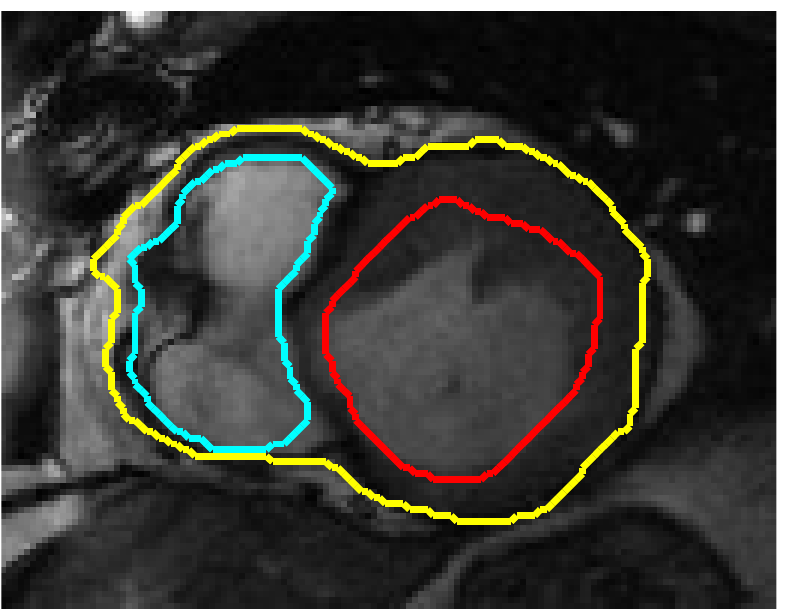}
    \includegraphics[width=.18\linewidth]{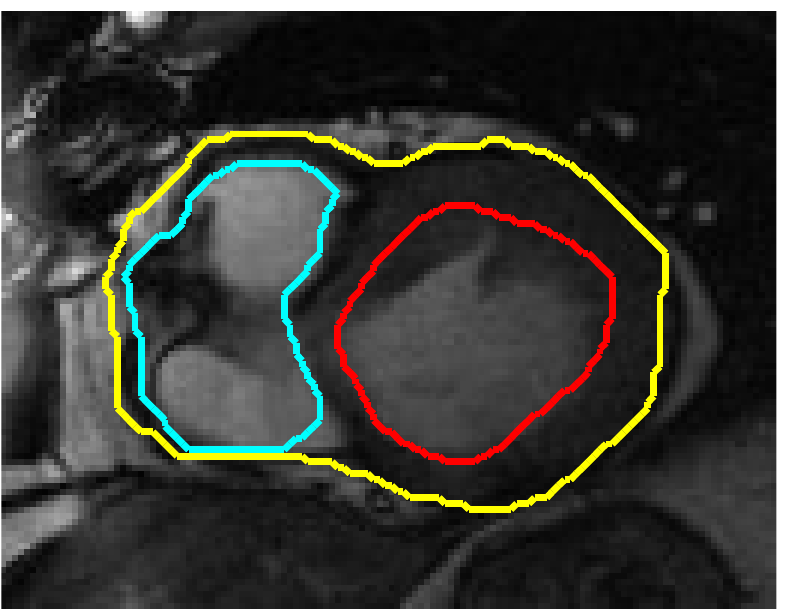}
    \includegraphics[width=.18\linewidth]{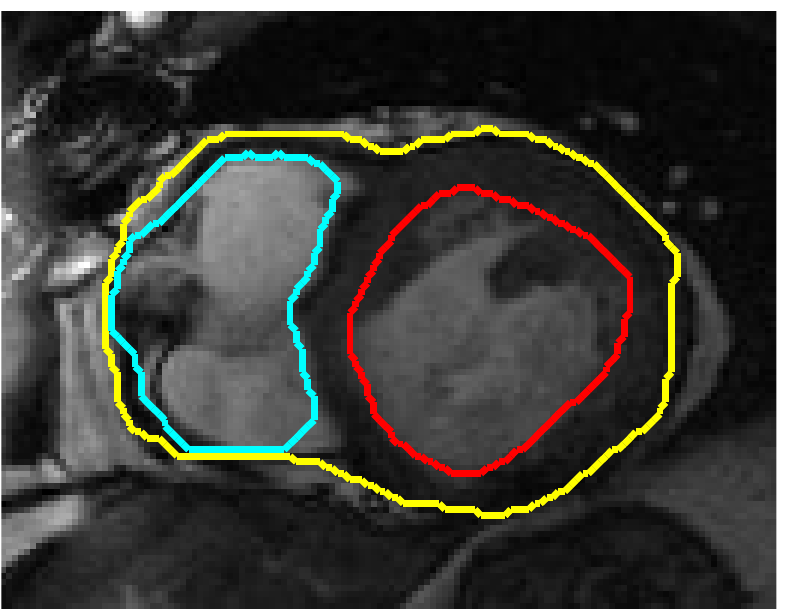}
    \includegraphics[width=.18\linewidth]{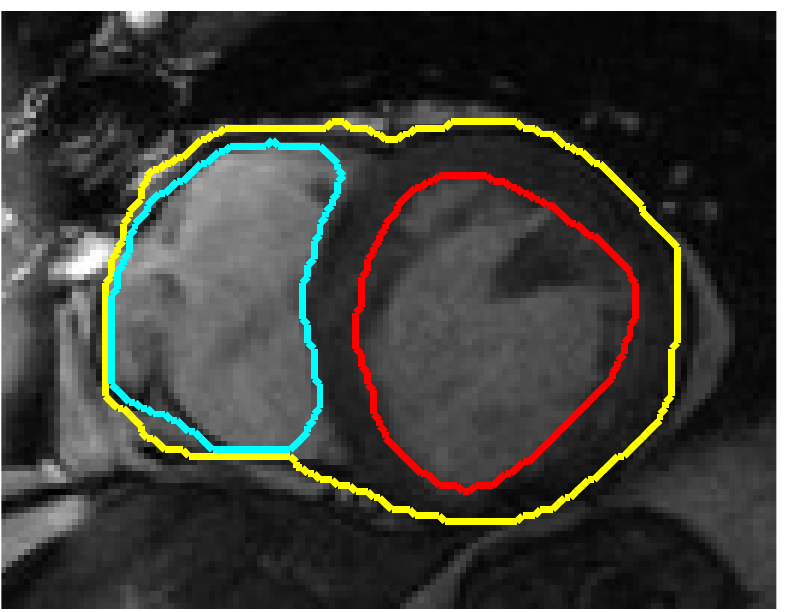}
  };
	
  \node[below of=two](three){
    \includegraphics[width=.18\linewidth,clip=true, trim=10 10 1 50]{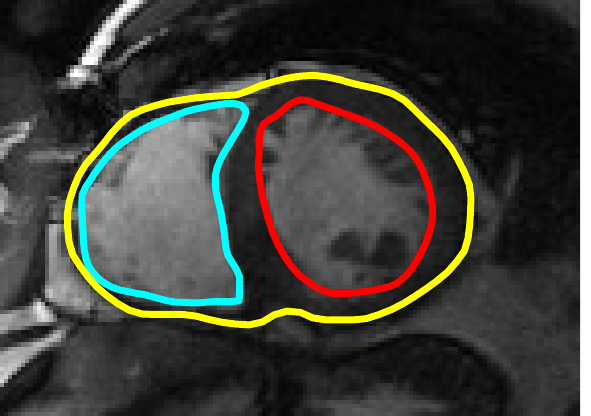}
    \includegraphics[width=.18\linewidth,clip=true, trim=10 10 1 50]{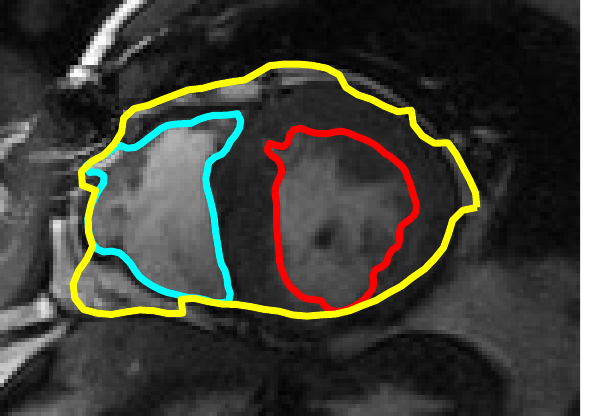}
    \includegraphics[width=.18\linewidth,clip=true, trim=10 10 1 50]{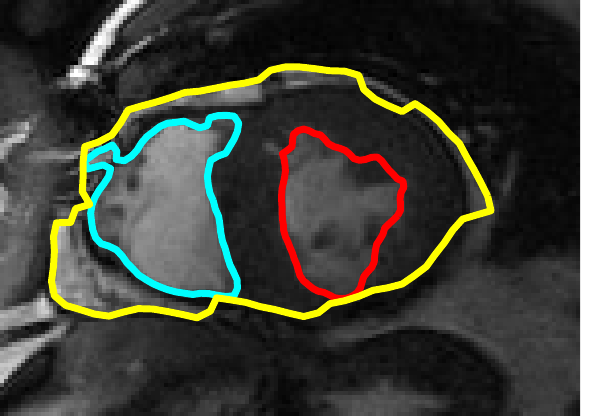}
    \includegraphics[width=.18\linewidth,clip=true, trim=10 10 1 50]{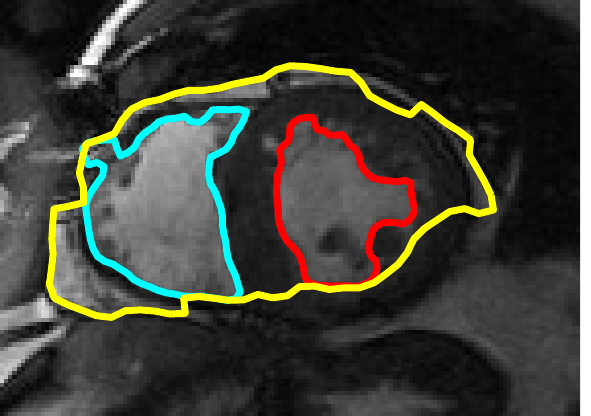}
    \includegraphics[width=.18\linewidth,clip=true, trim=10 10 1 50]{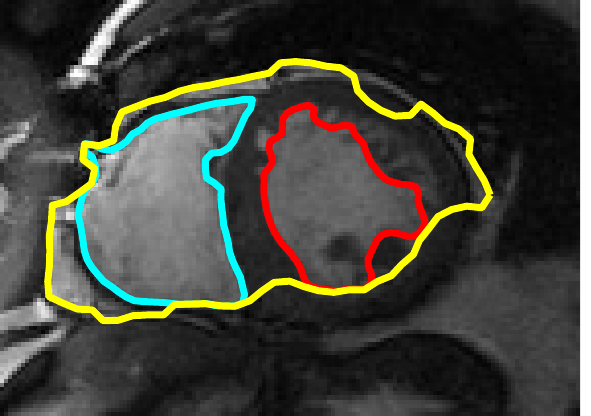}
  };
	\node[below of=three](four){
    \includegraphics[width=.18\linewidth]{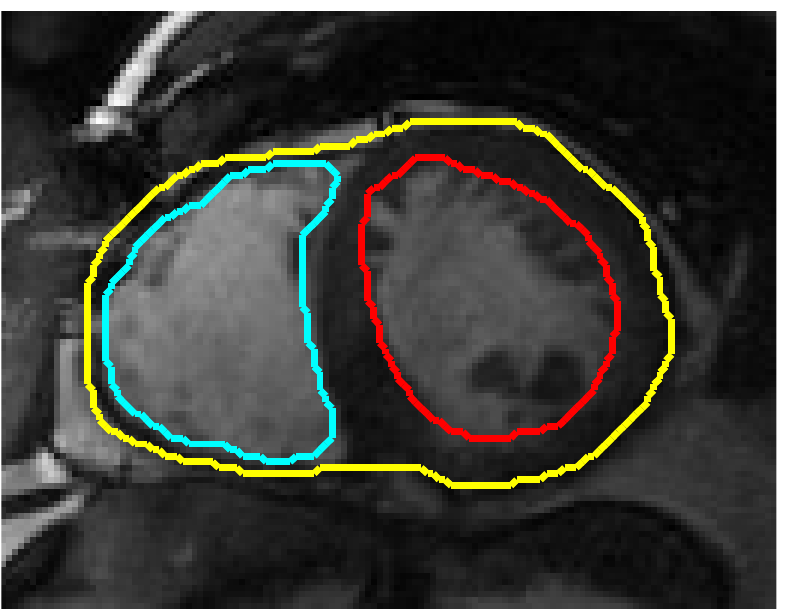}
    \includegraphics[width=.18\linewidth]{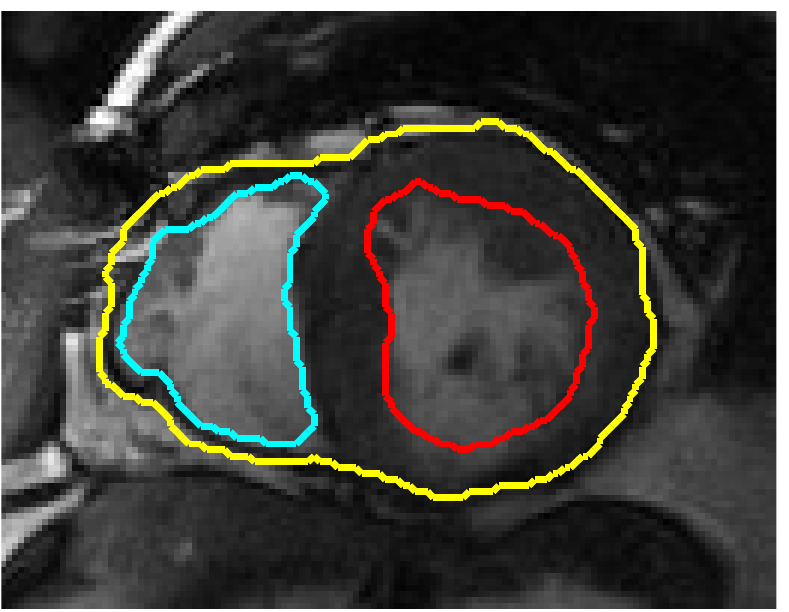}
    \includegraphics[width=.18\linewidth]{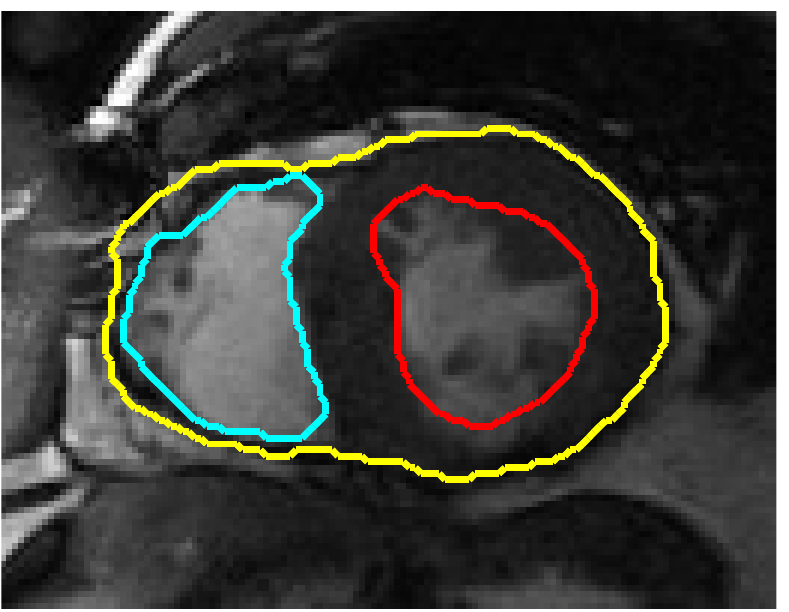}
    \includegraphics[width=.18\linewidth]{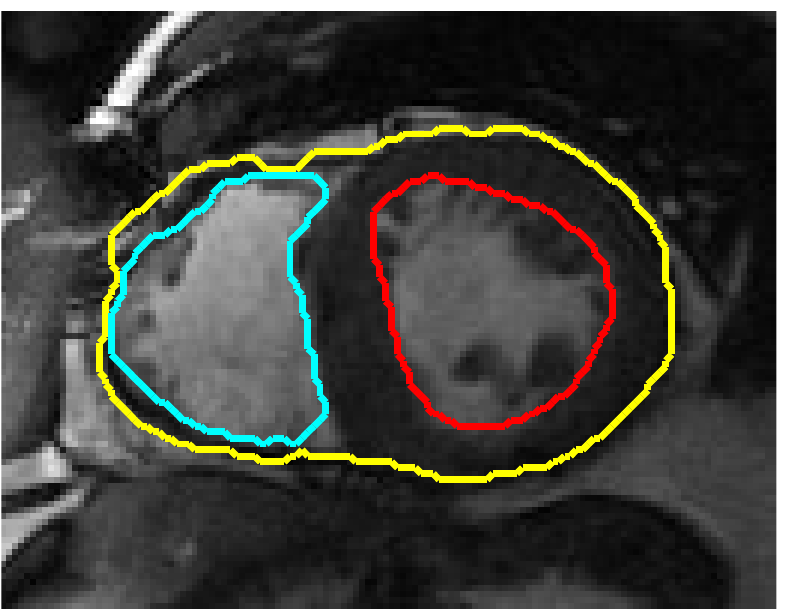}
    \includegraphics[width=.18\linewidth]{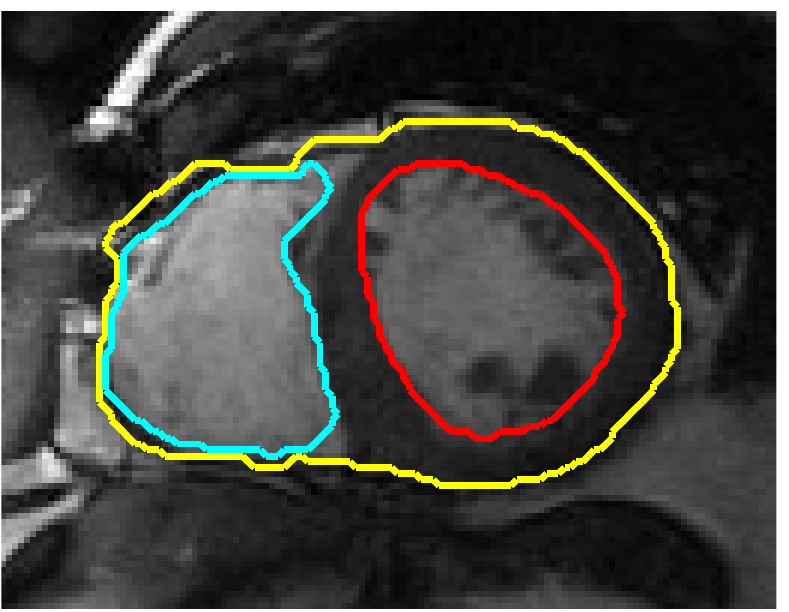}
  };
	
  \node[below of=four](five){
    \includegraphics[width=.18\linewidth,clip=true, trim=10 10 1 50]{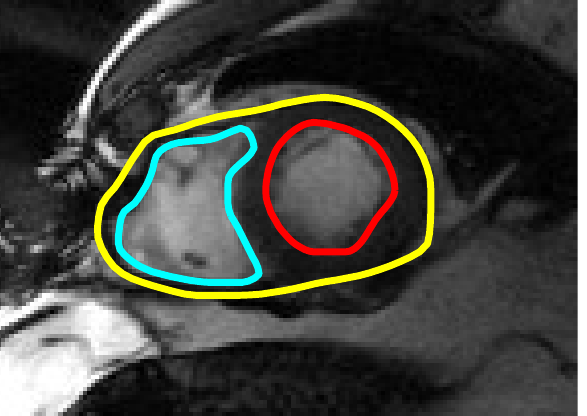}
    \includegraphics[width=.18\linewidth,clip=true, trim=10 10 1 50]{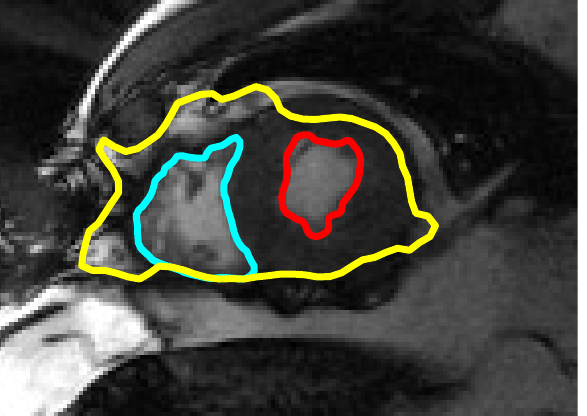}
    \includegraphics[width=.18\linewidth,clip=true, trim=10 10 1 50]{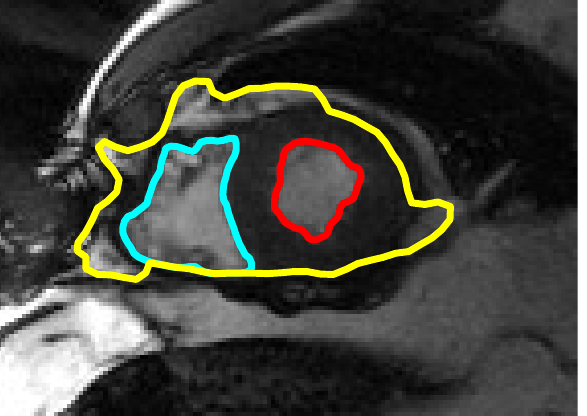}
    \includegraphics[width=.18\linewidth,clip=true, trim=10 10 1 50]{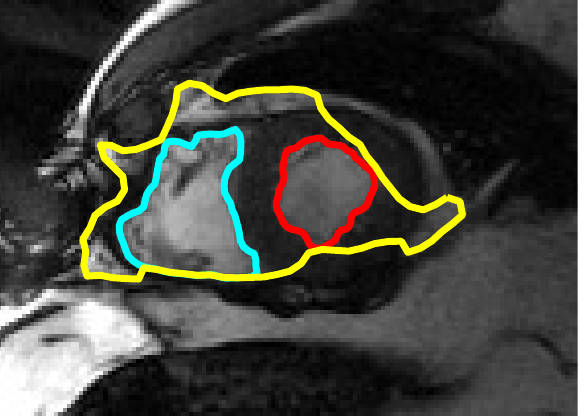}
    \includegraphics[width=.18\linewidth,clip=true, trim=10 10 1 50]{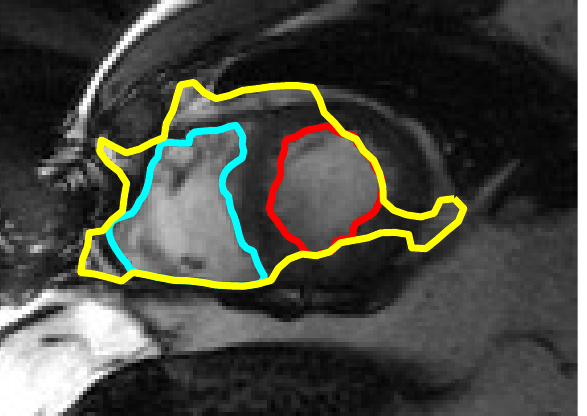}
  };
	\node[below of=five](six){
    \includegraphics[width=.18\linewidth]{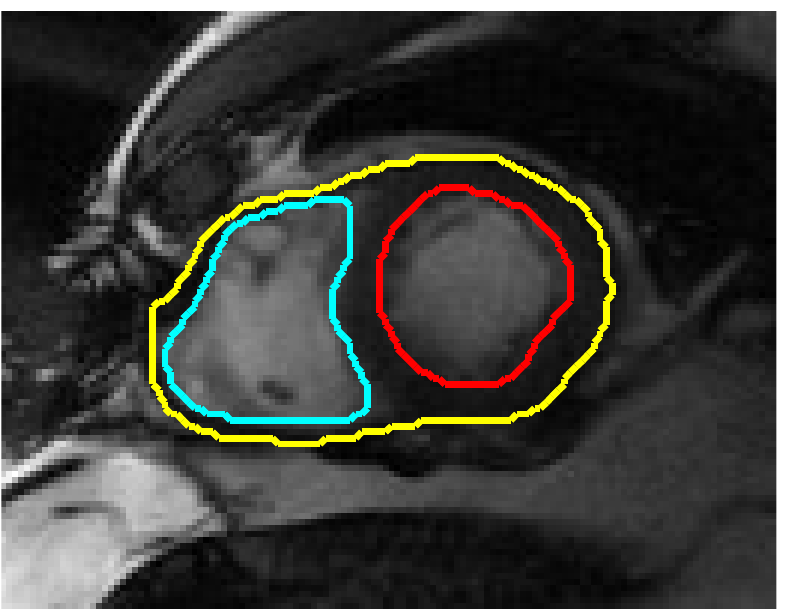}
    \includegraphics[width=.18\linewidth]{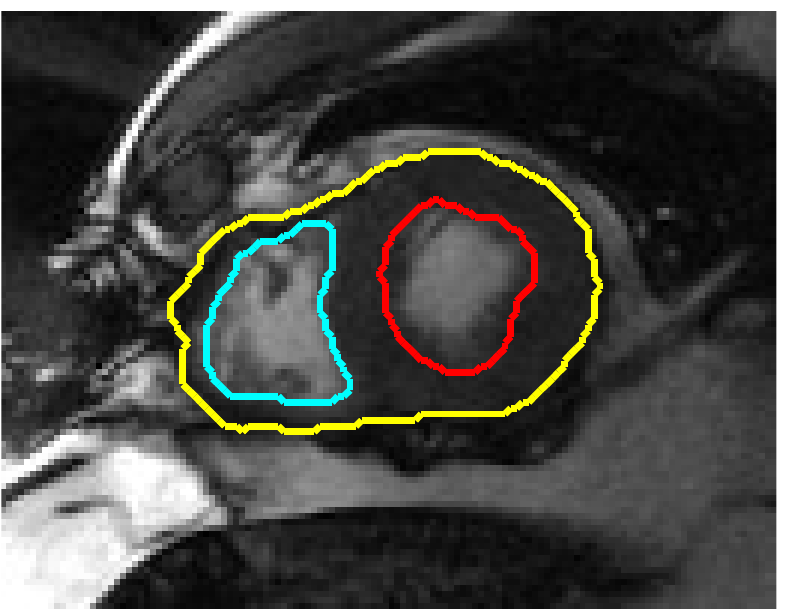}
    \includegraphics[width=.18\linewidth]{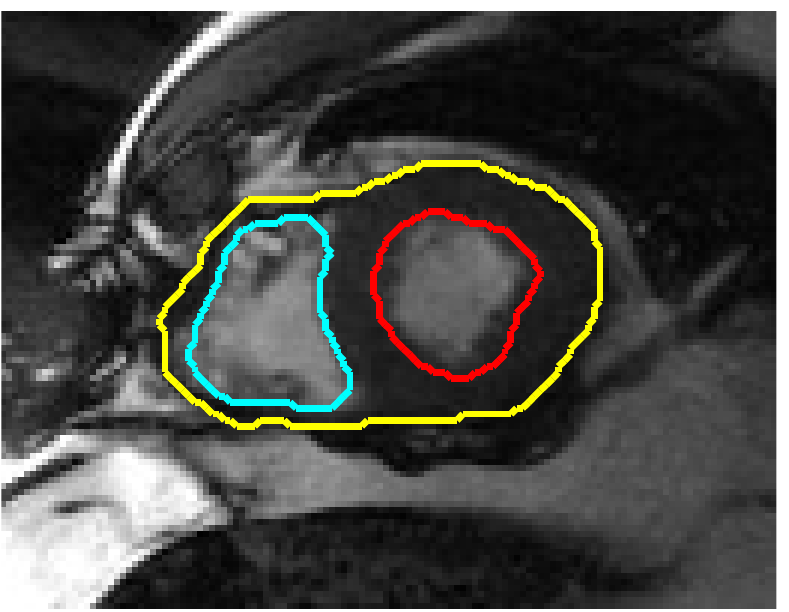}
    \includegraphics[width=.18\linewidth]{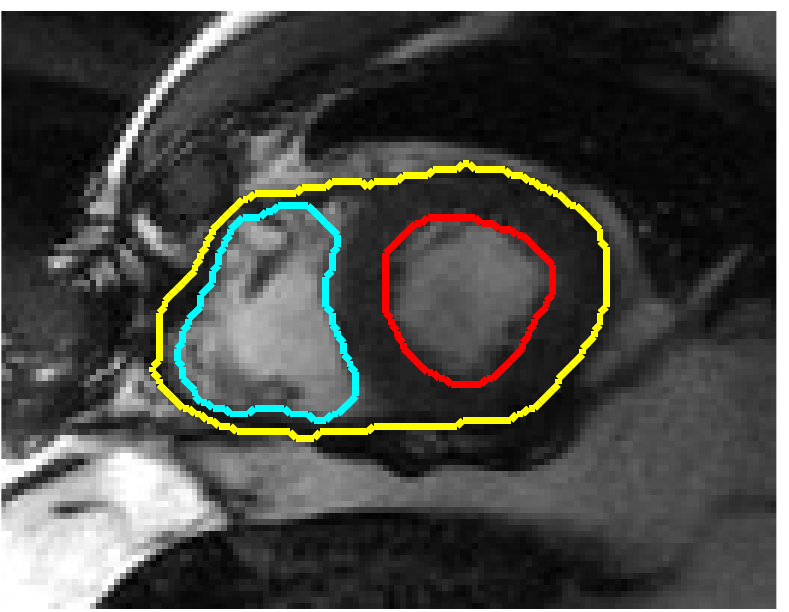}
    \includegraphics[width=.18\linewidth]{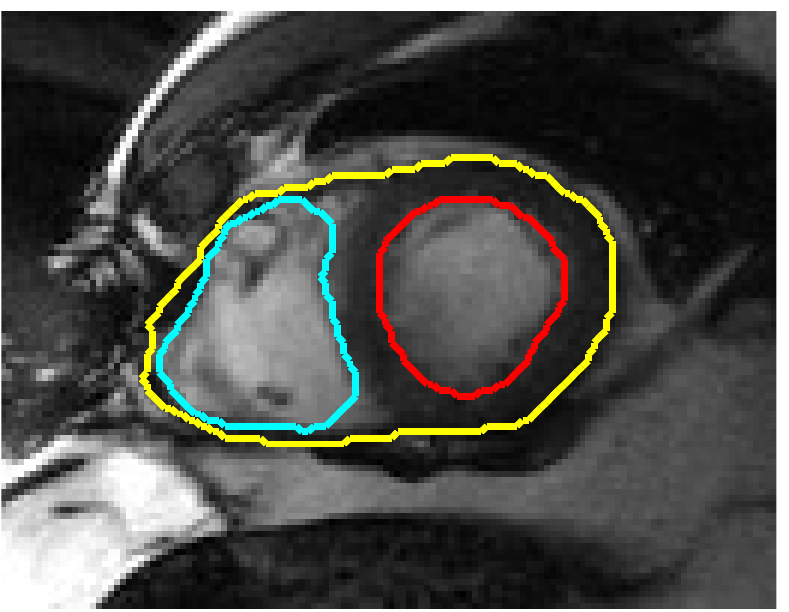}
  };
	
  \node[below of=six](seven){
    \includegraphics[width=.18\linewidth,clip=true, trim=10 10 1 50]{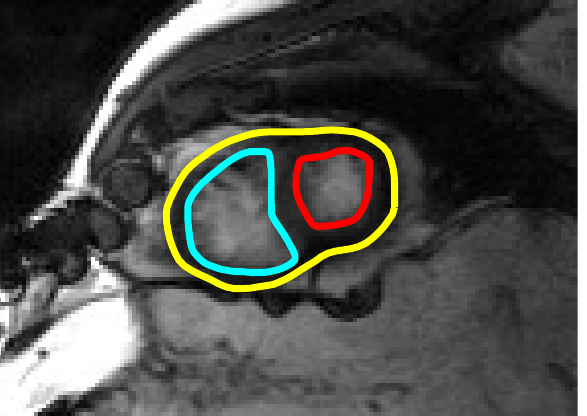}
    \includegraphics[width=.18\linewidth,clip=true, trim=10 10 1 50]{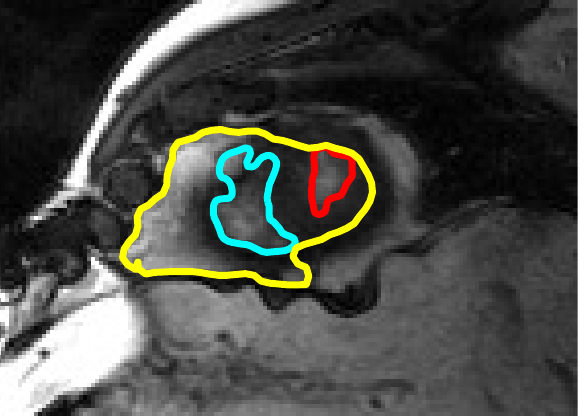}
    \includegraphics[width=.18\linewidth,clip=true, trim=10 10 1 50]{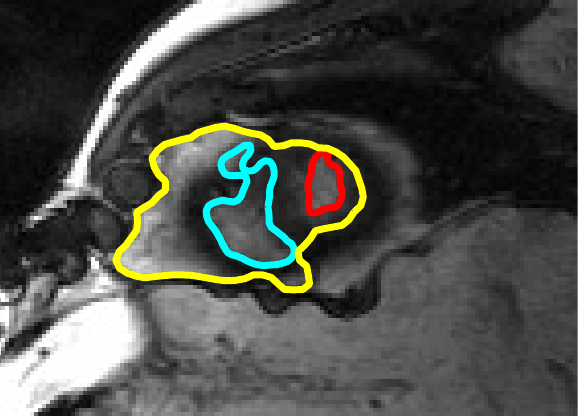}
    \includegraphics[width=.18\linewidth,clip=true, trim=10 10 1 50]{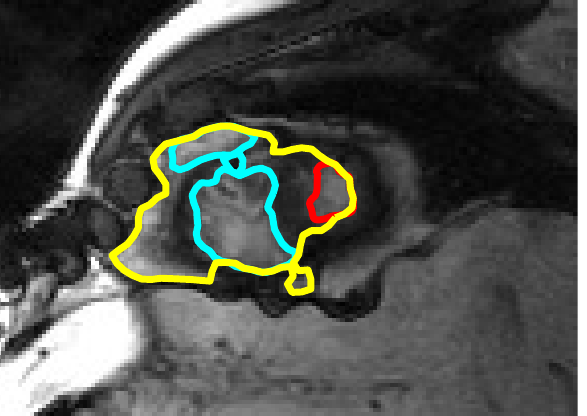}
    \includegraphics[width=.18\linewidth,clip=true, trim=10 10 1 50]{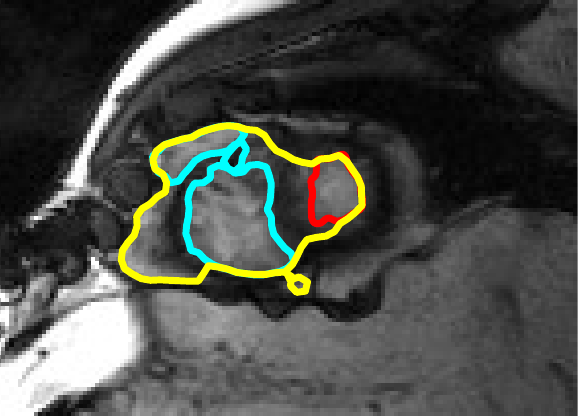}
  };
	\node[below of=seven](eight){
    \includegraphics[width=.18\linewidth]{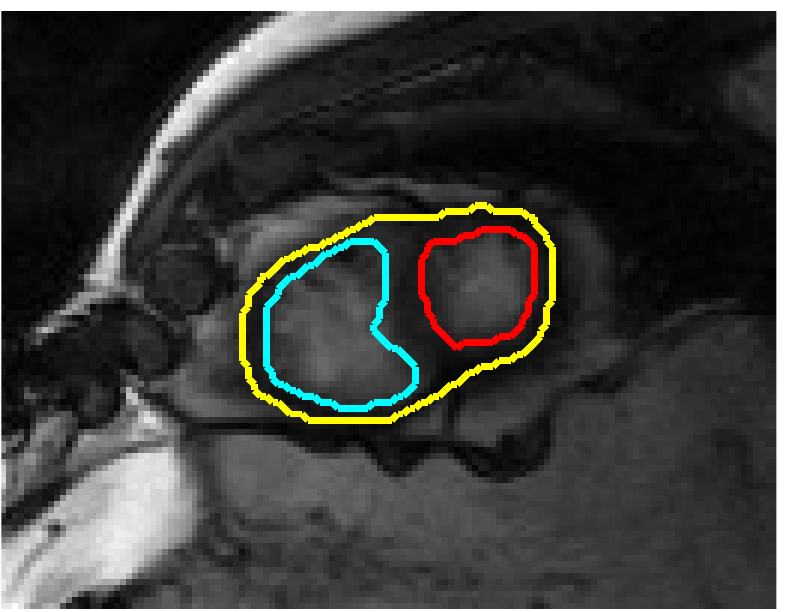}
    \includegraphics[width=.18\linewidth]{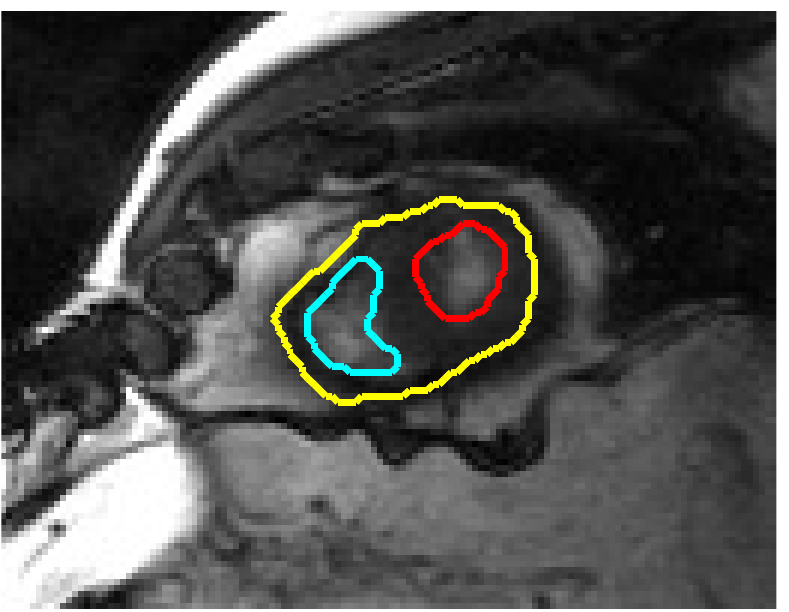}
    \includegraphics[width=.18\linewidth]{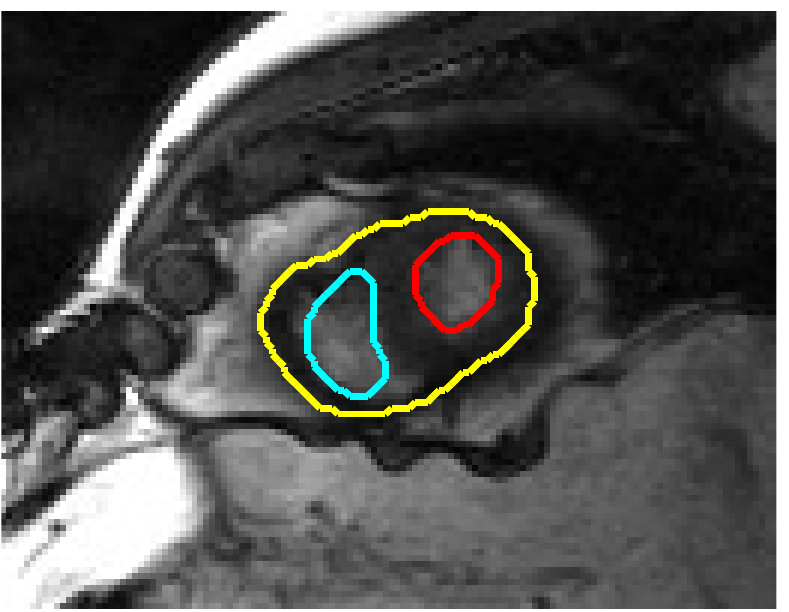}
    \includegraphics[width=.18\linewidth]{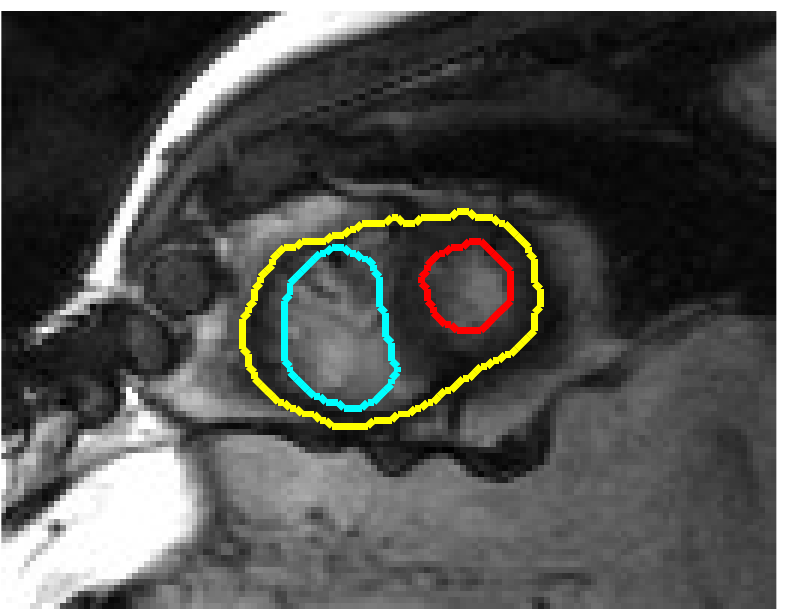}
    \includegraphics[width=.18\linewidth]{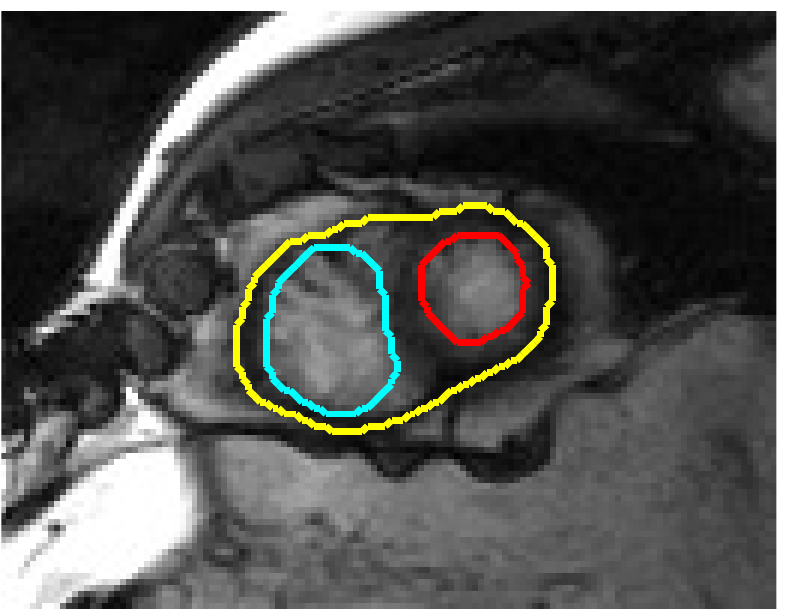}
  };
	
  \node[right of =one, rotate=90, node distance=3.4in]{Medviso};
  \node[right of =two, rotate=90, node distance=3.4in]{Ours};
  
  \node[right of =three, rotate=90, node distance=3.4in]{Medviso};
  \node[right of =four, rotate=90, node distance=3.4in]{Ours};
	
  \node[right of =five, rotate=90, node distance=3.4in]{Medviso};
  \node[right of =six, rotate=90, node distance=3.4in]{Ours};
	
  \node[right of =seven, rotate=90, node distance=3.4in]{Medviso};
  \node[right of =eight, rotate=90, node distance=3.4in]{Ours};
\end{tikzpicture}

\caption{{\bf Comparison on Multiple Region Segmentation}. [Top to
  Bottom]: $1^{st},3^{rd},6^{th}$ and $8^{th}$ slices shown. The LV
  (red), RV (cyan) and myocardium outer boundary (yellow) are
  simultaneously segmented using our proposed technique. Comparison
  is shown to Medviso. Visual results indicate that our technique is
  more accurate in segmenting all structures.}
\label{fig:compare2medviso}
\end{figure}

\begin{figure}
  \centering
  tracked in time $\rightarrow$\\
  \rotatebox{90}{\quad\quad Medviso}\,\,
  \includegraphics[width=.18\linewidth]{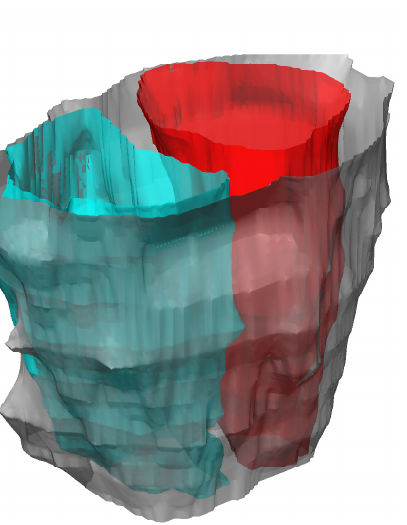}
  \includegraphics[width=.18\linewidth]{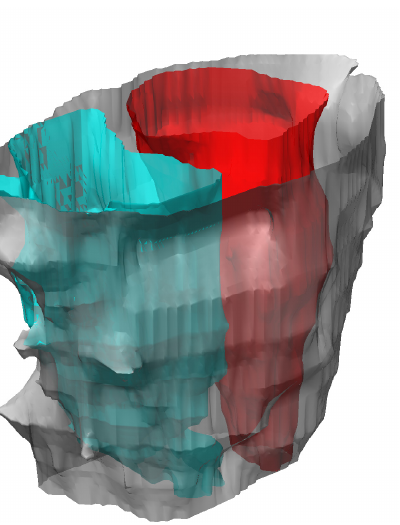}
  \includegraphics[width=.18\linewidth]{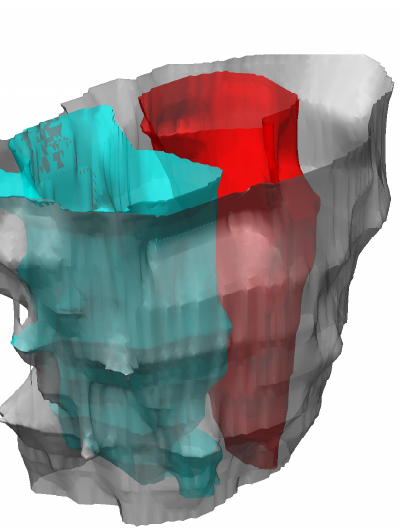}
  \includegraphics[width=.18\linewidth]{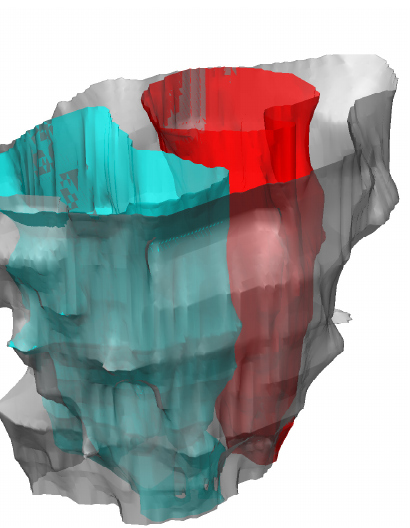}
  \includegraphics[width=.18\linewidth]{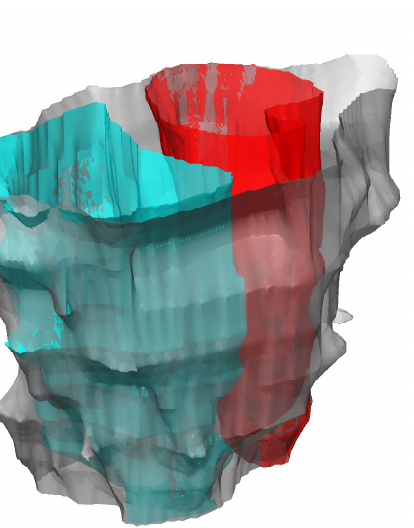}\\
  \rotatebox{90}{\quad\quad our method}\,\,
  \includegraphics[width=.18\linewidth]{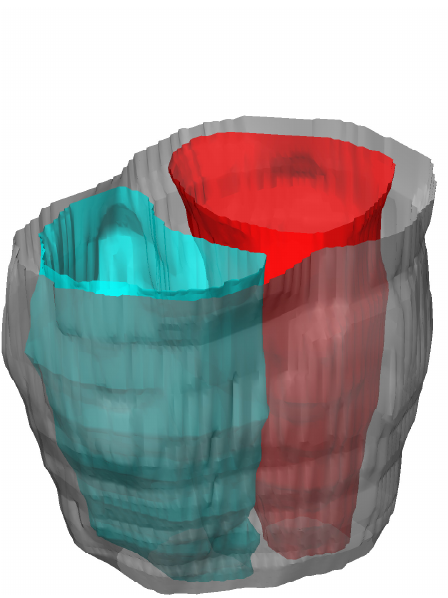}
  \includegraphics[width=.18\linewidth]{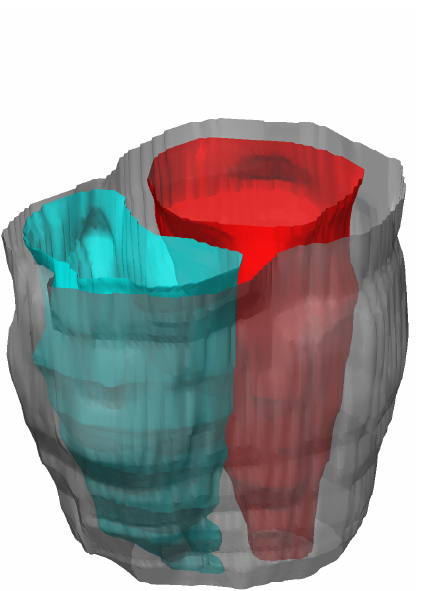}
  \includegraphics[width=.18\linewidth]{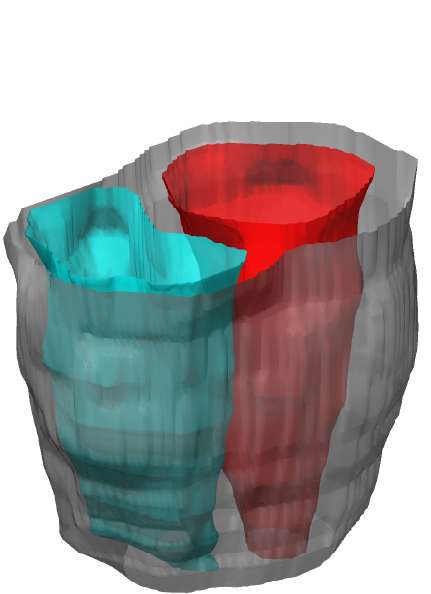}
  \includegraphics[width=.18\linewidth]{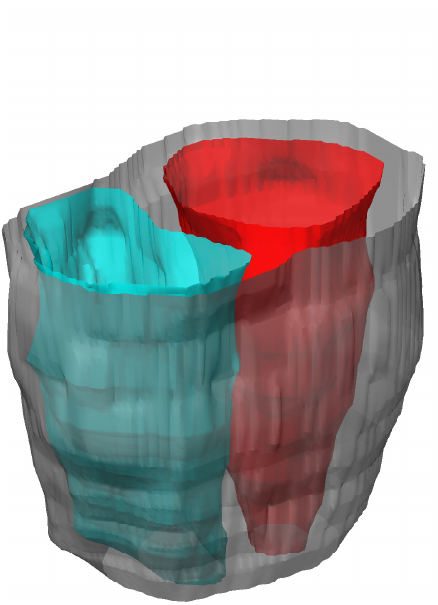}
  \includegraphics[width=.18\linewidth]{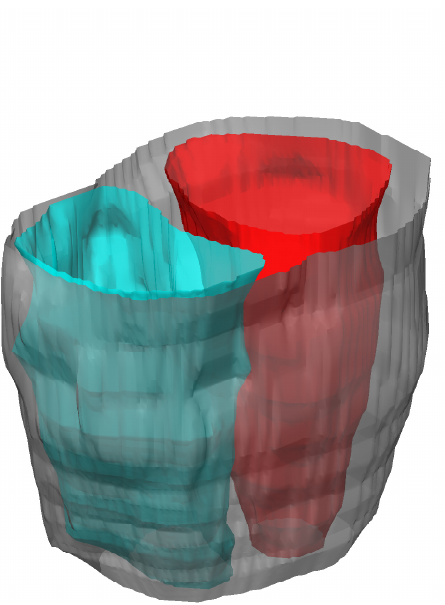}
  \caption{{\bf Comparison on Multiple Region Segmentation (3D
      Visualization)}. [Top row]: Medviso result, [Bottom row]: our
    result. Grey: Myocardium, cyan: right ventricle, red: left
    ventricle. Only 5 of 20 frames shown. Results indicate that the
    results of our method resemble heart structure.}
  \label{fig:3d}
\end{figure}

Figure~\ref{fig:compare2medviso} shows the slice-wise results of our
method and Medviso on a full 3D cardiac MRI sequence for a full
cardiac cycle. Results indicate that our method is more accurate in
capturing the shape of the ventricles and epicardium, and our method
is especially more promising on the RV and epicardium.
Figure~\ref{fig:3d} shows visualization of the results in 3D, and that
our method more accurately resembles the structure of the heart.

%Figure \ref{fig:samples} shows some more examples of tracking results
%obtained using our method.

\comment{
\begin{figure}
  \centering
  \includegraphics[width=.18\linewidth]{multi/SC-HF-I-05-IM-0081}
  \includegraphics[width=.18\linewidth]{multi/SC-HF-I-05-IM-0085}
  \includegraphics[width=.18\linewidth]{multi/SC-HF-I-05-IM-0090}
  \includegraphics[width=.18\linewidth]{multi/SC-HF-I-05-IM-0095}
  \includegraphics[width=.18\linewidth]{multi/SC-HF-I-05-IM-0100}
	
  \includegraphics[width=.18\linewidth]{multi/e2/SC-HF-I-05-IM-0101}
  \includegraphics[width=.18\linewidth]{multi/e2/SC-HF-I-05-IM-0105}
  \includegraphics[width=.18\linewidth]{multi/e2/SC-HF-I-05-IM-0110}
  \includegraphics[width=.18\linewidth]{multi/e2/SC-HF-I-05-IM-0115}
  \includegraphics[width=.18\linewidth]{multi/e2/SC-HF-I-05-IM-0120}
  
  \caption{{\bf Left and right ventricle and myocardium tracking}
    using the proposed method. 5 of 20 frames shown.}
  \label{fig:samples}
\end{figure}
}

\section{Conclusion}
We have presented an algorithm for propagating the segmentation from
one frame in an image sequence to another via a novel registration
algorithm. The registration is physically motivated by the multi-modal
motions among sub-structures and the physical constraints between
motions in adjacent regions, specifically the matching conditions of
normal velocities. Traditional registration algorithms apply global
regularization smoothing across region boundaries, mixing motions of
differing sub-structures, are not physically motivated, and therefore
yield inaccurate registrations and therefore segmentation
propagations. The presented technique solves the registration via a
variational formulation that results in PDEs in regions coupled via
boundary conditions that resemble Robin boundary conditions. This
leads to a computationally efficient technique that has the same cost
as traditional regularization.

Experiments have shown that our method is more effective than global
regularization in propagating segmentations in cardiac MRI data of the
heart. Moreover, the main motivation for this work has been to improve
existing interactive segmentation techniques, which are commonly used
commercially, for cardiac MRI segmentation by better predicting a
segmentation in the next frame from the current frame. We have compared
our technique both qualitatively and quantitatively against a recent
and widely used commercial software, Medviso, and results indicate
that our method would require less manual interaction for segmentation
correction, specifically in LV, RV and epicardium segmentation.

\appendices

\section{Proof of Positive-Definiteness of Motion Operators}
\label{app:positive_definiteness}
This appendix shows that the PDEs with the boundary conditions
\eqref{eq:vin_hard_PDE} and \eqref{eq:vout_hard_PDE} (and similarly
\eqref{eq:vin_soft_PDE} and \eqref{eq:vout_soft_PDE} in the previous
subsection) can be solved with a conjugate gradient solver by showing
the corresponding linear operators are positive-semi definite. This
can be shown by considering the following standard $\mathbb L^2$ inner
product on the space of admissible velocities $v_i,v_o$:
\[
\ip{v^1}{v^2}{} = \int_R v^1_i\cdot v^2_i \ud x + 
\int_{\Omega\backslash R} v^1_o\cdot v^2_o \ud x.
\]
Given the operator $A = (-\alpha_i \Delta + \nabla I \nabla I^T, \, -\alpha_o \Delta + \nabla I \nabla I)$ acting on $v=(v_i,v_o)$, positive semi-definiteness is shown by verifying $\ip{Av}{v}{}\geq 0$:
\begin{align*}
  \ip{Av}{v}{} &= \int_R (-\alpha_i \Delta v_i + \nabla I \nabla I^Tv_i)\cdot v_i \ud x
  + \int_R (-\alpha_o \Delta v_o + \nabla I \nabla I^Tv_o)\cdot v_o \ud x\\
  &= \int_R \alpha_i |\nabla v_i|^2 + (\nabla I \cdot v_i)^2 \ud x + 
  \int_{\Omega\backslash R} \alpha_o |\nabla v_o|^2 + (\nabla I \cdot v_o)^2 \ud x\\
  &-\int_{\partial R} \alpha_i (\nabla v_i \cdot N)\cdot v_i - \alpha_o (\nabla v_o \cdot N)\cdot v_o \ud s \\
  &= \int_R \alpha_i |\nabla v_i|^2 + (\nabla I \cdot v_i)^2 \ud x + 
  \int_{\Omega\backslash R} \alpha_o |\nabla v_o|^2 + (\nabla I \cdot v_o)^2 \ud x \geq 0.
\end{align*}
The last equality is due to the vanishing boundary integrals, which is
obtained by noting the boundary conditions
\eqref{eq:boundary_soft_first} and \eqref{eq:boundary_soft_last} or
\eqref{eq:boundary_hard_first}-\eqref{eq:boundary_hard_last}. Indeed,
noting \eqref{eq:boundary_soft_first} and
\eqref{eq:boundary_soft_last}, we have that $\alpha_i (\nabla v_i
\cdot N) = \alpha_o (\nabla v_o \cdot N)$, which implies the vanishing
boundary term above. Similarly,
\eqref{eq:boundary_hard_first}-\eqref{eq:boundary_hard_last} implies
that
\[
\alpha_i (\nabla v_i \cdot N)\cdot v_i - \alpha_o (\nabla v_o \cdot N)\cdot v_o = 
\alpha_i (\nabla v_i \cdot N)\cdot N (v_i\cdot N) - \alpha_o
(\nabla v_o \cdot N)\cdot N (v_o\cdot N) =0.
\]
Thus, since positive definiteness is shown, the conjugate gradient
algorithm may be applied.

\bibliographystyle{ieee}
\bibliography{miccai_ref,cardiac_interactive}

\end{document}